UNIVERSITY OF CALIFORNIA

Los Angeles

Signal Use and Emergent Cooperation

A dissertation submitted in partial satisfaction of the

requirements for the degree Doctor of Philosophy

in Computer Science

by

Michael Williams

2006

The dissertation of Michael Williams is approved.

_______________________________________

Milos Ercegovac

_______________________________________

Petros Faloutsos

_______________________________________

Charles Taylor

_______________________________________

Michael Dyer, Committee Chair

University of California, Los Angeles

2006

TABLE OF CONTENTS

# LIST OF FIGURES

# LIST OF TABLES

# ACKNOWLEDGEMENTS

Firstly, I would like to thank my advisor Michael Dyer for the direction he provided and for helping me make this dissertation the best it could be.

I would also like to thank my committee for their interest in my research and for their participation, and I'd like to thank the rest of the Computer Science department, faculty and staff, for everything they taught me.

Finally, I would also like to thank UCLA for a decade full of trial and triumph.

# VITA

December 7, 1977          Born, Los Angeles, California

1998-2001                 Software Engineer

                          Credit Interlink of America

                          Los Angeles, California

1999                      B.S., Computer Science

                          University of California, Los Angeles

2001                      M.S., Computer Science

                          University of California, Los Angeles

2001-2005                 Consulting Engineer

                          Boeing Satellite Systems

                          El Segundo, California

2005-present              Software Engineer

                          Advanced Software Systems

                          Northrop Grumman Corporation

                          El Segundo, California

ABSTRACT OF THE DISSERTATION

Signal Use and Emergent Cooperation

by

Michael Williams

Doctor of Philosophy in Computer Science

University of California, Los Angeles, 2006

Professor Michael Dyer, Chair

This dissertation provides insight into the use of signals by Distributed Autonomous Communicators (DACs) using the NEC-DAC system. NEC-DAC is an acronym that stands for *Neurally Encoded Culture - Distributed Autonomous Communicators*. The NEC-DAC animats learn to use signals to coordinate their activities in a way that increases the efficiency of their tribe and to then transmit their culture to new tribe members. The animats (or agents) are organized into tribes but are autonomous because they each contain their own neural decision-making engine. They operate independently from each other and share information via discrete communication signals. Communication is essential for cooperation, but in order for communication to be effective an organism must not only have the capability to transmit information to its peers, it must also know when and how to use the capabilities is possesses. Each animat encodes its individual understanding of its tribe's shared behavioral system within its neural networks, and this system of shared behavior can be considered a *culture* because it is acquired through learning and propagated by signals. My research is focused on how culture self-organizes (emerges) within a population of creatures and on how communication increases fitness in combination with varying behavioral capabilities. By comparing several different social (authority) structures I demonstrate that a group's culture of cooperation can have a significant effect on performance, and that the use of signals contributes not just to the emergence of culture but also to the transmission of culture across generations. I also examine the utility of coordinating behavior and signaling between the neural networks in a single animat.

# 1  Introduction

This dissertation provides insight into the use of signals by Distributed Autonomous Communicators (DACs) using the NEC-DAC system.  NEC-DAC is an acronym that stands for *Neurally Encoded Culture - Distributed Autonomous Communicators*.  The NEC-DAC animats learn to use signals to coordinate their activities in a way that increases the efficiency of their tribe and to then transmit their culture to new tribe members.   The animats (or agents) are organized into tribes but are autonomous because they each contain their own neural decision-making engine.  They operate independently from each other and share information via discrete communication signals.  Communication is essential for cooperation, but in order for communication to be effective an organism must not only have the capability to transmit information to its peers, it must also know when and how to use the capabilities is possesses.  Each animat encodes its individual understanding of its tribe's shared behavioral system within its neural networks, and this system of shared behavior can be considered a *culture* because it is acquired through learning and propagated by signals.

There are seven major sections:

1.  The *Introduction*, which you are enjoying right now.

2.  A *Project Overview* that describes the aspects of distributed autonomous communicators that I explored and why my conclusions are significant.

3.  A description of the *NEC-DAC Architecture* and how it is suited to the issues explored.

4.  The *Experiments* I performed and their results.

5. The *Conclusions* I drew from the experiments along with an explanation of NEC-DAC's current limitations and how the system relates to other work in the field.
6. The *Appendices*.
7. The *References*.

My research is focused on how culture self-organizes (emerges) within a population of creatures and on how communication increases fitness in combination with varying behavioral capabilities.  By comparing several different social (authority) structures I demonstrate that a group's culture of cooperation can have a significant effect on performance, and that the use of signals contributes not just to the emergence of culture but also to the transmission of culture across generations.  I also examine the utility of coordinating behavior and signaling between the neural networks in a single animat.

There are two components to my research:
1. Creation of an architecture that can learn to handle the large problem space required for representing an interesting set of signaling and behavioral capabilities.
2. Development of experimental models that will serve as examples of how culture can emerge within a population whose members possess various types of signaling capabilities and the ability to learn.

The philosophy behind my design is that the animats are similar to chimpanzees: they have the "physical" tools required for advantageous communication (Savage-Rumbaugh, Rumbaugh & Boysen, 1980), but need to learn to use those tools through experience (by developing a culture). I have abstracted many common signaling considerations so as to focus primarily on how groups (not individuals) can benefit from signaling and cooperation and on how cooperation can arise from suitable building blocks.

## 2  Project Overview

I have created a series of experiments in which animats organized into tribes compete for scarce resources. Animat performance is judged based on a brick-stacking task in which animats search for bricks, carry them back to their tribe's tower, and then place them in the correct order. The characteristics of the animats and their world were varied across numerous trials to examine the effects of environment on communication, cooperation, and behavior.

There are four areas I explore using NEC-DAC:

1. Probabilistic reasoning;
2. Social structures;
3. Role differentiation;
4. and some final Miscellaneous variations.

The focus of these experiments is a tower building task that requires animats in a tribe to cooperate with each other to complete several subtasks:

1. Locate bricks;
2. Fetch bricks from the world and bring them to the tower;
3. Stack bricks onto tower foundation;
4. Attack enemy animats to disrupt the building of enemies' tower;
5. Defend against attacks from enemy animats.

Each tribe has one tower, and most trials have two tribes. (Some trials were run "solo" with just one tribe, in which case the tribe is evaluated based on its scoring rate.) The brick-stacking subtask is serial; each brick is numbered from 1 to 5 and a tower can only be built with bricks stacked in the proper order up to a maximum height of five bricks. The other subtasks can be performed in parallel, and a tribe that coordinates its scouting and fetching activities can build its tower more quickly and efficiently than can a tribe that uses straightforward greedy tactics. Animats are rewarded for placing the next required brick on top of their tower, and several things occur when the final (fifth) brick is placed:

- the score for the tribe is incremented,
- the tower height is reset to zero,
- the bricks that made up the tower are scattered randomly in the world.

Building a tower is a task that can be completed more efficiently by agents that communicate, and it is also directly analogous to the activities of many animals and human groups. Beavers build dams, birds build nests, and humans build skyscrapers. Even aside from construction, there are many jobs with a mix of serial and parallel subtasks that can be completed more quickly if agents communicate and share information. A law firm preparing a big case will have dozens of lawyers and paralegals working on different angles, all of which need to be brought together at trial. Communication reduces redundancy and enables work to be distributed among agents, which is exactly what occurs in NEC-DAC. Because finding a mathematically optimal solution to such tasks is impractical, tribe efficiency was measured through competition: tribes with differing characteristics and signals were matched against each another to determine which characteristics had the most effect on performance.

The brain of each animat is a pair of recurrent neural networks (Haykin, 1998; Omlin and Giles, 1994; Klapper-Rybicka, Schraudolph and Schmidhuber, 2001) -- the Action Neural Network controls the animat's behavior, and the Signal Neural Network controls its signal utterances. Each animat maintains four circular history queues, each with 100 entries:

1. The *Action Queue* records the behavioral states the animat selects, which are the output of the Action Neural Network.
2. The *Signal Queue* records the signals the animat utters (which are the output of the Signal Neural Network).
3. The *Input Queue* records sensory input (from sight and knowledge).
4. The *Score Queue* records the animat score, based on performance.

Every time an animat executes its main thinking loop (and chooses a new behavior), these queues are updated and the inputs and decisions are saved in the circular history queues. The Score Queue is used to provide feedback as the Q function for Q-learning (Watkins, 1989) (a form of temporal difference learning (Sutton and Barto, 1998)), and the other queues are used for determining the training eligibility for each past decision. The training eligibility of a state is a measure of how heavily an animat's neural networks will weigh the results of a previous decision when it looks back in time for training purposes; a decision with an eligibility level of zero will not be used for training at all.

(Kirby, 2002) identifies three complex adaptive systems that contribute to communication in living organisms: learning (for yourself), culture (learning from another), and biological evolution. Much animat research focusing on collective behavior takes an evolutionary approach (Ward, Gobet and Kendall, 2001; Jim and Giles, 2000; Reggia, Schulz, Wilkinson and Uriagereka, 2001), but since NEC-DAC is primarily concerned with *culture* a tribe-based learning approach is more appropriate. As such, the animats do not mate and have offspring; rather, each animat is assigned a fixed lifespan, after which it is destroyed and removed from the tribe and a new animat is created with randomized neural networks to take its place. There is "population flow" as described in (Steels, 2003), but there are no "offspring" and no "inheritance". There are no genes passed between generations, but tribes with signals that enable animats to teach each other are able to pass on beneficial behaviors from oldsters to youngsters. In this way, the behavior and signal combinations that constitute a tribe's *culture* are wholly contained in the weights of their neural networks, and wholly transmitted between animats via teaching signals. Other research also tends to focus on how communication arises from situations in which it does not already exist (Cangelosi and Parisi, 2001), but this study of culture instead investigates how signals are used beneficially, in concert, once they have already been created.

(Vogel, 1999) gives a succinct definition of culture that she says is used by many biologists: culture is "any behaviors common to a population that are learned from fellow group members rather than inherited through genes". Under this definition, full language use is not required for a population to possess a culture, and in fact many animals appear to acquire and transmit culture by observation rather than through the use of any signals whatsoever. Unlike humans, Vogel writes that primates are unable to build on previously learned ideas and unable to combine ideas into larger useful constructs. (Horner and Whiten, 2005) describe the difference between chimp and human learning as the difference between *emulation* and *imitation*. (Emulating a process brings about the same end result, whereas imitation copies a process step-by-step.) In their experiment, young wild-born chimps and 3- to 4-year-old humans were set to the task of opening a box after watching an adult human open the box using a combination of useful and superfluous movements.  They first used an opaque box so that the test subjects could not determine which movements were required to open the box and which were not, and both the chimps and the humans imitated all the movements of the demonstrator.  Then they performed the same trials using a clear box so that the subjects could observe which movements were actually required.  The chimps quickly learned to eliminate the unnecessary movements and opened the box in a more efficient manner than the demonstrator -- emulation.  The humans, however, continued to imitate the unnecessary steps they saw the adult perform, even though they could now see into the box and observe that some of the steps were unnecessary.  At first glance it may appear that the chimps performed more "smartly" than the humans because they opened the box without wasted movements, but the real significance of the result is what it tells us about *how* humans learn.  Unlike the chimps, the humans learned not just that the box contained a reward, but they also learned to perform a certain set of movements, some of which were only peripherally related to opening the box.  This is culture, as can attest any person who was forced into a tuxedo for a wedding as a child and later required his own sons to wear tuxedos because *that's just what you wear to a wedding*.  The chimps essentially reinvent the box-opening solution once they know there's a reward inside, but the humans actually learn to do what they see the demonstrator doing.  In sophistication, NEC-DAC's animats lie somewhere between chimpanzees and humans; their signals are nowhere near the complexity of human language, but they are able to learn to combine behaviors into short sequences when taught by their elders and through observing the behavior of their brethren.

## 2.1 Probabilistic reasoning

Typical artificial neural networks are deterministic in that, for a set of weight values, a given input vector (set of inputs) will always yield the same output vector (set of outputs). However, natural neural networks are far more complicated than any artificial network, and even if they are deterministic they are also chaotic to the point that they are unpredictable. To simulate this property, NEC-DAC uses probabilistic neural networks in which the output vector is normalized and used as a set of probabilities to pseudorandomly select a behavior or signal. This evaluation method helps the animats avoid overtraining and enables them to escape local maximums that could otherwise prevent the full exploration of their state space, resulting in probabilistic animats that outperform peers that use winner-takes-all evaluation.

Probabilistic parameters were varied in the PROP-REAS experiment by:

1. *Probabilistic behavior* in which the neural networks output percentages for each behavior or signal and a random value is used to pick a winner.
2. Use of an *output bonus* to boost the probability of selecting a low-likelihood behavior;
3. Use of an *obedience multiplier* to vary the likelihood that a listening animat will obey a command signal given by another animat;
4. Presence or absence of a *random start phase*.

5. Use of *memory registers* that enable animats to store state information through time and pass information between neural networks.

## 2.2 Social Structures

There is no limited to the number of types of signals that can be communicated within a suitably complex environment, but it is not always obvious how useful any given signal will be to a tribe of communicating agents. My experiments focus primarily on three types of signals and evaluate the utility of specific instances of these types:

1. *Command signals* (which are obeyed according to the *authority* of the signaler);
2. *Information signals* (which ignore authority).
3. *Education signals*.

Command signals are issued by an animat with *authority* to a listener that then bases its next behavioral decision on the command received. The listener may not automatically obey the command, depending on the type of probabilistic reasoning used in the specific experiment. The question of which animats have authority over which other animats is determined by the social structure built into each experiment. There are an innumerable number of social structures that can be seen in real life, but in my SOC-STRUCT experiment I dealt with five:

1. *No authority*, in which no animats can command any others;
2. *No hierarchy*, in which every animat can command any animat in its tribe;
3. *Age hierarchy*, in which an animat can command every younger animat in its tribe;

4. *Merit hierarchy*, in which an animat can command every lower-performing animat in its tribe;
5. *Age and merit hierarchy*, in which an animat can command every other animat in its tribe that is both younger and has a lower score.

The first three types of hierarchy are static, in that if animat A has authority over animat B at time *t*, then A also has authority over B at time *t+1*. In the merit-based system, A only has authority over B if A's score is higher than B's score at that time, and in the age and merit system A must also be older than B. In all cases authority is transitive -- if A has authority over B and B has authority over C, then A has authority over C.

Information signals are used to share sensory data among animats in the same tribe. The data that could potentially be shared is constrained only by the contents and complexity of the world, but the signals in these experiments were limited to the following subset:

1. Internal state information of a specific animat;
2. Location of objects in the world;
3. Future intentions of the signaler;
4. Past events;
5. Properties of the world;
6. Properties of a task.

Education signals provide feedback to a commanding animat that indicates whether or not the results of its command were beneficial to the animat that obeyed the command.

By testing these types of signals, NEC-DAC can be used to explore how information is used to facilitate cooperation. My research indicates that more information is not always beneficial, and can even be harmful.

## 2.3 Role differentiation

A *role* is defined by the *American Heritage Dictionary* as "the characteristic and expected social behavior of an individual", and animat simulations are an excellent tool for exploring the "role of roles" within social structures (Odell, Parunak and Fleischer, 2003). For example, how do parents and children know how they are supposed to behave within the family relationship? How do co-workers know how to cooperate? How does a sports team decide who plays in which position? It is intuitive that the age-old nature vs. nurture dichotomy is simplistic and that both genetics and learning play a role in these complex social interactions, but the extent of the influence of each is unclear.

Odell et al. defined two kinds of methods of role assignment (and the terms can be generalized to apply to any type of simulation feature):

- *endogenous*, in which roles emerge as animats self-organize,
- *exogenous*, in which roles are created by the designer of the simulation.

There is nothing about any particular feature or assignment of roles that renders it inherently exogenous because a role built into a simulation could also potentially be generated through evolutionary means. It is only within the context of a particular simulation that the *method* used for role assignment is endogenous or exogenous. For example, NEC-DAC's signals are exogenous because I designed them myself, but a future extension could result in the same signals arising through evolution. That said, given the set of signals I created I wanted to examine the roles that would arise, so in the ROLE-DIFF experiment roles arise endogenously. Every animat in a tribe begins with identical capabilities -- that is, they are "genetically" homogeneous. NEC-DAC is a learning simulation and its animats do not evolve or change in construction or kind over the course of an experiment. Behavioral differences between animats within a tribe will only arise due to changes of neural weights within the brains of the animats, leading the animats to make different decisions. The amalgamation of the decisions that an animat makes is considered to constitute a distinct *role* if its behavior follows a discernable pattern and if the level of differentiation between that animat and the others in its tribe can be measured. The measurement used for this evaluation is the *Differentiation Factor* (DF) calculated for a whole tribe. Two DFs are used: the Action Differentiation Factor (ADF) measures differentiation of behavior sequences and the Signal Differentiation Factor (SDF) measure differentiation of signal sequences.

For *N* behaviors or signals, there are $N^2$ sequences of length 2 comprised of decisions from time *t* and time *t-1*, which we'll call the *time pair* (*t*, *t-1*). Each animat keeps track of how often it performs each of the $N^2$ sequences, and at the end of a trial these counts are normalized to sum to 1, yielding the *sequence frequencies*. The sequence frequencies for each sequence for each animat in a tribe are then used to calculate the sequence frequency standard deviation for that sequence (Equation 1 and Equation 2). The $N^2$ frequency standard deviations are then averaged together to produce the final result: the Differentiation Factor (Equation 3). Table 1 describes the variables used in the equations below.

**Table 1 -- The meanings of the variables used in the Differentiation Factor equations.**

| Variable | Meaning |
|---|---|
| $A$ | Number of animats in a tribe |
| $N$ | Number of behaviors or signals |
| $N^2$ | Number of behavior or signal sequences of length 2 |
| $sf_{a,n}$ | Sequence frequency of sequence $n$ for animat $a$ |
| $msf_n$ | Mean sequence frequency for sequence $n$ for all animats in the tribe |
| $\sigma_n$ | Standard deviation of sequence frequencies for sequence $n$ |
| $DF$ | Differentiation factor |

$$msf_n = \frac{1}{A} \sum_{\forall a} sf_{a,n}$$

**Equation 1**

$$\sigma_n = \sqrt{\frac{1}{A} \sum_{\forall a} (sf_{a,n} - msf_n)^2}$$

**Equation 2**

$$DF = \frac{1}{N^2} \sum_{\forall n} \sigma_n$$

**Equation 3**

Essentially, the DF measures the extent to which the animats in a tribe behave similarly or differently; a higher DF indicates that the animats have more widely varying sequence frequencies, and thus are more differentiated. These differences may not be easily labeled with names that correspond to roles among real animals (such as *father* or *street-sweeper*), but a high DF indicates that differentiation is occurring. Longer sequences could be used to calculate higher-order Differentiation Factors, but sequences of length 2 were found to be sufficient for noticing chains of behaviors of any length. Additionally, since the number of possible sequences increases exponentially proportional to the sequence length, it was impractical to use longer sequences. An animat typically makes approximately 100,000 decisions with each neural network during its life; with 12 possible behaviors, there are 144 sequences of length 2 but 1728 sequences of length 3, making it exponentially more difficult to extract patterns from random noise for longer sequences. However, I investigated length 2 sequences comprised of behaviors from time pairs other than just *t* and *t-1* and put the results in Table 2. The absolute values of the numbers themselves will not mean much right now, but DF values for random behavior is included for comparison and the context will be explained further when the experiments are discussed in section 4. The differentiation measurement decreases as the time separation increases, which indicates less of a connection between behaviors and signals that are far apart in time, as might be expected. The correlation between differentiation factor and score also drops as the time separation increases, which indicates that the increased differentiation doesn't contribute to score and isn't purposeful.

Table 2 -- This table compares the relative values of Action DF (ADF) and Signal DF (SDF) when the differentiation is calculated with four different time pairs. DF and correlation values are also included for random behavior for comparison purposes.

| Time pair | ADF | Correlation of ADF to Score | SDF | Correlation of SDF to Score |
|---|---|---|---|---|
| ($t$, $t-1$) | 38.2 | 0.58 | 12.0 | 0.58 |
| ($t$, $t-2$) | 38.9 | 0.48 | 12.7 | 0.48 |
| ($t$, $t-4$) | 39.5 | 0.35 | 14.1 | 0.48 |
| ($t$, $t-8$) | 42.3 | 0.22 | 16.0 | 0.29 |
| ($t$, $t-1$) (random behavior) | 105.2 | 0.01 | 32.5 | 0.00 |

Each tribe's Action Differentiation Factor (ADF) and Signal Differentiation Factor (SDF) are calculated every 1,000 seconds during a trial and the final values for a tribe are the average values of the DFs collected during the trial. DF fluctuates as animats learn, die, and new animats are created, so taking an average value of many snapshots is the best way to reduce the ADF and SDF information into two stand-alone numbers. In addition, the intermediate values of the ADF and SDF can be plotted to examine more closely how they track the tribe's score, and how they relate to other events within the trial.

# 3  NEC-DAC Architecture

This section has two parts:

1. *World Architecture,* which describes the world that is used as the environment for the simulations;

2. *Animat Architecture,* which explains and justifies the design of the animats and their learning process.

The project was built on the foundation of the "Quake 2" source code, which has been released under the Gnu Public License and can be downloaded from id Software's website at http://www.idsoftware.com/business/home/techdownloads/. The "Quake 2" source was used for its 3D graphics-rendering engine and to handle the mundane details that are typically associated with building a virtual world, such as collision detection.

All programming was done in C, and every object in the world is structure-based and self-contained. Simulation output data is saved to text files and PNG images. Spreadsheets are generated based on the text files, and movies and slideshows can be created from the images.

## 3.1  World Architecture

The NEC-DAC world is a non-wrapping two-dimensional 2048x2048 units square surface. A non-wrapping world eases line-of-sight and range calculations for sight and signaling purposes and increases the distance possible between objects without decreasing density. Time is discrete with 10 ticks per simulation second and objects in the world call "think" functions periodically to govern their operation. The time scale is variable to enable fast execution when desired.  At maximum speed on my Athlon 3GHz home machine the software executes approximately 70 simulation seconds per real second.

The 3D representation of the world was built using the "WorldCraft" software (website: http://www.planetquake.com/worldcraft/index2.shtm) -- a free software toolkit for creating Binary Space Partition (BSP) models.  Models and lighting from the "Quake 2" software are used to represent the animats and their surroundings.  Most simulations are run without using the graphical interface to facilitate faster execution.

Objects in the world consist of:

1. *Animats*, which are DACs.  The animats are non-blocking and cannot obstruct each other except by purposefully initiating combat behaviors -- this decision was made to preclude difficult-to-measure side effects of design that could be unrelated to behavior, and to reduce the sensory burden.  Animats should focus on completing their task and cooperating by coordinating subtasks, not on the physics of obstruction.
2. *Bricks*, which are immobile and block each other, so that only one brick can be in any 64x64 location at a time.  Bricks are numbered and can only be stacked onto a tower in the right order.

The animats scour the world for bricks, bring them back to their tower foundation, and then stack them in the proper order.  An animat is rewarded for placing the correct brick on top of its tribe's tower and for combat against an enemy animat.  The base reward values are 10 points for placing a brick, 1 point for winning a combat, and -1 point for losing a combat.  These base values were chosen because they produce a good mix of combat and building; if the relative reward for combat is increased then the animats attack more, and if the relative reward for brick stacking is increased they focus more on fetching and dropping bricks. Since tower building is the main task in the project and combat is a side complication, I decided to make brick placement more rewarding than combat for most experiments. (For a more detailed description of combat and a justification for the combat reward system, refer to section 3.2.1.)

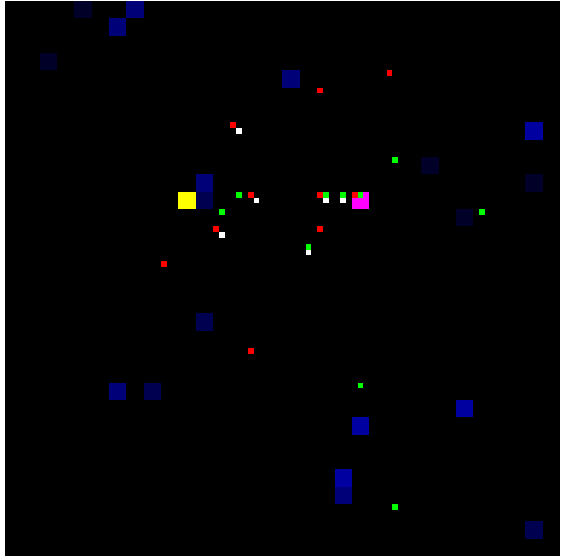

**Figure 1 -- Snapshot of world soon after creation.**

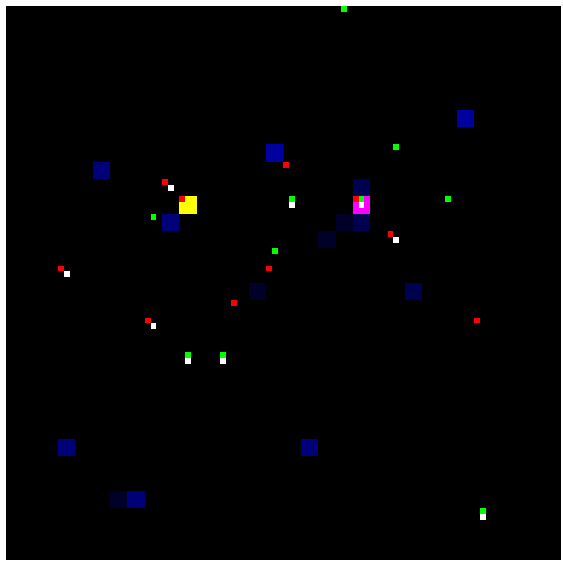

**Figure 2 -- Snapshot of world after random start phase.**

Figure 1 and Figure 2 show snapshots of the world taken near creation and then after the random start phase, respectively. In these figures we can see:

- Red and green dots are animats from tribes 1 and 2, respectively

- The large yellow and purpose squares are towers for tribes 1 and 2
- Large blue squares are bricks, with shade indicating cardinality
- Small white dots indicate recent combat locations
- Animats can be seen fighting primarily in the territory between the towers
- Animats around the periphery are searching for bricks.

Although a larger variety of objects would increase the complexity of the world, I believe that the brick-stacking game as described is sufficient for exploring the use of signals in coordinating parallel and serial tasks among distributed agents. Adding, for instance, walls that blocked movement and obscured vision would yield a more complex simulation world, but that new layer of complexity would not necessarily yield new brick stacking signal combinations; instead, useful new signal combinations would simply focus on completing yet another subtask: that of navigating walls.

In designing the world and the animats, three interrelated parameter values were chosen arbitrarily: the size of the world, the velocity of the animats, and the length of the animat think cycle. These three parameters together determine how far an animat can move between each decision it makes -- both in absolute terms and in proportion to the world -- and the values can only be considered relative to each other. The values chosen are shown in Table 3.

**Table 3 -- World parameter values.**

| Parameter | Value |
|---|---|
| Size of world | 2048 units on each side |
| Animat velocity | 23 units per second, plus or minus 3 |
| Animat think cycle | 2.1 seconds, plus or minus 1 |

The random values are chosen using a function called *crandom()* which returns a double wide real value in the interval [-1,1) using the algorithm described in (Matsumoto, 2004). (Early experimentation demonstrated that the default C pseudorandom number generator was not sufficiently random and led to behavioral "bunching" artifacts, so Matsumoto's Mersenne Twister algorithm was used instead.) The goal in selecting the values for these parameters was to make the world spacious in the sense that the animats would have the time to execute many think cycles during a complete traversal -- in this case, an animat will make an average of 42 decisions while crossing the world from one side to the other. Design experiments showed that these values do not significantly affect the behavior of the animats as long as they remain within the same order of magnitude.

Although there are only two types of objects in the world (bricks and animats), they interact in complex ways such that interesting behavior arises. Because bricks come in various numbered sizes, all of which are essential for building a tower, it is important for animats to tell the difference between the bricks and act accordingly; in the ROLE-DIFF experiment (section 4.3) animats within a tribe learn to divide up the brick-fetching task such that each animat specializes in bricks of a certain size. The presence of two tribes in a trial adds complexity as the opponents vie for scarce resources and compete directly (through combat) and indirectly (through brick fetching). The SOC-STRUCT experiment includes a comparison of behavior between trials with one tribe and trials with two tribes. Furthermore, the design of the system is simple enough that additional objects can be incorporated to accommodate future experimentation.

## 3.2 Animat Architecture

The animats were designed to have behaviors and signals added and removed as desired, and the following explanation refers to functionality that describes how standard behaviors and signals work. Exceptions will be noted in instances where certain behaviors and signals require special handling.

Each animat is controlled by a pair of recurrent neural networks:

1. *Action Neural Network*, controls the behavior of the animat;
2. *Signal Neural Network*, controls the signaling of the animat.

Neural networks can handle large, fuzzy state spaces and continuous input vectors more efficiently than other potential architectures, such as look-up tables. Neural networks are also well suited to adaptation as probabilistic decision-makers because they have many scalable outputs that can be converted to probabilities. Training is performed using a modified form of Q-learning with a hyperbolic tangent activation function (Sutton and Barto, 1998). Though some (Oliphant, 1997) have advocated the use of observational learning rather than reinforcement learning for learning communication, others (Steel and McIntyre, 1997; Hutchins and Hazelhurst, 1995) have used reinforcement learning to great success for soliciting the emergence of communication and I decided to follow in their footsteps due to the difficulty observational learning methods have bootstrapping.

The animats are physically instantiated as 3D objects and able to move around in the continuous 2D world. Animats have several attributes that determine their physical interaction with the world around them, including:

1. Volume and opacity, so they can be "seen" by other animats.

2. Velocity -- animats can move towards any target object in any direction; speed is standardized around 23 units per second, plus or minus 3. The small random variation prevents the animats from getting stuck on an implicit grid (for example, if speed were always exactly 23 units per second then animats would end every second at a location that is a multiple of 23). Animats cannot directly control their movement speed, though behavior procedures could be designed to use different speeds.
3. Health, which serves as a score to be punished and rewarded based on the animat's actions.
4. Behavioral state, which determines how the animat will act until the next time it thinks.

The wait times between think phases and the velocity ranges of the animats were initially chosen arbitrarily and then tweaked so that enough interaction would occur in each simulation that the animats would be able to reach an equilibrium state through learning. Small random variations are used to spread out animat activity across simulation ticks, preventing animats from thinking and moving all within the same tick and thereby providing a more asynchronous environment.

The following subsections describe four components of the animat architecture:

1. *Behaviors*;
2. *Signals*;
3. *Senses and Internal knowledge*;
4. *Thinking -- the main program loop.*

### 3.2.1 Behaviors

The behaviors available to the animats are procedures that enable the animat's neural networks to make abstract, "high level" decisions without having to control all of the underlying details; the outputs of a neural network indicate which behavior to invoke. For example, if the Action Neural Network decides to attack an enemy it doesn't have to manipulate dozens of independent muscles to move the animat into position and then strike a blow -- it merely initiates the Attack behavior and the details are executed procedurally by the Attack subroutine.

One disadvantage of this approach is that an animat can't fine-tune its behaviors to the degree it could if it had precise "muscular" control. However, if an animat could tweak the details of its attack effector in the same way a real animal can, the problem space would be enormous. By abstracting and encapsulating behaviors into procedural subroutines, the animats can focus on learning when and how to use their predetermined abilities.

Behavior and signal selection is performed using a probabilistic method in which the outputs from a neural network are normalized and used as a percentage value. A pseudorandom number generator is then used to select the behavior or signal to be executed. Thus, the networks do not have absolute control over the animat, they only set the probability that each behavior or signal will be chosen.

Animats do not specify their actions based on coordinate values, but once they have observed an object they can use its location implicitly. Neural networks cannot easily generate raw coordinates as outputs, so coordinates are pulled from a knowledge spatial map once the neural network decides what object it is interested in. The animats have the following types of behaviors available to them, though not all are enabled in every experiment.

1. **Go to friendly tower.** The animat goes to its tribe's tower.

2. **Go to enemy tower.** The animat goes to the enemy tower.

3. **Fetch brick +1/+2/+3/+4/+5.** If the animat knows the location of a brick numbered X higher than the one it believes to be on the top of the tower then it moves towards that brick and picks it up. Otherwise, the animat moves randomly until it sees such a brick.

4. **Fetch brick-1/2/3/4/5.** If the animat knows the location of a brick numbered exactly X then it moves towards that brick and picks it up. Otherwise, the animat moves randomly until it sees such a brick.

5. **Drop brick.** The animat drops any brick it is carrying at its current location. If the animat is at its friendly tower and the brick is the right number, the brick is put onto the tower.

6. **Explore**. The animat moves towards a random location in the world.

7. **Attack targeted enemy.** The animat attacks the enemy animat it has targeted, causing both to drop any bricks being carried. The attacker's combat power (described in 3.2.3) is set to 0 to prevent an immediate repeat attack from being successful. The animat that loses is stunned and unable to think for 5 seconds. This stun period imposes a time cost on the loser by preventing it from acting (and therefore scoring).

8. **Defend.** The animat stands still for a certain amount of time and waits to be attacked. After it is attacked, the animat stops waiting and chooses another behavior.

9. **Wait.** The animat waits immobile and does nothing; it also regenerates an additional 0.1 points of combat power.

Each of these behaviors persists until it completes (as described in Table 4) or the animat gives up. For instance, if an animat decides to fetch brick-1, it will move towards the nearest brick-1 it knows of, or begin exploring in search of a brick-1 if there aren't any known. Every 2.1 seconds it will check to see if it has successfully picked up a brick-1, and if it hasn't completed the behavior there is a 5% chance that it will give up. This "give up" decision is not made by the neural networks, but rather treated as a part of the behavior procedures. The 5% value was chosen arbitrarily to enable an animat to escape from a behavior that it cannot complete; a larger value would cause animats to abandon unsuccessful behaviors sooner and would limit, e.g., the area an animat could scour for a brick of a desired size.

Table 4 -- Behaviors and their completion events.

| Behavior | Completion Event |
|---|---|
| Return to tower | Completes when the animat is within 32 units of the tower that belongs to its tribe. |
| Go to enemy tower | Completes when the animat is within 32 units of the enemy tower. |
| Fetch brick +X | Completes when the animat has picked up a brick that is number X higher than the brick it thinks is on the top of its tower. |
| Fetch brick-X | Completes when the animat has picked up a brick that is numbered exactly X. |
| Drop brick | Completes when the animat has dropped any brick that it might be carrying. |
| Explore | Completes when the animat reaches the random location it chose to explore. |
| Attack targeted enemy | Completes when the animat executes its attack effector against its targeted enemy (i.e., the enemy indicated by the animat's enemy target variable). |
| Defend | Completes when the animat is attacked by an enemy. (Note: the defend behavior can also be terminated by giving up if the animat is never attacked.) |
| Wait | Completes when the animat gives up. |

The Attack behavior warrants further description. When animat A attacks enemy animat B, the following sequence of events occurs. The value $a\_str$ represents the attack strength of the attacker, A, and the value $b\_str$ represents the attack strength of the defender, B; higher attack strengths make an animat more likely to win the combat.

1. The distance between A and B is checked to make sure the animats are within sight range.

2. The Just Attacked variable of B is set to 1.

3. The Enemy Target variable of A is set to B, and the Enemy Target variable of B is set to A.

4. $a\_str \leftarrow$ Combat Power variable of A.

5. If A is not carrying a brick, a_str ← a_str + 1.

6. b_str ← Combat Power variable of B.

7. If B is not carrying a brick, b_str ← b_str + 1.

8. If B is within sight range of its tribe's tower and B is in the Defend state, b_str ← b_str + 1.

9. a_str and b_str are compared, and whichever animat has the highest strength is the winner of the combat.

10. The winner:

    a. gains 1 health (a reward of 1).

11. The loser:

    a. loses 1 health (a reward of -1);

    b. has its think delay increased by 5 seconds;

    c. is pushed in the direction away from the winner, causing it to move away until it thinks again and can change its direction.

I considered rewarding animats indirectly for combat rather than directly and ran a few experiments in this direction, but the results were not as interesting. For example, I tested a system in which animats received no reward for winning a combat other than that the loser was "stunned" (unable to act) for a length of time (5, 20, or 50 seconds) and was forced to drop any brick it was carrying. The losing animat was punished and the winner received some indirect benefits:

- the winner could pick up the brick dropped by the loser,
- the winner and its tribe faced less competition for bricks while the loser was stunned,

- the winner and its tribe were less likely to be attacked while the loser was stunned.

However, with only these indirect benefits to be reaped from combat the majority of animats found it more profitable to fetch and return bricks than to risk losing a fight, and so combat was comparatively rare. I wanted to increase the value of winning enough to encourage a critical mass of combat such that individual animats would choose to specialize in attack and defense behaviors, so I decided to add a direct reward for winning combat. In the final system, animats receive a direct reward for winning a combat in addition to the indirect tribal benefits listed above, and adding this reward had the desired effect.

### 3.2.2 Signals

Along with their behavioral abilities, the animats' signaling suite is the heart of the project and contains the building blocks the animats use to construct their cooperation schemes. In addition to performing the action selected by its Action Neural Network, an animat utters a signal selected by its Signal Neural Network in each iteration of its think cycle. When an animat utters a signal it is instantly heard by every animat within range, though only the targeted friendly animat may act on the signal if it is a command. For the sake of simplicity, the nearest friendly animat is always the one targeted; future work may be done to add a neural signal target selection mechanism. Whether or not the targeted animat obeys the signal (if it's a command signal) is determined by the listener's neural network and the tribe's social structure (see SOC-STRUCT, section 4.2). Animats can also decide to send a Null signal, in which case nothing happens.

Each signal is implemented procedurally (symbolically), similar to the way behavioral subroutines are handled as described in section 3.2.1. The signals are not created by the neural network, but rather the neural network selects which signal to use from the available set. The animats don't learn *how* to signal, they learn *how to effectively use* the signals they are capable of uttering and understanding. As described in the following subsections, each signal enables the signaler to communicate with its peers in a specific way, and the animats learn to combine these signals into a cohesive cooperation scheme. For example, animat A can command animat B to fetch brick-1 and then soon after animat A can command animat C to fetch brick-2, thereby accelerating the rate at which the tower is built by distributing the fetching subtasks and ensuring that animats B and C don't both fetch the same brick. This kind of sequencing also contributes to the role differentiation measured in the ROLE-DIFF experiment.

Not all signals are used in every experiment, and the specific range of capabilities available to each animat and tribe vary according to the scenario. There are four types of signals:

1. *Command signals* enable the speaker to influence the behavior the listening animat will select next time the listener makes a decision;
2. *Information signals* enable animats to exchange world knowledge;
3. *Education signals* are used by animats to teach each other and to provide feedback to teachers;
4. *Null signal*.

There is one command signal for each behavior specified in section 3.2.1. It is important that command signals overlap the behaviors that are otherwise possible because if hearing a signal enabled an animat to undertake a behavior that it couldn't perform on its own then it wouldn't be clear whether or not any advantage gained was due to the intelligent use of the signal or to the intrinsic nature of the new ability. By coordinating behaviors and command signals it is possible for animats to synchronize behaviors for mutual gain, e.g., for an animat to attack an enemy and simultaneously command another friendly animat to attack at the same time (although in NEC-DAC there is no mechanism for the commander to tell the listener *which* animat to attack at this time). By attacking together the attackers ensure that at most one of them will face an enemy with full combat power and the other(s) will be able to subsequently attack with a much greater chance of winning.

There are three classes of information signals, and each class represents a piece of information that can be shared; each class contains several distinct signals that behave similarly. Note that the neural network does not generate parameters to pass to these signals; the objects referred to by the signals, such as bricks or towers, are determined by the animat's senses and internal state.

1. **Tower height.** The signaler tells its listeners the height of the friendly or enemy tower and when the signaler saw it last. The listener stores this information in it's the appropriate internal state variable if it is more recent than the listener's current knowledge. Suppose animat A saw the enemy tower was height 2 at time 5 and animat B saw the enemy tower was height 1 at time 3. When A uses a tower height signal to communicate with B, B will update its internal knowledge to reflect A's information since it is more recent.

2. **Brick +1/+2/+3/+X location.** This signal tells the listeners the coordinate location of a known brick that is numbered X higher than the brick believed to be at the top of the tower by the signaler (based on the signaler's knowledge of the tower's state). The listener then stores this new knowledge in its spatial map. The specific brick size is transmitted and the animats do not have to be in agreement as to the height of the tower.

3. **Brick 1/2/3/X location.** This signal tells the listeners the coordinate location of a known brick that is numbered exactly X. As above, this brick location is incorporated into the listener's spatial map. Signals for both exact and relative brick sizes were used in order test the effect on signal sequencing. For instance, with relative signals an animat could signal "+1" every time and be useful, but with exact signals the animats would have to learn to use the signal for each specific size.

There are two education signals that are used for two different styles of animat education.

1. The **Teach/Learn** signal is used to directly share experiences between animats. When one animat utters TL and another hears it, the animat with the higher score (the "teacher") copies its Input, Action, Signal, and Score Queues to the "learner". The learner executes its training algorithm using the copied queues and then restores its own history data (returning it to its original state). This process is comparable to the teacher explaining its history and recounting the results of its behavior and signaling decisions to the learner. (The teacher is the animat with the higher score, regardless of which animat initiated the exchange.) This signal was not used in any of the final experiments.

2. The **Feedback** signal is used by an animat (while it is training its neural networks) to give feedback about a behavior it was commanded to perform. The animat (which was previously the recipient of a command) tells all friendly animats nearby (possibly, but not necessarily, including the animat who ordered the behavior) whether or not the command was beneficial. The animats that hear the feedback signal then use this information to train their Signal Neural Networks.

Additionally, there is a **Null** signal that Goes Nowhere and Does Nothing. This signal is used as a control for statistical analysis, and should have no effect on the success of any animat or tribe. Every animat is capable of generating this signal, and its frequency represents the value of uttering no signal at all. By using Null, an animat can avoid signaling and thus avoid the unwanted attention from enemies that might otherwise be attracted.

### 3.2.3 Senses and Internal Knowledge

Discussion of sensory apparatus and internal knowledge go together because they are closely related: each bit of information that the animats can sense has to be stored in an internal state variable or spatial map, so there is a one-to-one correspondence. Table 6 lists the types of knowledge that each animat possesses, along with its sensory source. There are three sensory sources:

1. *Sight*, which is omni-directional, has a range of 256 units (the *sight range*).

2. *Hearing*, which is omni-directional, has a range of 512 units (the *signal range*).
3. *Internal*, which has no range and simply represents knowledge that the animat has about itself.

Both sight and hearing ranges were varied in experiments, but these are the baseline values. Neither sight nor hearing is blocked by objects, nor does their accuracy attenuate with distance. They are dealt with in this simple manner because the size of the world enables distance alone to separate animats and bricks from each other without the need for other boundaries. Others (Jim and Giles, 2000) have used a message board approach to signaling in which animats post messages for each other, and this would be similar to using NEC-DAC with an infinite signal range. They argue that communication can help most significantly in partially observable environments, so they decide to limit the sight of their animats but allow their signals to be heard without a restriction on distance. Such a signaling system would reduce the performance limitations of geographic-based simulations such as NEC-DAC, however, and limit the flexibility of the system to incorporate modifications such as signal interception by enemy animats (section 4.4.3), so I decided to limit both my animats' sight and hearing.

The signal range value was determined using the density threshold specified by (Reggia, Schulz, Wilkinson and Uriagereka, 2001), namely that in their evolutionary experiment communication readily emerged when the expected average distance between animats was less than their communication range. Table 5 shows a comparison of the relevant areas for NEC-DAC. I chose these ranges because I wanted an average of at least one friendly animat within signal range of every other animat at any given time, and with 10 animats per tribe that means that the signal range should cover at least 10% of the world. I decided to bump this number up to 20% because much of the coverage can be wasted when animats are near the edge of the world.

**Table 5 -- A comparison of sight, hearing, and world areas.**

|  | Area | Comparison |
|---|---|---|
| World | 4,194,304 square units |  |
| Area within sight range | 205,887 square units (max) | ~5% of world area (max) |
| Area within hearing range | 823,549 square units (max) | ~20% of world area (max) |

**Table 6 -- This table lists every type of knowledge an animat can have and how it is sensed. Not all are used in every experiment.**

| Knowledge | Sense | Description |
|---|---|---|
| Health | Internal | The animat's current score. |
| Behavior | Internal | The animat's current behavioral state. |
| Destination | Internal | The world coordinates the animat is moving towards. |
| Enemy target | Sight | The enemy animat targeted for attack. |
| Enemy target combat power | Sight | The combat power of the enemy target. |
| Signal target | Sight | The selected animat that will be the recipient of an uttered signal (only used in experiments in which signaling is targeted rather than broadcast). |
| Signal target behavior | Sight | The behavior chosen by the animat currently indicated by the signal target variable. This enables the animat to see what its signal target is doing so it can choose an appropriate signal. |
| Think count | Internal | The number of times the animat has made behavior and signal decisions; directly proportional to the animat's age. |
| Memory registers | Internal | 10 registers that are used to hold values taken directly from the outputs of the neural networks, 5 from the ANN and 5 from the SNN. |
| Friendly tower height | Sight/Hearing | The height of the animat's tower when it was last seen by the animat, or the height relayed to the animat via a signal from an animat which had seen the friendly tower more recently. |

| Knowledge | Sense | Description |
| --- | --- | --- |
| Friendly tower observation time | Sight/Hearing | A time stamp of the freshness of the friendly tower height variable above that indicates when the observation was made. This time stamp is compared with the time stamps of signaling animats that attempt to share information about tower height and is used to ensure that only the most recent information is used. |
| Enemy tower height | Sight/Hearing | The height of the enemy tower when it was last seen by the animat, or the height relayed to the animat via a signal from an animat which had seen the enemy tower more recently. |
| Enemy tower observation time | Sight/Hearing | A time stamp of the freshness of the enemy tower height variable above that indicates when the observation was made. This time stamp is compared with the time stamps of signaling animats that attempt to share information about tower height and is used to ensure that only the most recent information is used. |
| Birth time | Internal | A time stamp to record the time at which the animat was created. This variable is used in experiments in which animats are created and destroyed over time. |

| Knowledge | Sense | Description |
|---|---|---|
| Combat power | Sight | A value that is reset to 0 every time an animat initiates an attack. Combat power is increased by 0.1 every think cycle that the animat doesn't attack, up to a maximum value of 1. This trait is used in determining the winner of a combat, and prevents constant attacking from being a successful strategy by simulating the energy cost of an attack. It also allows an animat to judge whether or not an attack is likely to be successful. |
| Just attacked | Internal | A value that is set to 1 when the animat is attacked by an enemy, and then decreases by 0.2 every think cycle, to a minimum of 0. |
| Same signal target as last | Sight | A binary value that indicates whether the current signal target is the same as the signal target last think cycle. This enables the animat to know if it is sending a signal to the same listener. |

Animats use a spatial map system originally based on (Panangadan and Dyer, 2001) to store the locations of bricks within the world. Panangadan and Dyer's egocentric spatial maps (ESMs) use a two-dimensional array of neurons arranged in a 100 x 100 grid to feed geographic brick coordinates, centered on the animat, directly into their animats' neural networks. Their animats then attempt to satisfy their goals of eating, drinking, and placing blocks in pre-designed two-dimensional patterns based on the activation patterns in these maps. Like the animats in Panangadan and Dyer's simulation, NEC-DAC animats are concerned with moving bricks around, but towards a simpler end. Their animats used blocks to build two-dimensional patterns, so it was critical for them to be able to assess block locations using their neural networks. In contrast, my brick-stacking task only requires my animats to deposit bricks at one pre-determined location (albeit in a specified order) and the focus is on the communication signals, not the building itself. In earlier versions of NEC-DAC I experimented with a neural-based spatial map approach based on Panangadan and Dyer's ESMs because I intended to make my brick location signals neural rather than symbolic, but since NEC-DAC animats don't need to drop their bricks in complex patterns it appeared not to be worth the added complexity.

My much-simplified spatial maps, then, are two-dimensional arrays of 32 x 32 *sectors*, each of which represents a 64 x 64 unit square in the simulation world. When an animat sees a brick, it stores a record of it in the corresponding sector of the spatial map, and the knowledge of that brick location decays over time. For example, when animat A is created its brick spatial map will be set to all zeros. As A is traveling around the world it sees a brick-4 located in sector (5,3), so it makes an entry in the (5,3) cell of its brick spatial map to indicate this bit of knowledge and assigns it a recency value of 1000. As A continues to wander, every time it executes a think cycle it will reduce the recency value of the knowledge in its brick spatial map by *random()\*5* units, meaning that it will forget about a brick approximately 400 think cycles after it is last seen. Later, when A decides to fetch a brick-4 it will consult its brick spatial map to determine the location of the nearest one it knows about, and if it hears a signal from a friendly animat asking for the location of a brick-4 it will share its knowledge with its compatriot. Thus, my spatial maps are used symbolically rather than neurally, although future work towards implementing neural-based signals could make use of more proper ESMs. Figure 3 shows a sample spatial map from the beginning of a trial, with most of the bricks randomly distributed.

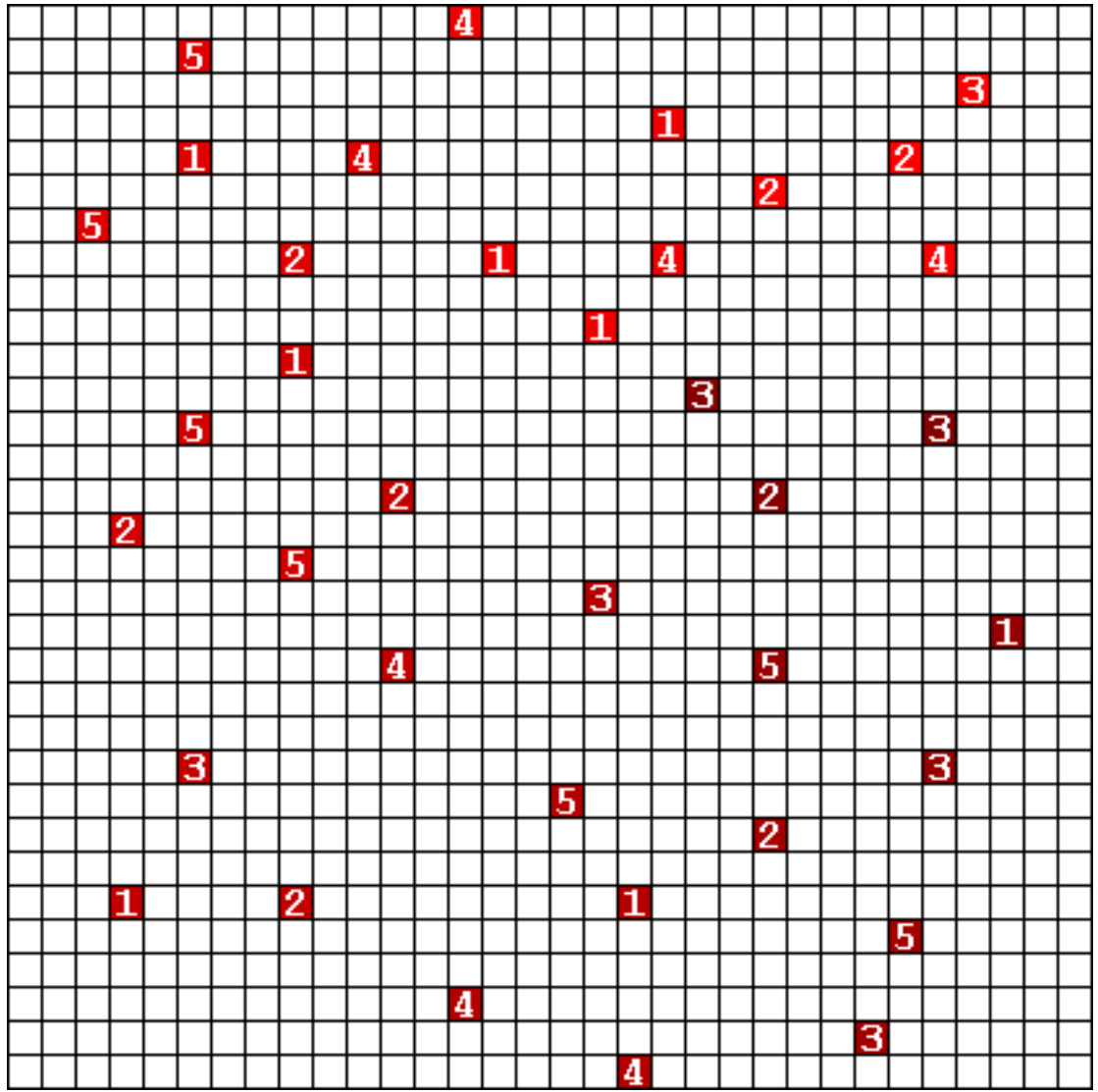

**Figure 3 -- This figure shows sample 32 by 32 sector spatial map with randomly distributed bricks. The numbers in the colored sectors indicate the brick number located in that sector, and the shade of**

**red indicates the recency value of the knowledge represented, with brighter shades meaning more recent.**

### 3.2.4 Neural Network Architecture

Each animat has two neural networks that match with two of the three systems identified by (Kirby, 2002): the Action Neural Network enables the animat to learn for itself, and the Signal Neural Network enables the animat to learn from (and contribute to) its tribe's culture. Their inputs and internal structures are identical but they have different outputs. The neural networks are standard recurrent networks with one hidden layer. The number of units in the hidden layer was varied over time until further increase yielded no advantage in performance; unless otherwise noted, all experiments were run with a number of hidden units equal to *((# of inputs)+(# of outputs))\*(2/3)*. There were typically 66 inputs, 23 outputs, and therefore 59 hidden units. The inputs and outputs are enumerated and described below in Table 8, Table 9, and Table 10.

Two activation functions were tested: the hyperbolic tangent and the sigmoid. I eventually selected the hyperbolic tangent function because it is balanced around zero and results in quicker and more accurate training. The sigmoid function was effective as well, and either one would have been sufficient.

Additionally, in some experiments persistent *memory registers* were used to enable an animat to store data for use in a later think cycle. The memory registers are an array of 10 floating-point values, 5 of which come from dedicated outputs from the ANN and 5 of which come from dedicated outputs from the SNN. All 10 values in the memory register are then used as inputs for each of the neural networks. (While designing the system I experimented with 2, 6, 10, and 20 registers and found that 10 worked as well as 20 and better than 2 or 6, so I stuck with 10.) Because both neural networks get all 10 values as input, the memory registers serve as a simulated *corpus callosum* that connects the two halves of an animat's brain, similar to the connection between the hemispheres of a human brain. The values in the registers don't have any imposed meaning and are not used for anything other than feeding back into the networks. Thus, the networks are free to store there whatever values they want, and the meanings of the registers are entirely dependent on how each network in each animat interprets them. The PROB-REAS experiments demonstrate that animats with memory registers significantly outperform those without, indicating that the networks are indeed making use of the registers even though it is essentially impossible to determine what meanings are being encoded -- and indeed there's no reason to believe that any two animats treat the memory registers similarly. Data from the experiments show that the volatility of the memory registers decreases over time so that a given register assumes a near-constant value that drifts slowly across many learning sessions. However, most high-performing animats do appear to associate specific memory registers with specific behaviors. (See section 4.1 for more analysis of memory register contents.)

Figure 4 illustrates the structure of the two neural networks and Table 7 defines the abbreviations used in the diagram.

Table 7 -- This table describes the designators used in Figure 4.

| Designator | Meaning |
|---|---|
| I1 | Inputs from the animat's internal knowledge variables (Table 6) |
| I2 | Inputs that indicate the behavior being performed by the animat's signal target. |
| I3 | Inputs that indicate the behavior the animat performed last time it made a decision. |
| I4 | Inputs that indicate the signal the animat uttered last time it made a decision. |
| I5 | Inputs that take the values of the memory registers |
| H | Hidden units |
| O1 | Outputs that indicate the behavior to be performed |
| O2 | Outputs that set the ANN's half of the memory registers |
| O3 | Outputs that indicate the signal to be uttered |
| O4 | Outputs that set the SNN's half of the memory registers |
| M1 | The five memory registers set by the ANN |
| M2 | The five memory registers set by the SNN |

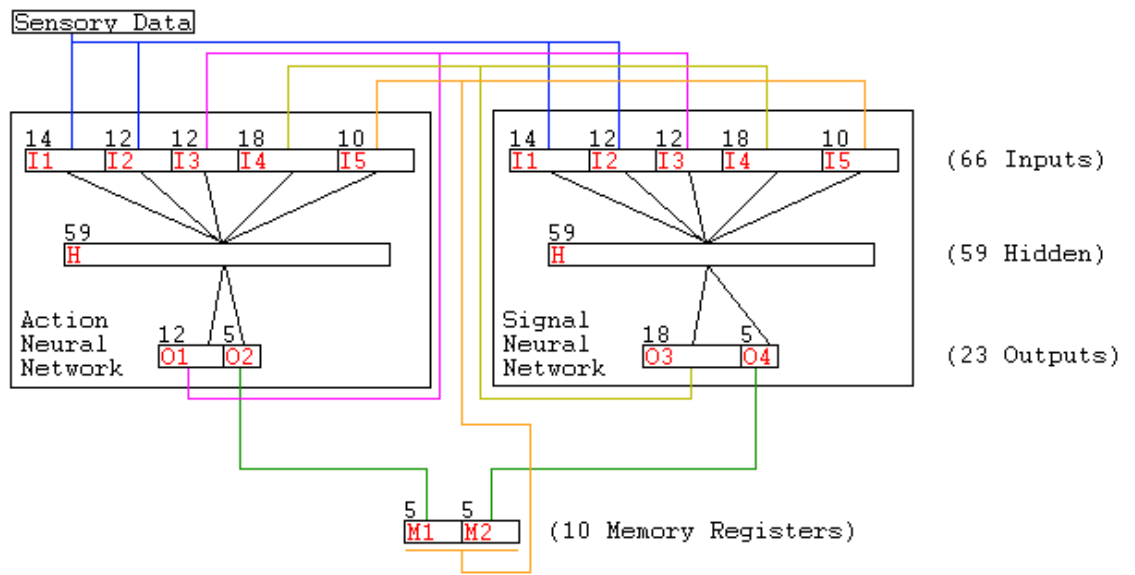

Figure 4 -- ANN and SNN inputs, outputs, and recurrent feedback loops. The numbers of input, hidden, and output units are typical for most experiments. The labels are explained in Table 7.

Figure 4 shows the architecture of the two neural networks and how they are connected. The inputs are located along the top of the ANN and the SNN and take in data from the animat's sensors, internal knowledge, and neural network outputs as shown by the colored connecting lines. Activation flows from the input layer to the hidden layer, and from the hidden layer to the output layer. The colors indicate:

- Blue: input from the environment and the animat's internal knowledge
- Purple: output from the ANN that is fed back into the ANN and the SNN
- Yellow: output from the SNN that is fed back into the ANN and the SNN
- Green: output from the ANN and SNN that is stored in the memory registers
- Orange: values from the memory registers that are fed into the ANN and SNN

When an animat is initialized, each neuron is fully connected to all the neurons in the layers above and below it. The connection weights are set randomly to *(crandom() / sf)*, where *sf* is a scaling factor equal to the number of neurons in the largest neural layer -- the hidden layer. This scaling is important because the sum of the inputs needs to fall within the useful range of the activation function, otherwise the activation of the neuron will always fall on either the high or low boundary.

The two neural networks use the same inputs but produce different outputs, and these inputs and outputs are varied depending on the experiment. Table 8 lists all the inputs ever used, though not all were used in any single experiment. All inputs are initially between 0 and 1 -- inclusive, because many inputs are binary -- before being fed into the neural network, where they are scaled to the range of -1 to 1. (This scaling is required by the hyperbolic tangent activation function. Zero is the middle value of its domain, and a negative value just means that the input is below the middle.  The value 0 will scale to -1, the value 0.5 will scale to 0, and the value 1 will scale to 1.)

Table 8 -- The inputs to both neural networks, with brief operational descriptions; for a more detailed explanation of the meaning of the inputs refer to Table 6.  † Can vary, depending on the number of bricks required to complete a tower.  ‡ Can vary, depending on the number of behaviors and signals available in the given trial.

| Number of Inputs | Bit-Width of Each Input | Input Name | Input Description |
|---|---|---|---|
| 1 | 1 | At friendly tower | 1 when at friendly tower, 0 otherwise. |
| 1 | 1 | See friendly tower | 1 when friendly tower is within sight range, 0 otherwise. |
| 1 | 1 | At enemy tower | 1 when at enemy tower, 0 otherwise. |
| 1 | 1 | See enemy tower | 1 when enemy tower is within sight range, 0 otherwise. |
| 1 | 1 | Have target animat | 1 when an enemy animat has been targeted, 0 otherwise. |
| 1 | 1 | Target has brick | 1 when an enemy animat has been targeted, and that animat is carrying a brick, 0 otherwise. |
| 1 | 1 | Have brick | 1 when carrying a brick, 0 otherwise. |
| 1 | 1 | Have brick, right | 1 when carrying a brick and that brick is the right number to fit onto the tower to the best of the animat's knowledge, 0 otherwise. |
| 1 | 1 | Have brick, too low | 1 when carrying a brick and that brick is too low to fit onto the tower to the best of the animat's knowledge, 0 otherwise. |
| 1 | 1 | Have brick, too high | 1 when carrying a brick and that brick is too high to fit into the tower to the best of the animat's knowledge, 0 otherwise. |
| 1 | 1 | Near brick | 1 when within pickup range of a free brick, 0 otherwise. |

| Number of Inputs | Bit-Width of Each Input | Input Name | Input Description |
|---|---|---|---|
| 5† | 1 | Know a brick size X | One input for each brick size used in an experiment. 1 if the animat knows the location of a brick of the appropriate size, 0 otherwise. |
| 1 | 4 | Friendly tower observed | Set to 0 when the friendly tower is within sight range, slowly climbs to 1 over time while the friendly tower is not in sight. |
| 1 | 4† | Friendly tower height | Scaled from 0 to 1 based on the observed height of the friendly tower. |
| 1 | 4 | Enemy tower observed | Set to 0 when the enemy tower is within sight range, slowly climbs to 1 over time while the enemy tower is not in sight. |
| 1 | 4† | Enemy tower height | Scaled from 0 to 1 based on the observed height of the enemy tower. |
| 1 | 1 | See friend | 1 when a friendly animat is within sight range, 0 otherwise. |
| 1 | 1 | See enemy | 1 when an enemy animat is within sight range, 0 otherwise. |
| 1 | 4 | Combat power | Between 0 and 1, based on the combat power of the animat. |
| 1 | 4 | Target combat power | Between 0 and 1, based on the combat power of the animat's target; 0 if the animat has no target. |
| 1 | 3 | Just attacked | Set to 1 when the animat is attacked then slowly decreases to 0 until the animat is attacked again. |

| Number of Inputs | Bit-Width of Each Input | Input Name | Input Description |
|---|---|---|---|
| 1 | 1 | Same signal target | 1 if the friendly animat targeted for signaling is the same as the last animat signaled to, 0 otherwise. |
| 12‡ | 1 | Signal target observed state | One input for each possible behavior. An input is set to 1 if the friendly animat targeted for signaling is using that behavior at the moment of observation, 0 otherwise. |
| 12‡ | 1 | Last behavior executed | One input for each possible behavior. An input is set to 1 if the animat selected this behavior last time it did a think operation, 0 otherwise. These values are taken from the Action Queue. |
| 18‡ | 1 | Last signal uttered | One input for each possible signal. An input is set to 1 if the animat uttered this signal last time it did a think operation, 0 otherwise. These values are taken from the Signal Queue. |
| 10 | 32 | Memory registers | These inputs represent the values stored in the 10 floating point memory registers. |

Each neural network has a different set of outputs that correspond to the behaviors or signals available in a given experiment in addition to the 5 outputs dedicated to its 5 memory registers (making up the total of 10 memory registers for both neural networks together).

Table 9 shows the outputs of the Action Neural Network, and Table 10 shows the outputs of the Signal Neural Network.  Inside the networks all activation values are between -1 and 1, and the final outputs are then scaled to the range between 0 and 1.  Behavior and signal selection is performed using a probabilistic method in which the outputs from a neural network are normalized and used as a percentage value.  A pseudorandom number generator is then used to select the behavior or signal to be executed.  Thus, the networks do not have absolute control over the animat, they only set the probability that each behavior or signal will be chosen.  The outputs from the ANN are saved in the Action Queue after they are used, and the outputs from the SNN are saved in the Signal Queue.

**Table 9 -- Action neural network outputs.**

| Number of Outputs | Bits per Output | Output Name | Output Description |
|---|---|---|---|
| Variable | 1 | Behavioral outputs | One output for each possible behavior.  An output is set to 1 for the behavior selected for execution, and 0 for the rest. |
| 5 | 32 | Memory registers | The values of these 5 outputs are fed into 5 of the 10 floating point memory registers. |

Table 10 -- Signal neural network outputs.

| Number of Outputs | Bits per Output | Output Name | Output Description |
|---|---|---|---|
| Variable | 1 | Signal outputs | One output for each possible signal. An output is set to 1 for the signal selected for utterance, and 0 for the rest. |
| 5 | 32 | Memory registers | The values of these 5 outputs are fed into 5 of the 10 floating point memory registers. |

### 3.2.5 Thinking -- The Main Program Loop

Each animat has a think routine that serves as the main program loop responsible for controlling that animat's behavior. The think routine of each animat is invoked every *(crandom() + 2.1)* seconds of simulation time, and the decisions that are made determine what the animat does until its think routine is invoked again. (With a lifespan of 200,000 seconds, an animat executes its think cycle approximately 100,000 times.) Each time an animat thinks it makes two main decisions: selecting a behavior and selecting a signal. Two neural networks are responsible for these decisions, and the think routine organizes the inputs to and outputs from these networks and then executes the chosen actions. Before exiting, the think routine sets up an internal trigger (with a small random variance) to tell the animat when to think again. The four phases of the thinking process are:

1. **Sense**. The animat uses its sensors to accumulate information about itself and its surroundings. This information is stored in the internal state variables (such as the friendly tower height and whether or not the animat knows the location of a certain size brick).

2. **Decide.** The animat feeds its sensory input into its neural networks and selects a behavior and signal.

3. **Learn.** The animat reviews its past decisions, evaluates them, and then modifies its neural networks to learn from these experiences.

4. **Act.** The animat calls the subroutines that execute the chosen behavior and signal.

No simulation time elapses while the animat works through the first three phases and begins acting on its decision (although the action itself requires time to execute). The following sections describe each of these phases in greater detail.

During the sense phase, the animat accumulates sensory data and stores the information it receives in its internal state variables and its spatial map to be fed later into its neural networks and used procedurally by the behavior and signal subroutines. The sensors and internal knowledge variables are listed in 3.2.3.

There are three steps in the decide phase:

1. Populate the neural network inputs;

2. Evaluate the neural networks;

3. Save the inputs and outputs into the history queues for later learning.

The animat populates the inputs listed in Table 8 by drawing from its sensor data and its internal state variables. The neural inputs all range from 0 to 1, so the data must be scaled to fit before it can be applied. The inputs to the two neural networks are not exactly the same; the ANN takes as inputs the outputs of itself and the SNN from the last think cycle, but the SNN takes as inputs the outputs of itself from the last think cycle and the outputs of the ANN *from the present think cycle*. This design decision was made to enable the SNN to select signals that would immediately complement behavior selected by the ANN; thus, the order that the neural networks are evaluated is important because their inputs are slightly different.

The two neural networks are then evaluated -- first the ANN and then the SNN. The outputs of the ANN are used as inputs to the SNN, but otherwise their inputs are identical. The inputs (which range from 0 to 1) are scaled to the range (-1, 1) and applied as the activation values of the input layer neurons. The activation values of the input layer neurons are then multiplied by the weights of the connections between the input layer neurons and the hidden layer neurons. The activation values for each hidden layer neuron are summed together and a hyperbolic tangent function is used to determine the activation and output level of each hidden neuron.

The hyperbolic tangent is defined as:

$$\tanh x = \frac{\sinh x}{\cosh x} = \frac{e^{2x} - 1}{e^{2x} + 1}$$

**Equation 4**

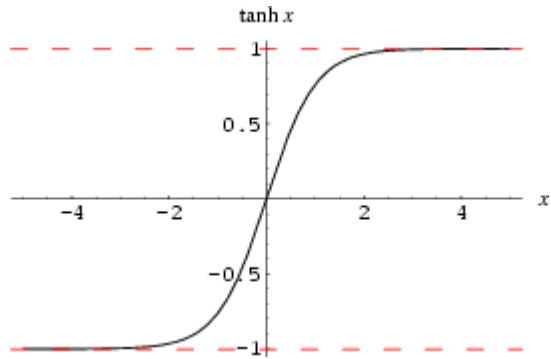

**Figure 5 -- http://mathworld.wolfram.com/HyperbolicTangent.html**

The output of each hidden neuron is multiplied by the weights of the connections between that neuron and the neurons in the output layer. The inputs for each output neuron are summed together and the activation function is applied to this value to determine the output value for each output neuron. So, for neurons *i* and *j*, let $in_j$ be the sum of all *n* inputs to *j* and let $out_j$ be the output from *j*. Let $w_{ij}$ be the weight of the connection between *i* and *j* and let *tanh()* be the hyperbolic tangent function. So,

$$out_i = \tanh(in_i)$$

**Equation 5**

$$in_j = \sum_{i=1,n} w_{ij} out_i$$

**Equation 6**

A small *output bonus* (typically 0.05) is then added to output value of each neuron in the output layer and the resulting output vector is normalized by dividing its value by the sum of all the output values so that the final sum of all the behavior/signal outputs is 1. Experiments demonstrate that this output bonus can help prevent overtraining and enable less rigid thinking.

An output winner is then selected probabilistically based on the normalized output vector. Each output has a chance of being selected equal to its normalized value. A random number between 0 and 1 is generated and then each output is subtracted in turn from this random number; the process is halted when a subtraction results in a negative number and the most-recently subtracted output is selected. The use of an output bonus combined with this probabilistic output selection ensures that the animat will be able to continually explore its state space and try new strategies even when it becomes highly trained.

Finally, the output winner is set to 1 and every other output is set to 0 when the evaluation routine completes and returns, except for the five outputs dedicated to the memory registers. The values of the memory register outputs are fed without modification into the five memory registers that belong to the neural network.

The only time a neural network is evaluated differently than above is during the *random start phase*, during which the animat ignores the input vector and each output is selected with an equal probability. The input context is stored normally, and the random actions are evaluated later during training. This random start phase enables the animat to build a base of experience to learn from and is typically 5% the length of the entire experiment (the length is varied for testing purposes). During the random start phase the outputs of the neural networks are ignored for the purposes of actually making decisions, but they are still used for learning.

Finally, the values in the inputs and outputs (the input and output vectors) are saved to the appropriate queue to be used later during training.  Each queue is circular and holds 100 records in chronological order.

1. *Action Queue* stores record of behavioral states (from the Action Neural Network).

2. *Signal Queue* stores record of signals uttered (from the Signal Neural Network).

3. *Input Queue* stores record of sensory input (from sight and knowledge).

4. *Score Queue* stores record of animat's individual score, based on performance.

The five columns of Table 11 below show the contents of the four input queues of a single animat taken as a snapshot during training.  The data has been translated into English descriptions because each entry is a multidimensional vector.  The queues are circular and the first column shows the index value.  In this example, the contents of line 0 are the farthest in the past and the contents of line 9 are from the most recent complete think phase.  The actual queues have 100 elements, but only 10 are shown in the snapshot below.  I chose a length of 100 because it is sufficient to store enough history to ensure that the causes of any effect would not be lost and it's small enough that learning happens frequently and animats can begin acting on their experiences without extreme delay.

Table 11 -- Below is an abridged example of a particularly interesting set of queues. The Comments column explains what is happening in each row and why it is significant, though not all behaviors can be explained or justified as optimal or even beneficial. Note that 1) the inputs "at X tower" and "near X tower" are *different*, which is why the animat can move from "near friendly tower" to "near enemy tower" between indexes 8 and 9, and 2) this animat was the oldest in its tribe at the time of this snapshot, so it did not obey any commands issued by any other animats. For more detailed, data-intensive examples of the history queues in action, see section 6.1 in the Appendix.

| Index | Input Queue | Action Queue (being performed) | Signal Queue (being sent) | Score Queue | Comments |
|---|---|---|---|---|---|
| 0 | Not carrying a brick; knows where a brick-2 is; thinks tower is height 1. | Fetch brick-2 | Query for tower height | 112 | The animat is fetching for a brick-2, but then learns from its query that the tower height has increased. |
| 1 | Not carrying a brick; updated tower to height 2; knows where brick-3 is. | Fetch brick-3 | Command to attack | 112 | It stops searching for brick-2 and starts fetching brick-3. |
| 2 | Have right size brick; near a friendly animat; near friendly tower. | Return to friendly tower | Command to return to friendly tower | 112 | The animat picks up a brick-3 and starts to return to the tower, issuing a command to do the same to a nearby animat. |
| 3 | Have right size brick; at friendly tower; near friendly animat; near enemy animat. | Drop brick | Command to fetch brick-4 | 122 | It drops its brick onto the tower and command a nearby animat to fetch brick-4, the next one required. |
| 4 | Near enemy animat; near friendly animat; near friendly tower; not carrying a brick; saw | Defend | Command to drop brick | 123 | It then assumes a defensive position to thwart a nearby enemy animat. It also commands a nearby animat to drop the brick it's carrying (which may or may not |

| Index | Input Queue | Action Queue (being performed) | Signal Queue (being sent) | Score Queue | Comments |
|---|---|---|---|---|---|
| | friendly animat fetch brick-4. | | | | have successfully fetched brick-4 yet). |
| 5 | Near enemy animat; combat power 1; target combat power 0; near friendly animat. | Attack | Command to move to enemy tower | 122 | Having just defeated an enemy animat by defending, the animat now sees that its combat power is higher than its enemy's, so it attacks. It also orders a nearby friendly animat to move to the enemy tower. |
| 6 | Not carrying a brick; near friendly tower. | Fetch brick-3 | Query for brick-3 location | 122 | The animat decides to fetch a brick-3, for some reason. |
| 7 | Has a brick that's too small; near friendly tower; near friendly animat; near enemy animat. | Drop brick | Command to attack | 122 | The animat then decides to drop the brick-3 because it's too small to fit on the tower now. It also commands a friendly animat to attack. |
| 8 | Near friendly tower; near friendly animat; near enemy animat. | Move to enemy tower | Query for tower height | 122 | It then decides to move to the enemy tower. |
| 9 | Near enemy tower; near enemy animat; near friendly animat; combat power 1; target | Attack | Command to defend | 123 | The animat decides to attack, even though it is matched for combat power with its target, perhaps hoping the target will attack the animat that is being commanded to defend. |

| Index | Input Queue | Action Queue (being performed) | Signal Queue (being sent) | Score Queue | Comments |
|---|---|---|---|---|---|
|  | combat power 1. |  |  |  |  |

Learning is accomplished using a modified form of Q-learning (a form of temporal difference learning) and is performed similarly on both neural networks. During the Sense and Decide phases the animat accumulates input information and makes decisions, and these pieces of data are stored in the circular history queues, each of length 100. When the animat enters the Learn phase of each think cycle there is a 1% chance that it will attempt to learn from the batch of information it has stored from the 100 most recent decisions. (The animat will not perform any learning until it has made at least 100 decisions.) This batch of 100 recent actions is considered together, and training is performed only periodically so that wide learning windows can be accommodated with low overlap and without excessive, repetitive training.

Each of the 100 most recent decisions is evaluated to determine whether or not it was beneficial -- that is, whether or not it resulted in an increased score for the animat. For every time $t$ in this set of 100, if the animat's score at time $t+1$ is different than it was at time $t$, training is performed using the inputs and outputs of time $t$ as described in the subsections below. However, not all behaviors and signals are immediately beneficial, and the training architecture takes this into account. For example, fetching a brick alone does not increase an animat's score, but fetching must be performed before the brick can be dropped on top of the tribal tower; thus, a (Fetch, Return to Tower, Drop Brick) sequence would result in a reward after the Drop Brick action is completed (assuming the dropped brick is the right size). To facilitate the learning of such sequences, a "forgiveness" factor *ff* is used, and generally set equal to 2. This means that if the animat's score at any of times *t+1* through *t+1+ff* is different from (greater than or less than) its score at time *t*, the inputs and outputs at time *t* will be used for positive or negative training, accordingly. This forgiveness factor enables the animats to learn beneficial behavioral sequences that don't have immediate payoffs by "bridging" over the intermediate unrewarded actions. Animats with a forgiveness factor of 0 can only learn sequences in which every step is immediately beneficial, because no harmful intermediate steps will be "forgiven".

This method of history queuing and forgiveness enables NEC-DAC to delay learning from an action $a_t$ taken at time *t* until later, after (potentially) many other actions have been taken and the results of $a_t$ can be known rather than only approximated or predicted, as is done in learning methods such as the Actor-Critic Method presented by (Barto, Sutton and Anderson, 1983). The Actor-Critic architecture consists of two components, an Actor and a Critic. The Actor takes input from the environment and then selects an action to perform. The Critic predicts the reward that will be received as a result of the action and receives feedback from the environment indicating the accuracy of its prediction. The Critic then generates a reinforcement signal for the Actor so the Actor can adapt its policy in a way the Critic predicts will yield greater future rewards. Over time the Critic is reinforced based on the accuracy of its predictions, and its ever-more-accurate predictions are used to generate better reinforcement signals for the Actor. One advantage of an Actor-Critic system is that it doesn't require frequent reinforcement from the environment in order to learn, instead, the Critic calculates its own feedback based on the accuracy of its predictions. NEC-DAC's forgiveness factor serves a similar function by allowing time periods with no feedback to be bridged. The advantage of NEC-DAC is that the actual results of actions can be used for training, but the downside is that you have to wait to train until you get the results and new information isn't incorporated immediately.

The Learn phase has four steps. These steps are similar to the phases of the think cycle because the learning process requires each decision to be reloaded into the animat's context and re-evaluated, essentially performing the initial think again.

1. Load the input and output vectors;
2. Evaluate the neural network;
3. Calculate the error vector;
4. Propagate the errors and correct them.

First, in order to learn from a past behavior and result, an animat's neural network is reloaded from the Input Queue with the input vector that led to the behavior. Then the behavior-to-be-learned (the target behavior) is loaded from the Output Queue. For example, if an animat is going to learn to perform behavior B when the input vector is I, we load I from the Input Queue and B from the Output Queue. These loads prepare the animat to recreate the context that led to the decision being learned from.

Second, once the context is loaded, the neural network is evaluated in the same manner as described above.

Third, the output from this evaluation is compared to the output from the original decision (the target behavior) and an error vector is calculated based on the difference.

Depending on whether the decision was beneficial or detrimental (determined by examining the Score Queue), the neural network will be trained either towards or away from the target outputs. Even mature networks will frequently have errors due to the probabilistic evaluation method.

Fourth, the differences between the expected outputs and the actual outputs are propagated up through the network. The standard Q-difference learning equations are used, with variables described in Table 12.

Table 12 -- The meanings of the variables in the Q-learning equations.

| Variable | Meaning |
|---|---|
| $t$ | Time |
| $s_t$ | Input vector (state) at time $t$ |
| $a_t$ | Behavior/signal selection (action) at time $t$ |
| $Q(s_t;a_t)$ | Learned action-value function for $a_t$ in $s_t$ |
| $\beta$ | Positive and constant learning parameter equal to 0.05 |
| $r_{t+1}$ | Reward gained for taking action $a_t$ from state $s_t$ |
| $\gamma$ | Discount factor used to indicate that past decisions become less important as they grow more distant in time |

The learning rule is:

$$Q(s_t;a_t) \leftarrow Q(s_t;a_t) + \beta[r_{t+1} + \gamma Q(s_{t+1};a_{t+1}) - Q(s_t;a_t)]$$

**Equation 7**

Every learning step improves $Q$ for a given state-action pair. The value of $r$ depends on the success of the state transition, and can be positive or negative. For the Action Neural Network, animats are rewarded for placing bricks on their tower and for winning combat. Table 13 shows the values used for *r* for each of these events. The relative values of these rewards are important, and when they are changed the animats modify their behavior to maximize their scores. The values below were chosen because they result in a mix of combat and brick placement that I judged to be balanced -- that is, neither set of behaviors pushes the other to zero.

Table 13 -- Action Neural Network reward values for events.

| Event | Reward *r* |
|---|---|
| Place brick on tower | 10 |
| Win combat | 1 |
| Lose combat | -1 |

Some actions that an animat takes are the result of being commanded to act by another animat. In such cases, the animat being trained utters a *feedback signal* indicating that the reward it earned for performing the action is attributable to the command it obeyed. Nearby friendly animats use this signal to train their Signal Neural Networks and to learn how to issue commands that lead to rewards for the listeners. The value of *r* for the Signal Neural Network is calculated based on these feedback signals. Rather than being based on events that affect the learning animat (the signaler), *r* is based on events affecting the animat that *listened to and obeyed* the signal (the listener) that the learner uttered. If animat A commands animat B to fetch a brick and B does so, and the fetching ultimately leads to B being rewarded, then A's SNN will be trained with an equivalent reward. The result is that SNNs are trained to utter signals that result in increased scores for listeners. Additionally, any other friendly animats within signaling range when B tells A about its success will be trained to utter the same signal A uttered in the same circumstances. For example, animat A commands animat B to fetch brick-3 and animat B obeys. Animat B then returns to its tower and drops brick-3 on top. Later, when animat B goes through its learning process it will see that the fetch brick-3 behavior led to the eventual reward obtained for stacking brick-3 onto the tower, and it will know that it fetched brick-3 because it was commanded to do so. As a result, animat B will utter a feedback signal (after learning) that includes information about its state at the time it was commanded, the command it received, and the result of that command. This feedback signal will be heard by every friendly animat within signal range, which may or may not include animat A.

The variable $\gamma$ equal to *(0.75^($\Delta t$))* was used to reduce the importance of past decisions exponentially, where $\Delta t$ is the number of think cycles between when the action was performed and when the reward was realized. Use of this variable along with past rewards enables animats to learn sequences of actions and signals that only pay off at the end and don't have intermediate rewards after each step in the sequence. For example, the fetch-return-drop sequence is very common and it's important that the fetch and return to friendly tower behaviors are recognized as contributing to the eventual reward gained from the drop. However, the contribution of past behaviors must be discounted relative to more recent behaviors to ensure that non-contributing behaviors (which are more likely to appear farther back in history as the animat learns to behave more efficiently) are eventually weeded out.

Several values for the coefficient of learning, $\beta$, were tested, and 0.05 was eventually selected because it works -- other values may also work. At values above approximately 0.1 the updating function (Equation 7) diverges; lower values result in slower training or no training at all. (Larger coefficients were needed when the sigmoid instead of the hyperbolic tangent was used as the activation function, but training with the sigmoid was never as fast as with the hyperbolic tangent.)

A value estimate is calculated for each output neuron based on its error share, and the error is propagated up through the layers of the neural network in a manner similar to standard backpropagation, with variables described in Table 14.

**Table 14 -- The meanings of the variables used in the backpropagation equations.**

| Variable | Meaning |
| --- | --- |
| $w_{ij}$ | Weight of the connection between neuron $i$ and neuron $j$, bounded between (-1,1) |
| $\beta$ | Coefficient of learning |
| $a_i$ | Activation of neuron $i$ |
| $\delta_j$ | Error of neuron $j$ |
| $\beta$ | Positive and constant learning parameter equal to 0.05 |
| $f$ | Activation function |
| $f'$ | Derivative of the activation function |
| $t_j$ | Target value |

At a given time step, the error signal for output neuron $j$ is:

$$\delta_j = (t_j - o_j)$$

**Equation 8**

And the weight delta is:

$$\Delta w_{ij} = \beta \delta_j f'(a_j) f(a_i)$$

**Equation 9**

The only difference if *j* is a hidden neuron is that its error is calculated as, for all output neurons *k*:

$$\delta_j = \sum_k (\delta_k + w_{jk})$$

**Equation 10**

The neural network implementation connects to the temporal difference learning theory thusly:

- the set of outputs corresponds to Q-learning value estimates,
- the error signals correspond to Q-learning error,
- the weights correspond to the policy being trained.

After a decision has been made and training has been performed the subroutine for the selected behavior is executed and the following maintenance steps are performed:

1. If the subroutine requires the animat to move it will set the destination variable to the appropriate location and the animat's new velocity will be calculated.
2. A check is made to verify that the animat stays within the world's boundaries.
3. The animat's health is bounded to minimum of 1 with no maximum.
4. If the Combat Power variable is less than 1 it is increased by 0.1 to a maximum of 1.
5. If Just Attacked is non-zero it is decreased by 0.2 to a minimum of 0.
6. The animat's health is stored in the Score Queue for later use in training.

# 4  Experiments

The experiments I performed fall into the three categories I described in the introduction and are named as indicated in bold:

1. **PROB-REAS:** Probabilistic reasoning. Modifications are tested successively.

   a. *Probabilistic behavior* in which the neural network outputs are percentages that are used by a random number generator to select a behavior to perform or a signal to utter;

   b. Use of an *output bonus* to boost the probability of selecting a low-likelihood behavior;

   c. Use of an *obedience multiplier* to vary the likelihood that a listening animat will obey a command signal given by another animat;

   d. Use of *memory registers* that enable animats to save arbitrary state information through time;

   e. Presence or absence of a *random start phase*.

2. **SOC-STRUCT:** Signals and information sharing within a social structure. Social structures compete against each other.

   a. *No authority*, in which no animats can command any others.

   b. *No hierarchy*, in which every animat can command any animat in its tribe;

   c. *Age hierarchy*, in which an animat can command every younger animat in its tribe;

   d. *Merit hierarchy*, in which an animat can command every lower-performing animat in its tribe.

    e. *Age and merit hierarchy,* in which an animat can command every other animat in its tribe that is both younger **and** has a lower score.
3. **ROLE-DIFF:** Role differentiation.
    a. Examining *differentiation* of behavior and signaling choices using the differentiation factor.

Following these three major experiments, this chapter also contains a section on Miscellaneous Experiments (4.4) and an elaboration on Neural Network Weights (4.5).

In these experiments animats are created and destroyed on a time schedule that is similar to "generations" of living organisms, but it's important to remember that there is no reproduction, recombination, or evolution occurring. Animat performance as measured by score fits roughly into a Pareto distribution, as can be seen in Figure 6. In the experiment data, scores are normalized when they are used to compare tribes in competition with each other so that the sum of the scores of the two tribes in a single trial always adds up to 1000.

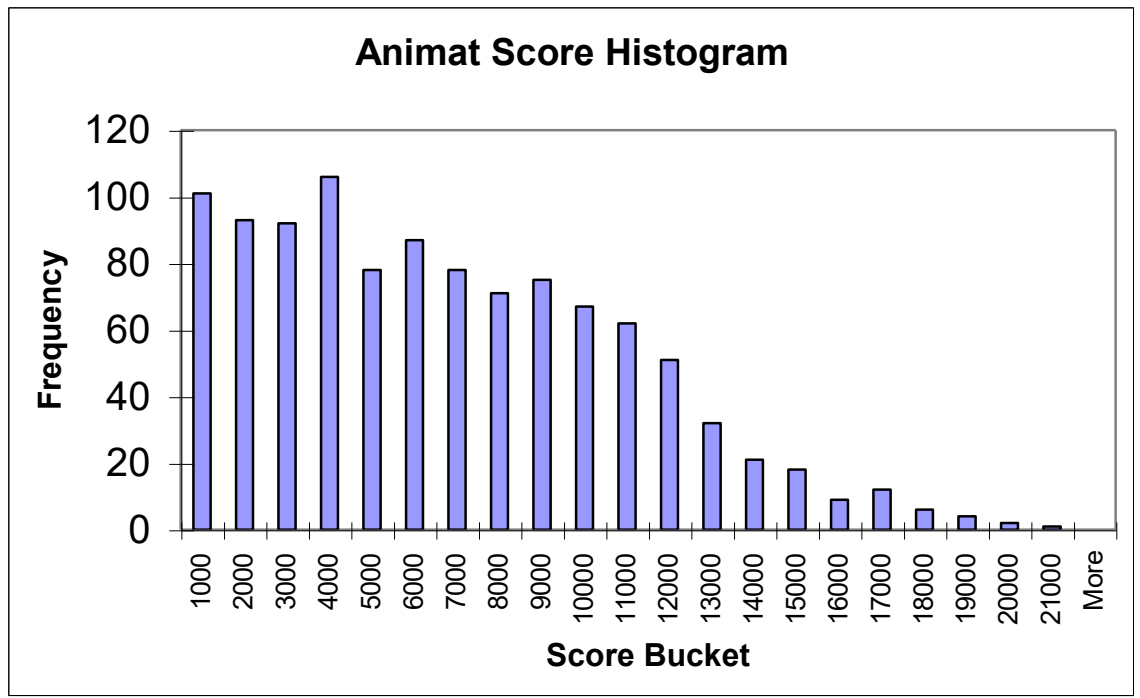

**Figure 6** -- This histogram shows the number of animats that scored within each bucket range, taken from a total population of 1534 over the course of 13 trials and roughly fits a Pareto distribution. Animats that did not live out their full lifespan (because they were created "old" near the beginning of a trial or died "young" when a trial ended) were removed from the set, leaving the 1066 animats that are represented in the figure.

Each trial lasts for 1 million simulation seconds (i.e., 10 million ticks), and each animat has a lifespan of 200,000 seconds (during which time it will make approximately 100,000 decisions with each neural network). At the beginning of a trial, animats are created with staggered ages so that they are destroyed in an evenly timed sequence. The exact timing depends on the number of animats in each tribe; most trials use tribes with 10 animats, in which case animats would be initially created with ages in increments of 20,000 seconds (200,000 divided by 10), starting with age 0. Thus, one animat from each tribe is destroyed every 20,000 seconds. Experiments show that an animat's performance does not improve much after 200,000 seconds of learning, which is why this age was chosen for destruction. The trial length was chosen arbitrarily, and doesn't matter as long as it is long enough to prevent any of the initial animats from living until the end. This aging mechanism has several advantages:

1. It simulates population flow over generations in a biological population, enabling the exploration of how culture is transmitted.
2. It mitigates the effects of any outlying animat behavior. Since an animat's score is removed from its tribe's total when the animat dies, an outlier animat can only have a limited effect on the overall performance of a tribe.
3. As animats are removed from the population, the effects of the culture can be isolated from any systematic quirks that arise within an animat but are not transmitted through signals.
4. The system can essentially run multiple related "mini-trials" within each trial, since there are 50 distinct sets of animats that make up each tribe over the course of a single run as individual animats are born and die, and hence are added and removed from the tribe. These 50 sets of animats overlap each other, and as time advances animats learn and change. The 50 sets are not independent, but there is complete turnover within a tribe every 200,000 seconds (giving five non-time-overlapping sets).

The following three sections describe the experimental set up used in each of these experiments and the results obtained. Correlation coefficients were calculated using simple linear correlation (Pearson r) and represent the square root of related variance between the two variables, $x$ and $y$ (Equation 11).

$$r = \frac{\sum xy}{\sqrt{\sum x^2 \sum y^2}}$$

**Equation 11**

Significance is measured as the p-value of a linear regression (performed with Microsoft Excel), with results accepted at the 95% confidence level unless otherwise noted -- lower p-values indicate a higher degree of confidence that the independent variable in question influences score. Where p-values are not explicitly listed, I follow the standard convention of marking 95% confident values with an asterisk (*) and 99% confident values with a double-asterisk (**). The primary variables measured are score and the two differentiation factors, and for smoothing purposes their values are calculated for each tribe as the average of 1,000 intermediate measurements taken 1,000 seconds apart (ignoring any measurements taken during a random start phase, if applicable). Score is usually a dependent variable, and the regressions are used to demonstrate that other variables influence score.

## *4.1 PROB-REAS*

The PROB-REAS experiments demonstrate that tribes of animats that incorporate probabilistic reasoning into their decision-making process can outperform tribes that do not. Five probabilistic mechanisms are introduced in these trials, and tribes with the modifications are pitted against tribes without them. The five mechanisms, with trial designators in parentheses, are:

1. *Probabilistic behavior* (PS) in which the neural network outputs are percentages that are used by a random number generator to select a behavior to perform or a signal to utter. Opposed by winner-takes-all (nPS) in which the output with the highest activation is always chosen.

2. Use of an *output bonus* (OB) to boost the probability of selecting a low-likelihood behavior. The output bonus is added to the neural network outputs before they are normalized and it prevents outputs from being trained down to zero probability. Opposed by no output bonus (nOB).

3. Use of an *obedience multiplier* (OM) to vary the likelihood that a listening animat will obey a command signal given by another animat. Obedience multiplier values from 0 to 64 were tested.

4. Use of *memory registers* (MR) that enable animats to save arbitrary state information through time. Opposed by no memory registers (nMR).

5. Use of a *random start* (RS) phase in which animats behave randomly for a period of time to accumulate data on the whole range of actions and signals. Opposed by no random start (nRS).

The results of PROB-REAS demonstrate the effects of each of these modifications, some of which improve the competitiveness of a tribe of animats while others are neutral. The effective modifications were later used in the ROLE-DIFF and SOC-STRUCT experiments. Within each trial, a random subset of signals is disabled for each tribe (the disabled subset is different for each tribe and each trial); each signal has a 50% chance of being enabled for a particular tribe in a particular trial. By running dozens of trials, statistically significant data was accumulated and the performance of animats with varying capabilities was analyzed by controlling for the enabling/disabling of signals. As explained in the introduction to this chapter, each trial can be considered as a collection of 50 interdependent mini-trials (by animat turnover between generations), so the statistical relevance is actually higher than would be supposed just from the number of trials. Furthermore, as we will see in section 4.2 (Table 21 and Table 23), the specific signals enabled or disabled did not have a large effect on animat performance.

Each set of trials in the PROB-REAS experiment is labeled according to the probabilistic mechanism being tested. For instance, the "PSvnPS" set involves two tribes: tribe 1 uses *probabilistic output selection* (PS) and tribe 2 uses winner-takes-all output selection (nPS). In nPS, whichever output has the highest activation is automatically selected as the behavior or signal to be executed. In PS, the winning output is selected probabilistically as described in section 3.2.5 -- the output activation values are normalized and a random number generator is used to pick a winning behavior proportionately to the relative output values. Table 15 shows the result of PSvnPS.

Table 15 -- PS tribes completely dominate nPS tribes.

| Tribes | PS Tribe Wins | nPS Tribe Wins | Average PS Tribe Score | Average nPS Tribe Score | Correlation between score and PS | P-value |
|---|---|---|---|---|---|---|
| PSvnPS | 9/9, 100% | 0/9, 0% | 952 | 48 | 0.987 | 3.6E-14 |

I only ran 9 PSvnPS trials because it became obvious that probabilistic output selection created such an enormous advantage that further demonstration was unnecessary. Winner-takes-all selection squelches the behavioral variety required to successfully complete the brick-stacking task.

However, not all of my probabilistic modifications were as fruitful. In earlier, less controlled experiments it appeared that adding a small *output bonus* to neural network outputs before normalizing them aided performance and problem-space exploration by preventing any actions or signals from being trained to zero probability, no matter how badly they hurt animat scoring (see section 3.2.5). In the final configuration, the results of OBvnOB show that use of a small output bonus (5%) has no significant effect on performance, and analysis of the underlying data reveals that none of the neural network outputs is trained to a value of less than 3% even without any bonus. Larger output bonus values (10%+) quickly put a tribe at a distinct disadvantage -- a large output bonus makes an animat behave randomly as its unmodified output values are overwhelmed by the bonus.

The *obedience multiplier* modification increases the neural network output associated with an action that an animat has been commanded to perform by another animat with authority (Equation 12). For instance, if animat A was commanded by animat B to execute the drop brick action, the next time A performs a think cycle the value of the drop brick output ($o_b$) on the ANN will be multiplied by the obedience multiplier ($x$) before it is normalized with the other outputs (yielding $m_b$). Since there will be some delay between when A utters the signal and B hears it and acts, poor commands coupled with high obedience multipliers could be disadvantageous. Table 16 defines the variables used in the obedience multiplier equations.

Table 16 -- The variables used in the obedience multiplier equation.

| Variable | Meaning |
|---|---|
| $b$ | Behavior being commanded |
| $o_b$ | Value of output $b$ from the neural network (the raw, non-normalized probability that behavior $b$ will be chosen) |
| $m_b$ | Modified value of output $b$ after the obedience multiplier is applied |
| $x$ | Obedience multiplier |

$$m_b = xo_b$$

**Equation 12**

Obedience multipliers from 0 to 64 were tested by pitting tribes against each other with randomly assigned values. Animats in a tribe with an obedience multiplier of 0 would be incredibly perverse, utterly refusing to select any action they're commanded to perform. Animats in a tribe with an obedience multiplier of 1 would essentially ignore commands. Values of 2 or greater would correspondingly increase the likelihood of a command being obeyed. Figure 7 shows the results of 80 trials -- 160 tribes in total -- with obedience multipliers between 0 and 9 and Figure 8 includes values for obedience multipliers of 32 and 64.

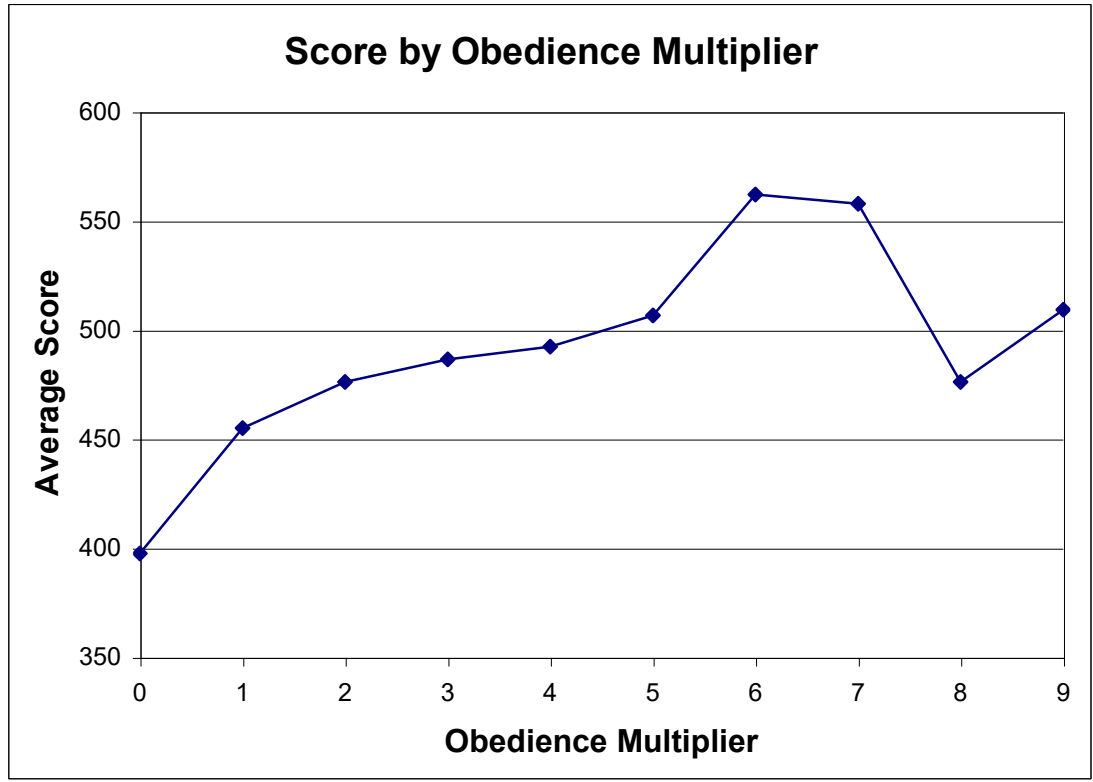

**Figure 7 -- This chart shows the average scores of tribes using the Obedience Multiplier value indicated on the X-axis.**

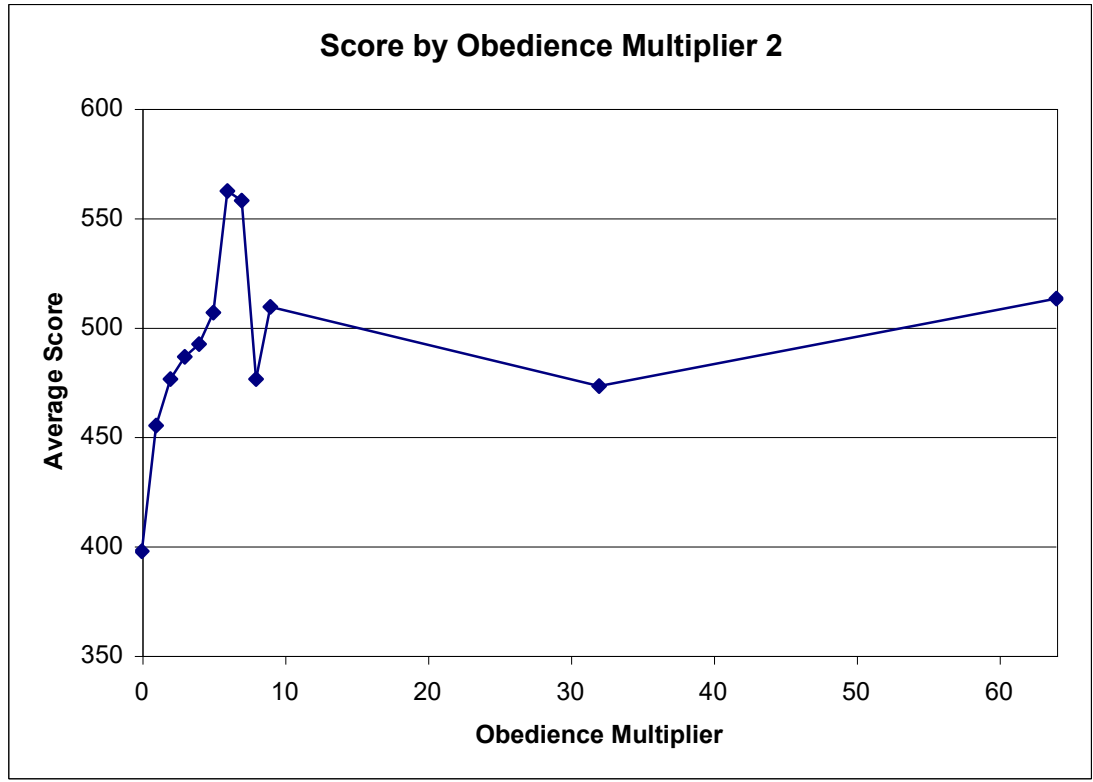

**Figure 8 -- This extended graphs shows that higher obedience multiplier values also stay close to 500.**

The data indicates that a tribe with a finite obedience multiplier will out-perform a tribe that obeys absolutely (i.e., an infinite obedience multiplier). This result is logical considering the results of the PSvnPS experiment, because absolute obedience would greatly reduce the number of instances in which probabilistic output selection would come into play. Animats with the ability to disobey authority benefited from doing so occasionally, and the obedience multiplier mechanism leads to a level of disobedience proportional to the inclination of the animat towards a behavior other than the one commanded. This process is very similar to the way humans decide whether or not to obey authority, and the performance results mirror my intuition of humanity. People who are more "loyal" to the social structure give more weight to the commands given to them and less weight to their own desires and perceptions. A human who occasionally disobeys authority (finite, non-zero OM) will almost certainly prosper above one who always obeys (OM = infinity), yet a human who never obeys authority (OM = 0) will do worse than either of the other two.

The *memory registers* (MR) modification is probabilistic in the sense that the registers enable an animat to store values from the initially random outputs of its neural networks and assign arbitrary meanings to them that develop over time. MR animats can use their memory registers to store state information between decisions and to share information between their Action and Signal Neural Networks (section 3.2.4). In the current implementation animats can only store and read values from these registers, but future experiments could include registers that perform more complex mathematical operations on their contents. Table 17 shows the performance of tribes with (MR) and without (nMR) memory registers, and it's clear that although the encodings used to store information in the registers are nearly indecipherable, tribes with memory registers have a distinct advantage.

Table 17 -- Tribes with memory registers almost always defeat and significantly outscore tribes without.

| Tribes | MR Tribe Wins | nMR Wins | Average MR Tribe Score | Average nMR Tribe Score | Correlation between score and MR | P-value |
|---|---|---|---|---|---|---|
| MRvnMR | 17/18, 94% | 1/18, 6% | 548 | 452 | 0.753 | 1.1E-07 |

The values in the memory registers are extremely volatile in the early part of an animat's life, but over time they stabilize and drift more slowly, reflecting the overall increased stability shown by the rest of the animat's behavior. Still, even when the values become more stable the most highly performing animats have registers that are related to their most-used behaviors. Consider Animat 5002 from tribe 050523-002009-0043-2 (Tribe 43-2), one of the top-performers in its trial. Animat 5002 specialized in fetching brick-3 and returning it to the tower almost to the exclusion of all the other bricks, and the five memory register values generated by its Signal Neural Network show strong correspondence with the fetch-return-drop sequence for brick 3.

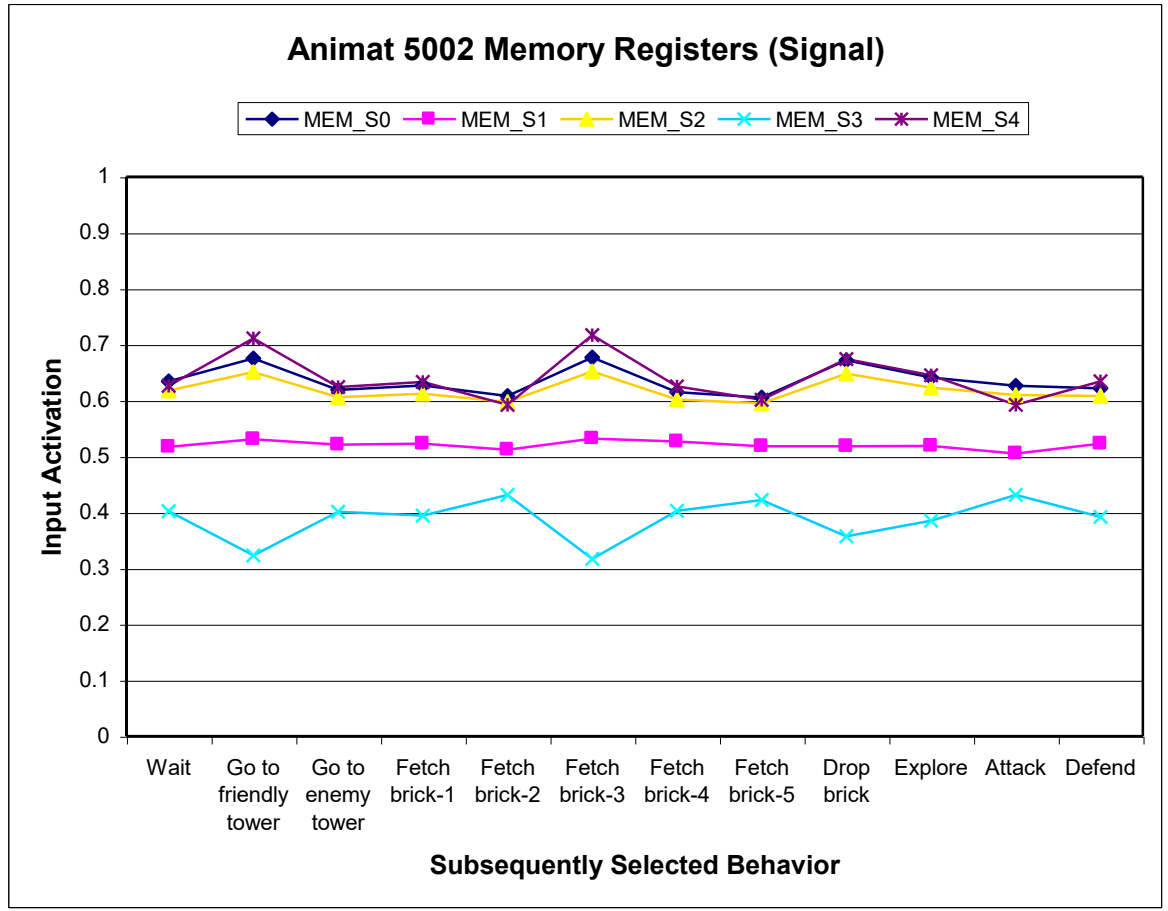

**Figure 9** -- This figure shows the average activation values of the inputs corresponding to Animat 5002's five signal memory registers broken down by the behavior that was subsequently selected as a result of those inputs. This means that, e.g., the input corresponding to MEM_S4 had an average activation value of 0.72 when Fetch Brick-3 was chosen and 0.5 when Fetch Brick-2 was chosen.

From Figure 9 we can see that, except for MEM_S1, which is relatively flat, the average input activation values of Animat 5002's signal memory registers show spikes for the Go to Friendly Tower, Fetch Brick-3, and Drop Brick behaviors. The MEM_S3 register appears to spike oppositely from MEM_S0, MEM_S2 and MEM_S4, spiking down for those behaviors while the others spike up. These memory registers are set by the Signal Neural Network one think cycle before they are seen by the Action Neural

Network, and they appear to be being used to let the SNN indicate what behavior it wants the ANN to execute. (We'll revisit Animat 5002 in section 4.2.) Data also shows that the memory registers fed from the SNN are significantly more important than the memory registers fed from the ANN (see section 4.5 in the for more details).

The final modification listed above, use of a *random start phase*, was the most important to my experimentation process; preventing newly created animats of generations other than the first from experiencing a random start phase was the change that led me to the first scenarios in which teaching and learning signals became significantly advantageous. The random start phase was initially implemented so that when an animat was created (in any generation within a trial) it went through a random start phase during which it acted randomly and explored its state space before settling down and obeying the outputs of its neural network. Animats using this random start phase drastically outperformed animats with no random start phase and no teaching/learning.

However, I had discovered that cross-generational teaching and command authority yielded little additional improvement. Tribes that could teach performed nearly the same as non-teaching tribes that used a random start phase. However, when I changed the random start phase implementation so that only animats in the first generation experienced a random phase (a weaker implementation), the benefits of cross-generational teaching became evident. Newly created animats in generations other than the first could no longer ignore their neural networks and had to, from the beginning, deal with the scoring consequences of their actions. As such, they had difficulty escaping local maximums and never learned on their own how to perform as successfully as animats in the first generation that did use a random start phase. However, tribes with a *social structure* and *command authority*, tribes that could teach and learn, continued to prosper because older animats taught younger animats how to behave and how to succeed. The process of learning from older animats essentially replaced the random start phase with an apprenticeship phase that lessens in effect as the number of animats older than the learner shrinks over time -- similar to how "heat" is reduced over time in simulated annealing algorithms (Kirkpatrick, Gelatt, and Vecchi, 1983).

Which raises the question: why are teaching and learning signals valuable if similar performance can be achieved using a random start phase? The answer is that a random start phase is only a viable option if the animat is allowed to ignore the consequences of its decisions. During my random start phase, animats ignored health and scoring penalties incurred from lost combat and did not take advantage of score bonuses received for correctly placing bricks. It is impossible for a real organism to similarly ignore the consequences of its actions in a real environment. The following SOC-STRUCT experiment elaborates on the different types of social structures I tested and their effects.

## 4.2 SOC-STRUCT

The purpose of SOC-STRUCT is to test the ability of the animats to learn to use their signaling capabilities within five types of social hierarchies and to examine the effects of the social structures on animat performance. The social structures are defined by the rules that determine which animats can give commands to which other animats, called *authority*.

1. *No authority (NA)*, in which no animats can command any others.
2. *No hierarchy (NH)*, in which every animat can command any animat in its tribe;
3. *Age hierarchy (AH)*, in which an animat can command every younger animat in its tribe;
4. *Merit hierarchy (MH)*, in which an animat can command every other animat in its tribe with a lower score;
5. *Age and merit hierarchy (AM)*, in which an animat can command every other animat in its tribe that is both younger and has a lower score.

Within each of these social structures, the performance of the animats was evaluated in light of two questions:

1. *Social structure.* What effect do social structures have on tribe performance?
2. *Signal usefulness.* Which signal combinations enable fastest learning and greatest competitiveness? Which individual signals contribute most to success?

Just as in PROB-REAS, within each trial a random subset of signals is enabled for each tribe (the enabled subset is different for each tribe and in each trial); any given signal has a 50% chance of being enabled for a particular tribe in a particular trial. By running dozens of trials, statistically significant data was accumulated and the performance of animats with varying capabilities was analyzed by controlling for the enabling/disabling of signals and combinations of signals. As explained in the introduction of this chapter, each trial can be considered as a collection of 50 interdependent mini-trials (by animat turnover between generations), so the statistical relevance is actually higher than would be supposed just from the number of trials. Over time it became clear that the specific signals that were enabled had much less effect on performance than did the number of enabled signals. Table 21 and Table 23 show that the specific signals enabled for a tribe in a given trial do not greatly affect the outcome.

Each set of trials in the SOC-STRUCT experiment is labeled according to the authority structure used by the tribes in the set. For instance, the NHvMH set involves two tribes: tribe 1 uses the No Hierarchy structure and tribe 2 uses the Merit Hierarchy structure. The various structures were compared in two ways:

1. Competitively, by pitting them against each other, with results are shown in Table 18;
2. Solo, by running trials with a single tribe that is rated based on the number of seconds it requires to reach a score of 200, with results farther down in this section.

Table 18 -- Results of SOC-STRUCT competition between tribes with differing social structures. Note the marked p-values: † not within 95% confidence; ‡ no significance.

| Tribes | Tribe 1 Wins | Tribe 2 Wins | Average Tribe 1 Score | Average Tribe 2 Score | Correlation between score and Tribe 1 | P-value |
|---|---|---|---|---|---|---|
| NAvNH | 3/21, 14% | 18/21, 86% | 395 | 605 | -0.63 | 7.3E-06 |
| NAvAH | 1/20, 5% | 19/20, 95% | 362 | 638 | -0.89 | 1.2E-14 |
| NHvMH | 15/18, 83% | 3/18, 17% | 584 | 416 | 0.52 | 1.1E-03 |
| NHvAH | 7/24, 29% | 17/24, 71% | 421 | 579 | -0.47 | 7.3E-04 |
| MHvAH | 13/29, 45% | 16/29, 55% | 469 | 531 | -0.23 | 0.088† |
| AHvAM | 32/61, 52% | 29/61, 48% | 509 | 491 | 0.05 | 0.571‡ |

Table 18 shows that the social structure of a tribe has a significant effect on competitiveness. I didn't run every possible match-up due to time constraints (each set of trials can take a week or more) but I ran enough combinations to rank the social structures in order of power, from most to least:

1. Age Hierarchy and Age and Merit Hierarchy, tied;
2. No Hierarchy;
3. Merit Hierarchy;
4. No Authority.

The results of the MHvAH set did not meet the 95% confidence requirement (with a p-value of 0.088) but I decided to accept them anyway because NH significantly beat MH in NHvMH and AH significantly beat NH in NHvAH. It appears there is triangle effect such that MH performs disproportionately better against AH than it does against NH, but it is still pretty clear that AH is the dominant social structure.

My original intuition was that MH would beat AH, but it appears that age and experience give an advantage during instruction that isn't reflected by an animat's score -- it is advantageous for a younger animat to obey an elder *even if* the younger animat has a higher score than the elder. The experience accumulated by an older animat is not directly reflected in its score, and when a younger animat obeys an elder the younger animat learns to behave in a way that leads it to a higher score. As an example, consider an older animat that knows, in its old age, how best to behave, but has a low score because it behaved poorly as a youth. However, even in that circumstance, why should a younger animat that has already surpassed an elder benefit from obeying the old-timer? I thought that perhaps the advantage of AH was due not to young, successful animats obeying their elders, but to old animats that didn't have to change their ways to obey youngsters. I developed the Age and Merit Hierarchy social structure to examine this scenario more closely, but as Table 18 shows AM and AH are evenly matched after 61 trials, with AH having a very slight edge.  Therefore I conclude that the advantage AH has over MH is due to young, high-scoring animats obeying orders from older, lower-scoring animats.

Similar, though less pronounced effects were seen in the solo trials, in which a single tribe per trial is rated based on the time required to reach a score of 200.  The 200-point target was chosen because it is high enough that it is impossible to reach until significantly past the random start phase, but low enough that no trial is likely to last longer than the competitive trials above.  Table 19 shows the results of the social structures run solo.

Table 19 -- Average number of seconds required for solo tribes of each social structure to score 200 points.  Each social structure was run for between 40 and 50 trials.  AM is omitted due to insufficient data.

| Social Structure | Average seconds to score 200 |
|---|---|
| AH | 294,000 |
| MH | 320,000 |
| NA | 327,000 |
| NH | 332,000 |

As can be seen, the performance numbers for the social structures are far closer when they are run non-competitively.  Not only do the animats not have to deal with attacking and defending behaviors, but resources are relatively more abundant and less likely to be controlled or occupied when needed.

We can use the *relative average scores* shown in Table 20 to evaluate whether or not the effectiveness of a signal depends on the social structure in use. The relative average score for a signal/social structure pair (A,B) is calculated as follows:

- X is the set of all tribes using social structure B and having signal A enabled,

- Y is the set of all tribes using social structure B (X is a proper subset of Y),

- The relative average score of (A,B) equals the average score of all the tribes in X divided by the average score of all the tribes in Y.

Table 20 -- Relative average scores of tribes by social structure and signal capability. Each cell represents the relative average score of the tribes with the enabled signal indicated on the left divided by the average score of all tribes within the social structure indicated by the column. The cells marked with parenthetical numbers are mentioned in the following paragraphs. It is critical to remember that each tribe in this data set fought only against opposing tribes with different social structures; e.g., no MH tribe fought against another MH tribe. * indicates p-value less than 0.05.

| Signal Enabled | NH Tribes | MH Tribes | AH Tribes | Std. Dev. |
|---|---|---|---|---|
| Null Signal | 1.00 | 1.00 | 1.00 | 0.00 |
| Go to Friendly Tower | (1) 1.22* | 1.12* | 1.07* | 0.08 |
| Go to Enemy Tower | 1.02 | 0.96 | 0.99 | 0.03 |
| Fetch Brick 1 | 0.99 | 0.99 | 0.93 | 0.03 |
| Fetch Brick 2 | 0.97 | 0.97 | 1.02 | 0.03 |
| Fetch Brick 3 | 0.97 | 1.04 | 0.98 | 0.04 |
| Fetch Brick 4 | 0.98 | 0.97 | 1.01 | 0.02 |
| Fetch Brick 5 (2) | 1.06 | 1.07 | 1.03 | 0.02 |
| Drop Brick (3) | 1.06 | 1.17* | 1.11* | 0.06 |
| Explore | 0.99 | 0.97 | 1.01 | 0.02 |
| Attack | 1.00 | 0.99 | 1.03 | 0.02 |
| Defend | 1.02 | 1.04 | 0.94 | 0.05 |
| Tower Height | (4) 0.89* | 1.02 | 1.04 | 0.08 |
| Brick-1 Location | 1.06 | 1.01 | 0.94 | 0.06 |
| Brick-2 Location | 1.06 | 0.99 | 0.99 | 0.04 |
| Brick-3 Location | 1.00 | 0.96 | 1.01 | 0.03 |
| Brick-4 Location | (5) 0.91* | 1.05 | 1.04 | 0.08 |
| Brick-5 Location | 1.03 | 1.00 | 0.97 | 0.03 |
| Std. Dev. (6) | 0.070 | 0.057 | 0.046 | |

From Table 20 we see that signaling capabilities did not make much of a difference in tribal competitiveness. Signaling capabilities had the most effect on the NH tribes, as shown by the standard deviations (6) in the final row. The most successful social structure, AH, had the smallest standard deviation between the relative average scores of tribes with various signaling capabilities. Still, some signals did show an effect. The Go to Friendly Tower signal capability led to higher scores for tribes in all social structures (particularly NH (1)), as did Drop Brick (3). It should be clear why these two signals are important: both are key to the fetch-return-drop sequence that most directly results in high score rewards. The rewards of the fetch step of that sequence must be distributed across the fetch commands for bricks of all sizes, however, which is probably why none of them stands out significantly. Fetch Brick-5 (2) does appear to be a bit more significant than the others, but I don't know why. It is also unknown why Tower Height (4) and Brick-4 Location (5) did so poorly for NH tribes specifically.

Why did the signaling capabilities of the tribes have so little impact on score? My first hypothesis was that the effects of the social structures were overwhelming the effects of the signaling capabilities; however, Table 21 shows that even when tribes with the same social structure are matched against each other no single signal has a drastic effect on performance. What *does* happen when like fights like is that the correlation between the *number* of enabled signals and the score increases, as can be seen in Table 22. As more signals are disabled there is a graceful degradation of performance. The later ROLE-DIFF experiments confirmed that Score and ADF were only slightly affected by the number of enabled signals, but SDF was impacted more strongly (see Table 23).

Table 21 -- This table shows the relative average scores for AH tribes with various signals enabled and the correlation between signal enabling and score . In these trials, AH tribes were matched up against each other. As can be seen, no signal stands out as particularly advantageous, but Go to Friendly Tower and Drop Brick rank highest. * indicates p-value less than 0.05. ** indicates p-value less than 0.01.

| Signal Enabled | Relative Average Score | Correlation Between Signal Enabling and Score |
|---|---|---|
| Null Signal | 1.00 | --- |
| Go to Friendly Tower | 1.09** | 0.34** |
| Go to Enemy Tower | 1.02 | 0.05 |
| Fetch Brick 1 | 1.03 | 0.10 |
| Fetch Brick 2 | 1.01 | 0.01 |
| Fetch Brick 3 | 1.02 | 0.08 |
| Fetch Brick 4 | 1.01 | 0.05 |
| Fetch Brick 5 | 1.01 | 0.01 |
| Drop Brick | 1.09** | 0.41** |
| Explore | 0.97 | -0.09 |
| Attack | 0.98 | -0.09 |
| Defend | 0.98 | -0.11 |
| Tower Height | 1.00 | 0.03 |
| Brick 1 Location | 1.04* | 0.19* |
| Brick 2 Location | 1.04 | 0.20 |
| Brick 3 Location | 0.99 | -0.02 |
| Brick 4 Location | 1.00 | 0.01 |
| Brick 5 Location | 0.96 | -0.14 |
| Std. Dev. | 0.04 | |

**Table 22** -- Correlation between tribal score and the number of enabled signals. This correlation measures the relationship between the number of signals a tribe has enabled and its score; a positive correlation indicates that a high number of enabled signals leads to a high score, and a negative correlation indicates that a high number of enabled signals leads to a low score. The signals were broken into three subsets. * indicates p-value less than 0.05.

| Signal Subset | NA vs. Others | NH vs. Others | MH vs. Others | AH vs. Others | AH vs. AH |
|---|---|---|---|---|---|
| All Signals | 0.13 | 0.17* | 0.21* | 0.13* | 0.24* |
| Fetch Commands | 0.12 | -0.01 | 0.07 | -0.03 | 0.11* |
| Brick Locations | 0.14 | 0.08 | 0.03 | -0.08 | 0.11* |

The three signal subsets used in Table 22 are:

1. *All Signals*, which contains the entire set of signals.

2. *Fetch Commands*, which contains Fetch Brick-1 through Fetch Brick-5.

3. *Brick Locations*, which contains Brick-1 Location through Brick-5 Location.

Only the All Signals subset contains the Go to Friendly Tower and Drop Brick commands, so it's clear that the Fetch and Location signals also correlate positively with high scores when the social structure of both tribes is the same. (The data in the NA vs. Others column is misleading though included for completeness; because the NA social structure performed so poorly and received uniformly low scores, slight fluctuations with no significance appear to have caused meaningless correlations with high p-values. Clearly, a tribe with animats that will not obey each other cannot benefit from signals that enable animats to send commands that will be ignored.)

These results do not match my initial hypotheses, but they indicate that a broader selection of signaling abilities gives a tribe an advantage even when no individual signal is decisive. I had expected that the neural network mechanism would identify a few key signals and use them advantageously and repetitively, and that this usage would be immediately visible in the correlation and relative average score statistics. However, it appears that the usage patterns are subtle, and that although correlations exist it is difficult to interpret the exact manner in which the signals are being used advantageously. Signaling capabilities are useful in the aggregate, and efficiency degrades gracefully when fewer signals are enabled.

## 4.3  ROLE-DIFF

In the ROLE-DIFF experiments, tribes of animats with randomly enabled signaling abilities (as above) were pitted against each other to determine the connection between endogenous role assignment (as measured using the Differentiation Factors) and success. Tribal differentiation was tracked across generations as animats were created and destroyed to determine under what circumstances role differentiation is advantageous. All the probabilistic modifications from PROB-REAS were included, except for the output bonus modification.

1. Probabilistic output selection.
2. The obedience multiplier was set to 6, the optimal value according to the results of PROP-REAS.
3. Memory registers.
4. An initial random start phase was used for animats in the first generation. Subsequent generations of animats within a trial did not have a random start phase.

Data from 454 tribes in 227 trials shows that tribes with higher signal and action differentiation factors (described in section 2.1) substantially outperform tribes with lower DFs, as shown in Table 23. As in the earlier experiments, within each trial a random subset of signals is enabled for each tribe (the enabled subset is different for each tribe and in each trial); any given signal has a 50% chance of being enabled for a particular tribe in a particular trial. As explained in section 4, each trial can be considered as a collection of 50 interdependent mini-trials (by animat turnover between generations), so the statistical relevance is actually higher than would be supposed just from the number of trials. By running hundreds of trials, statistically significant data can be accumulated and the performance of animats with varying capabilities can be analyzed by controlling for the enabling/disabling of signals and combinations of signals.

Table 23 -- Each cell in the table indicates the correlation between the DF in the column and the metric in the row. Some cells are duplicates, since ADF, SDF, and Score are in both rows and columns. ** indicates p-value less than 0.01.

| Correlation to: | Score | Action DF | Signal DF |
|---|---|---|---|
| Score | 1.00 | 0.70** | 0.61** |
| Action DF | 0.70** | 1.00 | 0.75** |
| Signal DF | 0.61** | 0.75** | 1.00 |
| Number of Enabled Signals (Total) | 0.17** | 0.17** | 0.44** |
| Number of Enabled Fetch Commands | -0.02 | -0.05 | 0.15** |
| Number of Enabled Brick Location Signals | 0.06 | 0.05 | 0.18** |

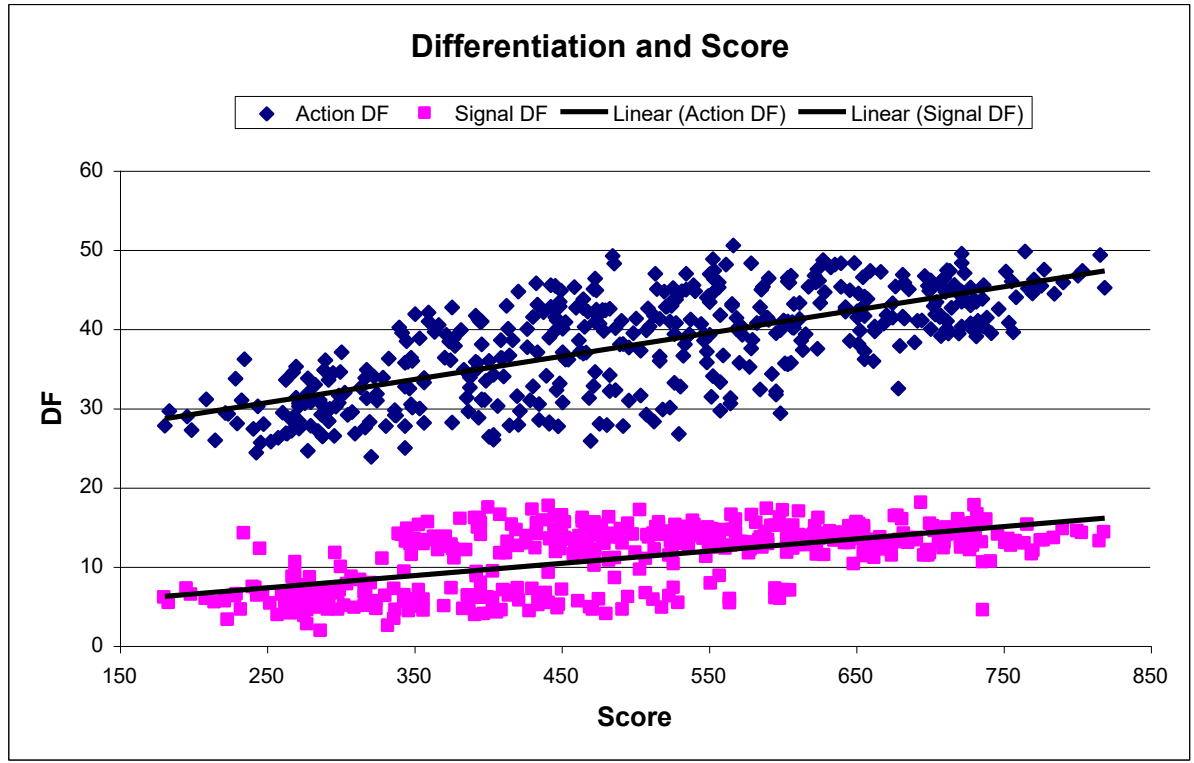

**Figure 10 -- This figure plots the DF values and scores of 454 tribes, overlaid by two trendlines.**

Several observations can be made:

1. Score correlates closely with both ADF and SDF.

2. The total number of enabled signals correlates closely with SDF, and less closely with ADF and score.

3. The number of enabled Fetch and Location signals has little effect on ADF or score.

4. ADF correlates closely with SDF.

The direction of causation is clear for observation (2), since the signals enabled in any given trial are determined before the trial begins -- the number of enabled signals influence differentiation and score. Observations (2) and (3) indicate that SDF is important for high scoring even when the number of enabled signals is low, since SDF correlates more closely to score than any of the enabled signals numbers. For observation (4), it is not known whether signal differentiation causes action differentiation or vice-versa, or neither, but since signals and actions are so closely intertwined (through commands, teaching, and information sharing) it is not surprising that their differentiation factors track each other closely.

The number of enabled signals in each group clearly causes SDF to increase and not vice-versa, but only the total number of enabled signals correlates with score, which implies that SDF has a lower effect on score than does ADF; most of the correlation between SDF and score is probably a side-effect of the high correlation between SDF and ADF. This makes sense because scoring in the brick-stacking game depends on performing actions in the proper order, and the signals are primarily used to command sequences of actions, thus the effect of SDF on score is indirect. However, since commands are given one at a time, the command to be issued depends less on the previous command generated than on the last observed action of the animat-to-be-commanded. Thus, SDF tracks ADF as command sequences track observed action sequences.

In order to see how final ADF and SDF values relate to score, it is useful to plot their intermediate values. Figure 11 and Figure 12 show the ADF, SDF, and score numbers for tribes 050523-002009-0036-1 (Tribe 36-1) and 050523-002009-0040-1 (Tribe 40-1), both of which beat their opponents by normalized scores of more than 700 to 300 (among the highest scores in the dataset).

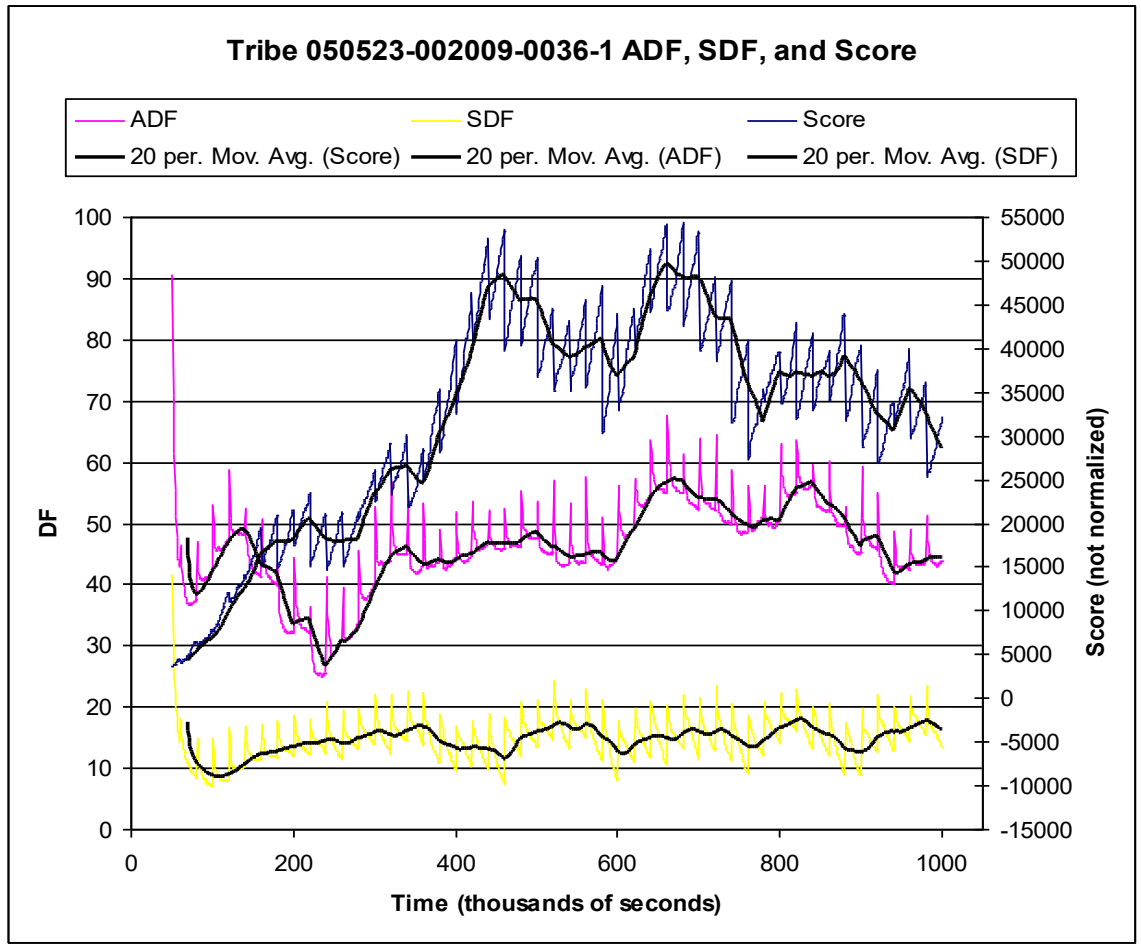

Figure 11 -- For Tribe 36-1, the correlation between ADF and Score is 0.50, and the correlation between SDF and score is 0.11. (Note that ADF and SDF use the Y-axis on the left, while the non-normalized Score uses the Y-axis on the right. The spikes in the plots occur every 20,000 seconds when the oldest animat is removed from the tribe and replaced with a new animat; score spikes down because the new animat has a score of zero, and ADF and SDF spike up because the new animat is untrained. The thicker black lines represent moving averages of 20 data points that smooth out the spikes.)

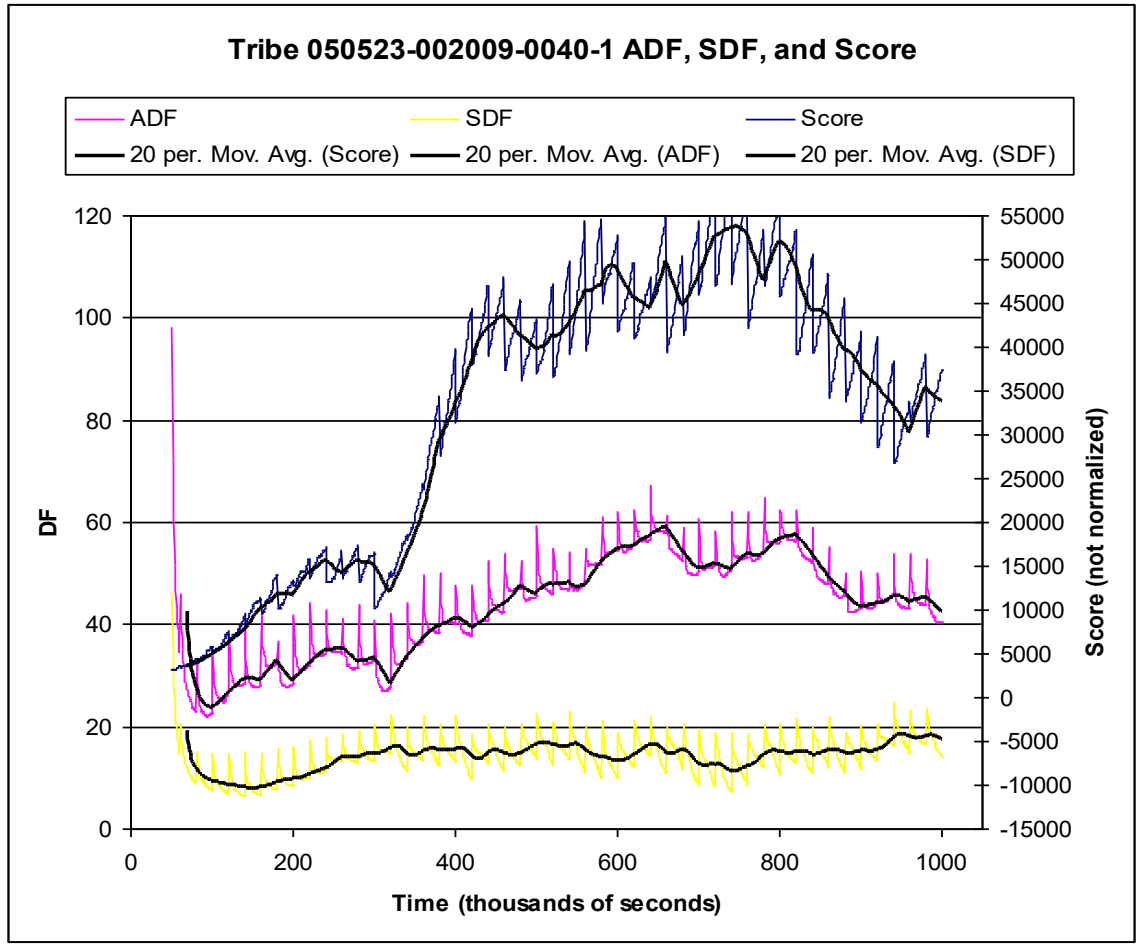

Figure 12 -- For Tribe 40-1, the correlation between ADF and Score is 0.84, and the correlation between SDF and score is 0.19. (Note that ADF and SDF use the Y-axis on the left, while the non-normalized Score uses the Y-axis on the right. The spikes in the plots occur every 20,000 seconds when the oldest animat is removed from the tribe and replaced with a new animat; score spikes down because the new animat has a score of zero, and ADF and SDF spike up because the new animat is untrained. The thicker black lines represent moving averages of 20 data points that smooth out the spikes.)

Although it is difficult to identify the actual characteristics of the amorphous roles that arise among the animats in a tribe, the differentiation factors prove they exist, and the correlation of DF to score demonstrates that differentiation is useful. (The direction of causation is known because score is entirely dependent on behavior.) Despite this difficulty, it is possible to look at snapshots of individual animats from the same tribe and hypothesize about how their recorded behaviors may fit in to a set of roles that contributes to their tribe's success.

  For example, consider the following six animats from tribe 050523-002009-0043-2 (Tribe 43-2). Animats 4997, 5002, 5004, 5006, 5008 and 5010 were consecutively created (except for Animat 4997), in the same tribe, and were all alive during the same period of time in the same trial, and by examining the data records we can see that each animat specialized in a different behavior. (The details of the animat numbering scheme aren't particularly important; in this case most of the odd-numbered animats were members of Tribe 43-1, the tribe opposing Tribe 43-2. These numbers are only for record-keeping purposes.)

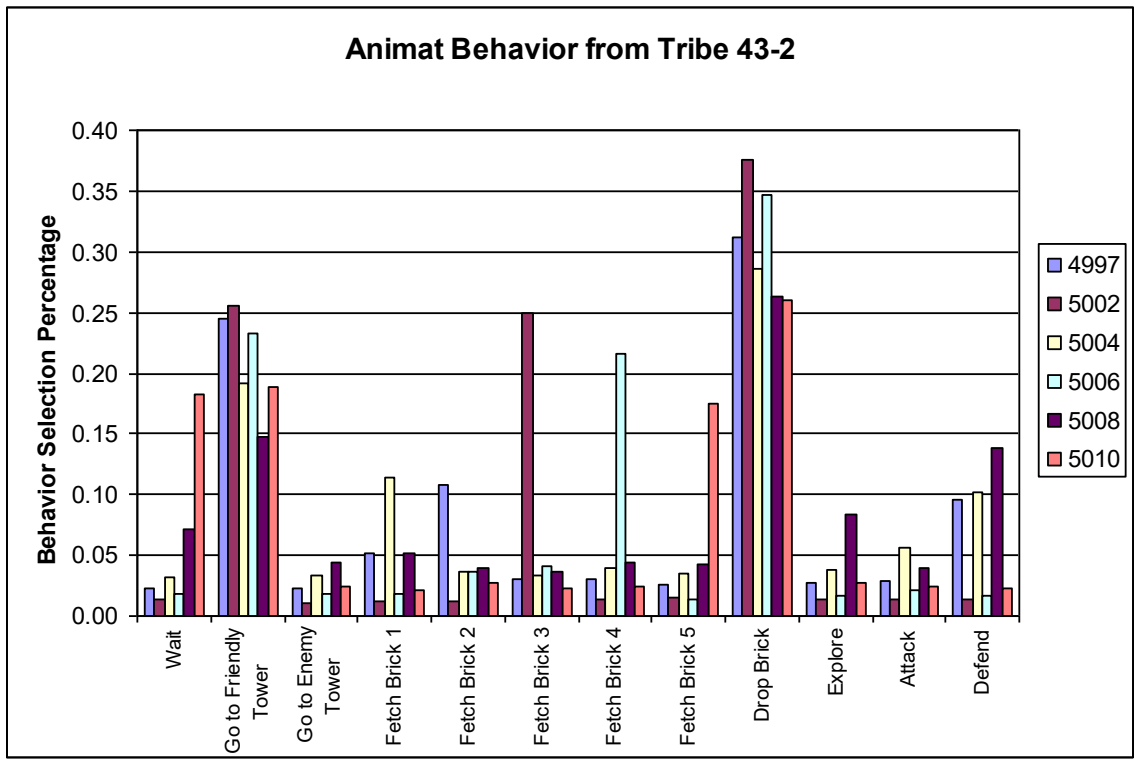

**Figure 13 -- This figure shows how often each animat chose each behavior.**

Figure 13 shows how often these six animats chose each behavior, and from these percentages we can discern the roles taken on by the animats. Most animats had high rates of returning to their tower and dropping bricks, so we should examine the other behaviors to determine their roles.

- Animat 4997: Fetching and returning brick-2; defending.
- Animat 5002: Fetching and returning brick-3.
- Animat 5004: Fetching and returning brick-1; defending and attacking.
- Animat 5006: Fetching and returning brick-4.

- Animat 5008: Exploring and defending; lower percentages for fetching bricks and returning to friendly tower than other animats.

- Animat 5010: Fetching and returning brick-5.

Among these six animats we have five that specialize in fetching bricks, one for each type, and another who explores and defends. Intuition tells us that this could make for an effective mixture of roles, and in fact during the time period these animats lived together Tribe 43-2 outscored Tribe 43-1 by 145 to 38 -- a ratio of 3.8:1, compared to a ratio of 2.3:1 for the trial as a whole. In Table 23 I showed that score correlated closely with ADF and SDF and these two tribes reflect that correlation, as seen in Table 24, below.

Table 24 -- This chart gives a comparison of the scores, differentiation factors and number of enabled signals of Tribes 43-1 and 43-2. Even with very similar numbers of enabled signals, Tribe 43-2 beat Tribe 43-1 because of its superior role differentiation.

|  | Tribe 43-1 | Tribe 43-2 |
|---|---|---|
| Score | 269 | 731 |
| Action DF | 28.43 | 39.75 |
| Signal DF | 6.63 | 17.63 |
| Total Number of Enabled Signals | 10 | 10 |
| Number of Enabled Fetch Commands | 2 | 3 |
| Number of Enabled Brick Location Signals | 3 | 3 |

Finally, Table 25 reveals three other results:

1. Tribes tend to differentiate approximately 350% more in a competitive environment than when they live alone, no matter what their social structure. Competitive environments have relatively sparser resources, and differentiation appears to be more useful when tribes have to coordinate offense and defense.

2. Tribes with no authority (NA), consisting of animats that cannot command each other, have lower differentiation factors than communicating tribes, particularly in competitive scenarios.
3. Tribes with age authority (AH) had almost the lowest DFs in solo trials, but they had the highest DFs in competitive trials.

Table 25 -- Comparison of differentiation factors for tribes with various social structures in solo and competitive environments.

| Social Structure | Solo Average Action DF | Solo Average Signal DF | Competitive Average Action DF | Competitive Average Signal DF |
| --- | --- | --- | --- | --- |
| AH | 10.95 | 2.67 | 39.31 | 11.55 |
| MH | 12.02 | 2.74 | 36.10 | 10.80 |
| NA | 12.62 | 1.75 | 32.62 | 5.69 |
| NH | 12.28 | 3.33 | 39.25 | 11.01 |

## *4.4 Miscellaneous Experiments*

In addition to the major experiments described above, I ran several smaller experiments and varied the brick-stacking task to demonstrate that the quality of the performance and results I recorded was not heavily dependent on the specifics of the task. These task modifications did affect the behavior of the tribes and their animats, but the differences follow logically from the alterations. All of these variations were tested competitively with two identical tribes using the Age Hierarchy social structure, and the results were compared with the results from the AHvAH trials in section 4.2 (which will be referred to as the *standard AHvAH* trials). In the interests of time, only fifteen trials were run for each of these experiments, which was not enough to generate data within the 95% confidence interval; however, most of the data would fall within a 90% confidence interval.

Four variations on the brick-stacking task were tested:

1. *Unordered stacking (US)*. Animats were allowed to stack bricks in any order rather than being restricted to numerical order.
2. *Spotters and fetchers (SF)*. Each animat is created as either a *spotter* or a *fetcher*. Spotters cannot pick up bricks and fetchers cannot see bricks but must be told where they are by the spotters.
3. *Signal interception (SI)*. Animats can hear and understand the brick location signals of enemies within their signal radius.
4. Trials were also run with five animats per tribe (5APT) and twenty animats per tribe (20ATP).

### 4.4.1 Unordered stacking

In US, animats were allowed to stack bricks onto their tower in any order rather than being required to stack numerically. When compared to the standard AHvAH trials, the US modification yielded several differences, as shown in Table 26.

Table 26 -- Side-by-side comparison of performance data between the standard AHvAH trials and the US trials.  Note that for US the correlations between DF and Score are not within the 95% confidence interval.

|  | Standard AHvAH | US |
| --- | --- | --- |
| Standard deviation of normalized score | 132 | 51 |
| Average ADF | 38 | 52 |
| Average SDF | 12 | 7 |
| Correlation of ADF to score | 0.58** | -0.13 |
| Correlation of SDF to score | 0.59** | 0.36 |

US tribes had a lower standard deviation for their scores and lower correlations between DF and score than the standard AHvAH tribes because the US task is less complex than the standard task.  US tribes benefited less from differentiation because the bricks could be fetched and dropped onto the tower in any order, and it was difficult for any tribe to beat its opponent by a large margin because no strategy was much better than any other.  It appears that high action differentiation may even *harm* US tribes (though the correlation is not statistically significant), possibly because the only differentiation possible is away from fetching and dropping altogether, which are the major scoring behaviors.

### 4.4.2  Spotters and fetchers

Animats in SF tribes were divided into spotters and fetchers randomly upon creation in a 1:1 ratio.  Spotters cannot pick up bricks and fetchers cannot see bricks but must be told where they are by the spotters.  Table 27 shows how some of the relevant data compares between SF and the standard AHvAH trials.

Table 27 -- Side-by-side comparison of performance data between the standard AHvAH trials and the SF trials.  Note that SF correlations are not within the 95% confidence interval.

|  | **Standard AHvAH** | **SF** |
|---|---|---|
| Average ADF | 38 | 76 |
| Standard deviation of ADF | 6.3 | 3.2 |
| Average SDF | 12 | 7 |
| Standard deviation of SDF | 3.7 | 1.1 |
| Correlation of total number of enabled signals to score | 0.24** | 0.01 |
| Correlation of number of enabled brick fetch signals to score | 0.11** | -0.01 |
| Correlation of number of enabled brick location signals to score | 0.11** | 0.17 |

The ADF for SF tribes is higher than the ADF for standard tribes because of the imposed bifurcation between spotters and fetchers.  Spotters tended to attack and defend more than fetchers because those were the only behaviors available to them that were rewarded by a score increase; fetchers fetched bricks and put them on the tower.  The low standard deviation of ADF is also due to the imposed bifurcation, since unlike standard tribes there was no chance that a SF tribe wouldn't be split into roles.  Similarly, the average SDF for SF tribes was lower than for standard tribes because of the increased importance of the brick location signals – animats used them almost exclusively.  Note also that although the total number of enabled signals and the number of enabled brick fetch signals had little effect on the SF tribes, having a higher number of enabled brick location signals had a measurable effect on score.  Since fetchers couldn't score without information from the spotters, the brick location signals were far more important than in the standard trials.

### 4.4.3 Signal interception

In SI trials animats could overhear brick location signals transmitted by enemy animats within signal range. Standard values for signal range (512 units) and sight range (256 units) were used in this experiment, which means that a signaling animat will not know for certain whether or not its signal will be overheard because the total area within signaling range is four times as large as the area within sight range. Table 28 shows how the ability to overhear location signals affected performance.

Table 28 -- Side-by-side comparison of performance data between the standard AHvAH trials and the SI trials. Note that SI correlations are not within the 95% confidence interval.

|  | **Standard AHvAH** | **SI** |
|---|---|---|
| Average ADF | 38 | 32 |
| Standard deviation of ADF | 6.3 | 2.8 |
| Average SDF | 12 | 6 |
| Standard deviation of SDF | 3.7 | 1.3 |
| Correlation of total number of enabled signals to score | 0.24** | 0.21 |
| Correlation of number of enabled brick fetch signals to score | 0.11** | -0.03 |
| Correlation of number of enabled brick location signals to score | 0.11** | -0.17 |

The most interesting result is that the correlation between the number of enabled brick location signals and score is negative, which is very intuitive. Given the ratio between the sizes of the areas covered by signals and sight, uttering a signal that can be overheard and understood is dangerous because it is very likely to help unseen enemies. The average SDF for SI tribes was low because animats learned to avoid using brick location signals, which gave them fewer signals to choose from. However, this was a hard habit to learn, so tribes with many enabled brick location signals were at a significant disadvantage. Since scoring is only indirectly related to the use of these the brick location signals, it was difficult for the SI tribes to realize that using brick location signals helped the enemy at least as much as it helped themselves. Overall, however, the effects of allowing information signal interception were fairly slight, which is roughly in line with the results of (Reggia, Schulz, Wilkinson and Uriagereka, 2001) who found that introducing a modest number of eavesdropping predators into their food gathering simulation only reduced the number of signaling animats from 100% to 95%. NEC-DAC animats can all overhear the signals of their enemies and discern their source, but SI animats could also glean knowledge about brick locations from the signals they intercept, which gave them an advantage roughly equivalent to that of the Reggia et al.'s predators by telling them the locations of their scoring objectives (the bricks rather than the animats).

### 4.4.4 Five and twenty animats per tribe

To test how the number of animats affected behavior I ran two more sets of trials, one with tribes of five animats and one with tribes of twenty animats. Neither of these experiments resulted in any significant difference from the results of the standard AHvAH trials, although DF measurements were slightly lower in both cases. Trials with fewer than five animats per tribe are too sparse for signaling to be very useful, and trials with more than twenty animats per tribe take an extremely long time to execute.

## 4.5 Neural Network Weights

Rather than describing the results of a specific set of experiments, this section illustrates the neural network weights of a typical animat. Here are four charts representing the connection weights from the neural networks of one average-performing animat.

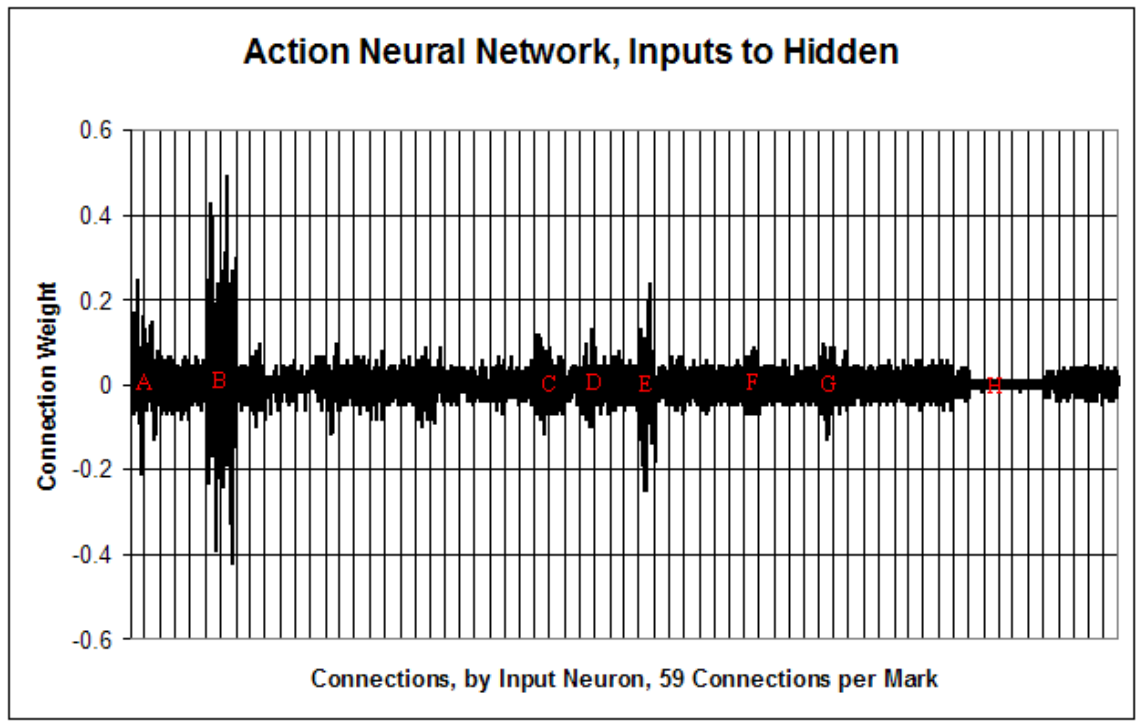

**Figure 14 -- The weights of the input-to-hidden connections in the ANN of average-performing Animat 72 (051119-222639-0001-2-72) at time of death. The regions between the vertical lines represent the set of connections leaving one input and going to the 59 hidden neurons.**

In Figure 14 we see eight areas of interest, marked by capital letters in the figure, that have connection weights leaving the inputs that are of significantly different magnitude than the other connections. The inputs associated with these areas are:

A. At Friendly Tower and See Friendly Tower inputs

B. Have Brick and Have Right Size Brick inputs

C. Last action was Go To Friendly Tower

D. Last action was Fetch Brick-2

E. Last action was Drop Brick

F. Last command was Go To Enemy Tower

G. Last command was Fetch Brick-5

H. Inputs from memory registers that take their values from the ANN

These results are of typical magnitude for all but the lowest-performing animats (which have very flat connection weight graphs) but the significant and insignificant areas vary between animats. Areas A and B indicate the four sensory inputs that have the most effect on behavior for Animat 72, and C, D, E, F, and G shows how Animat 72 values the inputs fed by the outputs of its own neural networks (i.e., E indicates that the knowledge that the last action Animat 72 performed was Drop Brick is particularly important). Area H shows that the weights of the connections leaving the inputs linked to the ANN memory registers are not very important, and this was the case with most animats. The inputs from the SNN memory registers (the area after H) were more active.

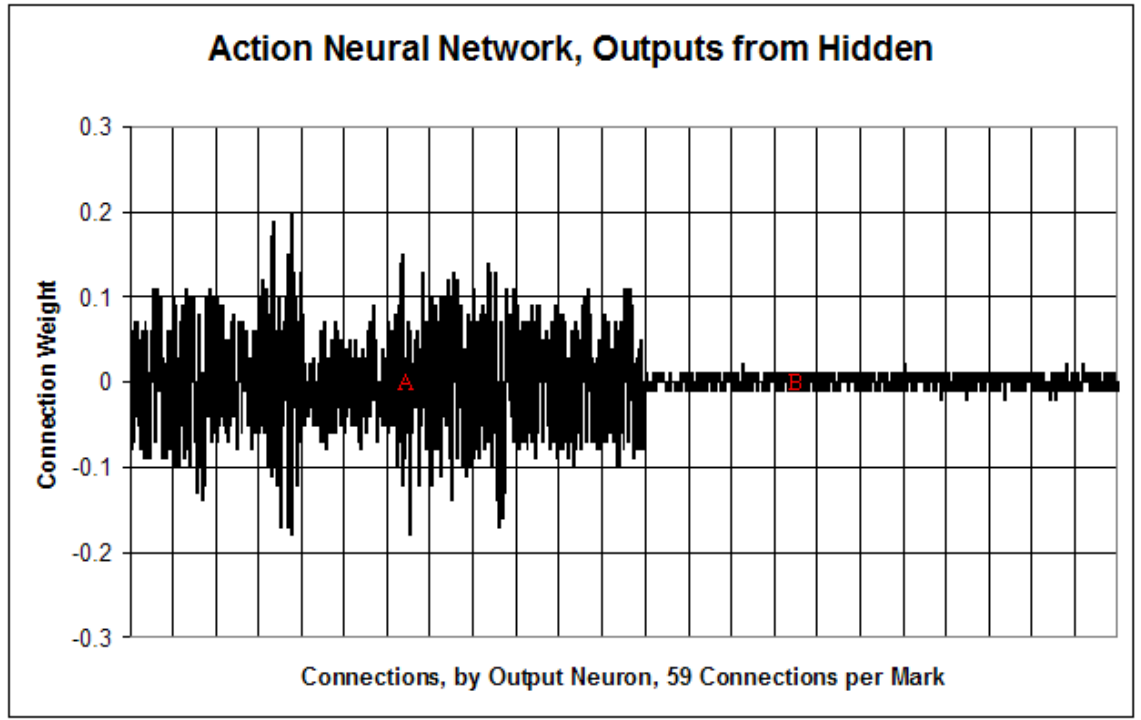

**Figure 15 -- The weights of the hidden-to-output connections arranged by output in the ANN of average-performing Animat 72 (051119-222639-0001-2-72) at time of death. The regions between the vertical lines represent the set of connections entering one output from the 59 hidden neurons.**

In Figure 15 we see two areas:

A. The outputs associated with the 12 behaviors

B. Six unused outputs followed by the five outputs that feed into the ANN memory registers

We can see that the connections that feed into the outputs associated with the 12 behaviors in area A are very diverse, but the connections to the five outputs that feed into the ANN memory registers in area B are nearly as flat as the connections to the unused outputs, indicating that the ANN memory registers are not very important.

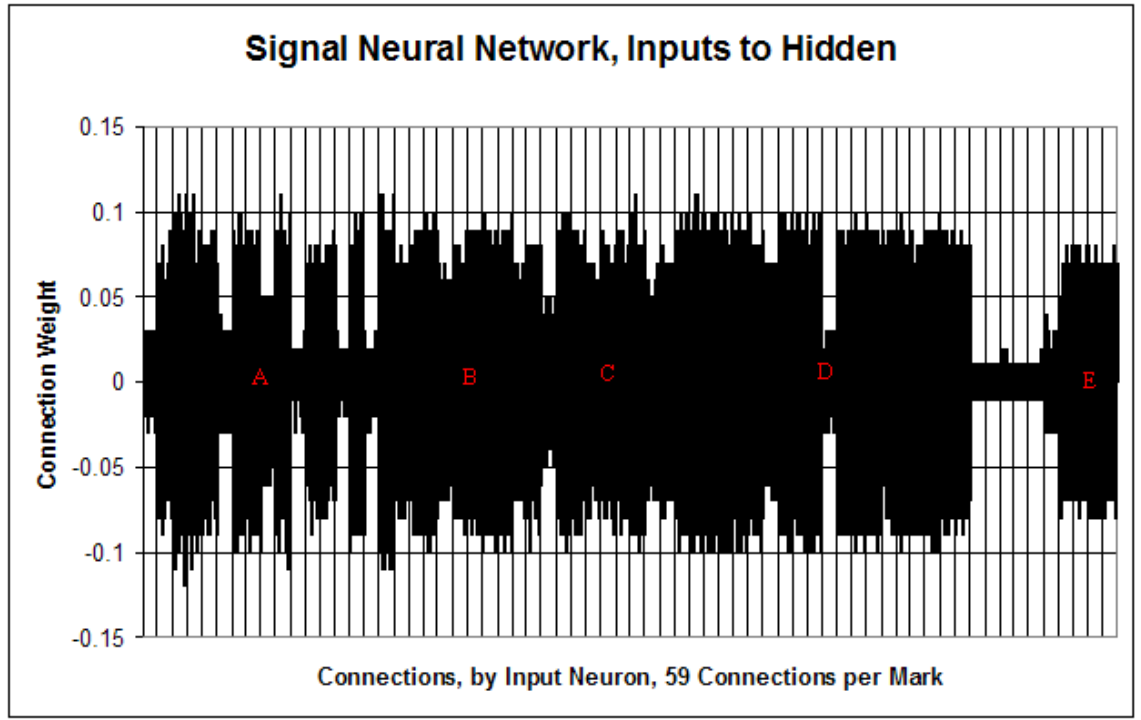

**Figure 16 -- The weights of the input-to-hidden connections in the SNN of average-performing Animat 72 (051119-222639-0001-2-72) at time of death. The regions between the vertical lines represent the set of connections leaving one input and going to the 59 hidden neurons.**

In Figure 16 we see that all areas of the input vector contain inputs that are significant to the decision of the SNN.

A. Sensory inputs

B. Behavior of last observed friendly animat

C. Last behavior of this animat

D. Last signal of this animat (the low point is the input that's triggered when the last signal uttered was Command to Fetch Brick-4)

E. The inputs associated with the SNN memory registers

In the flat spot right before Area E we see that the inputs associated with the ANN memory registers are, again, not significant, and it appears that only the SNN memory registers are getting much use.

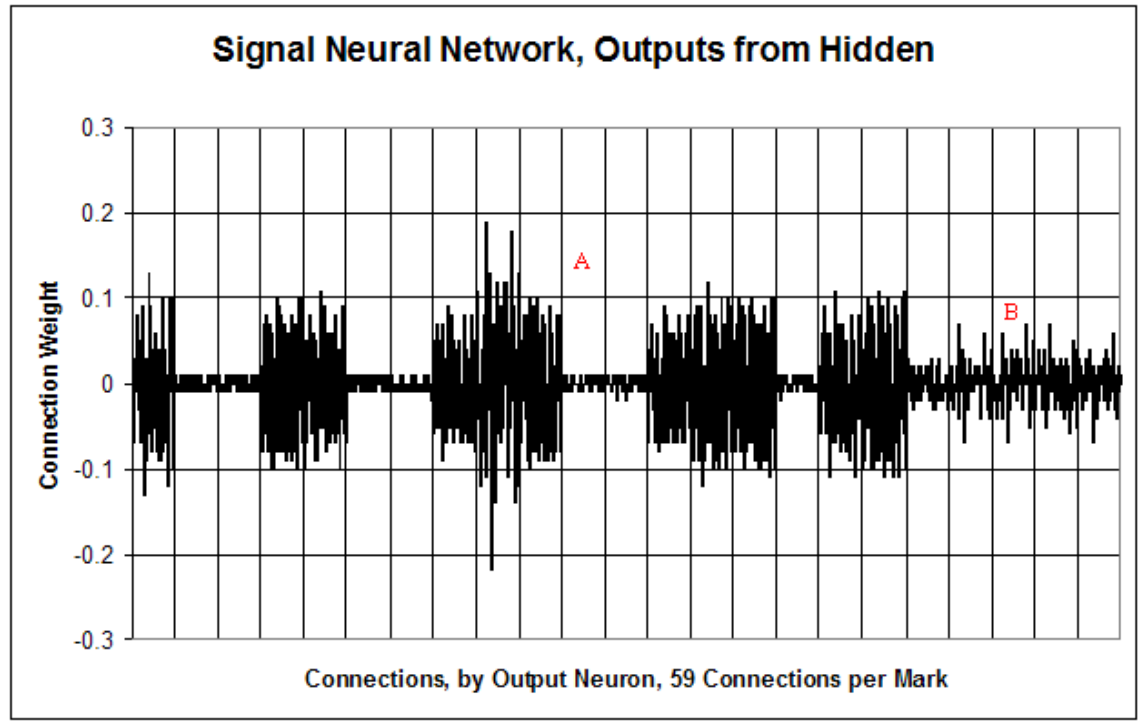

**Figure 17 -- The weights of the hidden-to-output connections arranged by output in the SNN of average-performing Animat 72 (051119-222639-0001-2-72) at time of death. The regions between the vertical lines represent the set of connections entering one output from the 59 hidden neurons.**

There are two main areas shown in Figure 17:

A. The 18 outputs that determine the signal to be uttered

B. The five outputs that feed into the SNN memory registers

The flat regions in Area A indicate signals that were disabled for Animat 72's tribe in this trial. Note that the connections into the memory register outputs in Area B are significantly more varied that the connections into the ANN memory registers in Figure 15, which were completely flat. This activity indicates that the SNN memory registers are far more significant than the ANN memory registers.

# 5  Conclusions and Related Work

With the rise of ubiquitous computing and the focus on netcentricity, it is inevitable that agent-based software solutions will become more and more prevalent. Netcentricity demands that network users be able to extract useful information from huge streams of raw data and adapt to rapidly changing circumstances, and with computers quickly outnumbering humans there's no question but that they will eventually be required to perform at high levels with minimal human intervention. These agents, whether robotic or purely software, will have to function within -- and eventually *create* -- their own social structures, and this dissertation describes a first step in that direction. The architectural features and social structures I explored can be used as components in the new science of social robotics and agent-based computing, and NEC-DAC can also be adapted to simulate agents in human social structures with real-life tasks and capabilities.

Three main sets of conclusions can be drawn from the three experiments detailed above:

1. Probabilistic architectural features can contribute to animat learning, particularly the learning of interrelated sets of behaviors and signals.
2. Simple social structures have a significant effect on both the creation and evaluation of signaling schemes.
3. Advantageous role differentiation can arise spontaneously as the result of learning combined with scoring pressure.

I also discuss in this section how my conclusions relate to other work in the field and conclude with a section describing potentially useful future work.

## 5.1  Probabilistic Architecture

Most of the probabilistic modifications to NEC-DAC yielded a performance improvement, and most of them could also be trained towards deterministic behavior by the neural networks. That is, for any specific animat, if it was not advantageous to behave probabilistically then its neural network could learn to push the percentages towards 0 or 100, as appropriate.

Of the five modifications:

1. Probabilistic output selection beat winner-takes-all output selection in 100% of the trials.
2. Output bonuses had negligible effect.
3. Various obedience multipliers were tested, and the data shows that middling values around 6 outperform low or high/infinite values.
4. Tribes with memory registers beat those without in 94% of trials.
5. A random start phase can provide a valuable leg up for a new animat, even to the extent of rendering teaching and learning signals useless. However, since a random start phase is not realistic, it is reasonable to favor teaching and learning.

An animat can reduce its probabilistic output selection and the effects of output bonuses by training its neural weights; memory registers can be ignored. The obedience multiplier, however, was predetermined for each trial; multiple values had to be explicitly tested to establish the usefulness of the modification, and some values were clearly more beneficial than others.  A future design might incorporate a mechanism to enable the animats to modify their own obedience multipliers as they weigh the effectiveness of the commands they receive -- animats could even have different multipliers for each of their compatriots, depending on their perceived reliability.

The random start phase modification was different from the others; despite its usefulness, I decided to scale it back because it interferes with the teaching and learning signals that are critical to my experiments.  Consequence-free random start phases are unrealistic in the natural world.  In circumstances where parents protect children to the degree that random exploration does not result in death, parents and children also engage in teaching and learning activities, either actively or passively, such that the randomness of exploration is limited.  Replacing this random start with an adaptive curiosity approach (Schmidhuber, 1991) could lead to a more natural simulation of the early phases of a real organism's life.  Such an approach would enable an animat to guide its explorations based on the predicted results of its actions rather than the feedback from a random walk, and it's possible that the animat would then be better able to optimize its behavior later in life. By making predictions and then purposefully seeking out behaviors that cannot be predicted with current knowledge, the time currently allocated to the random start phase could be used more efficiently and the problem space could be explored more thoroughly.

The advantage of probabilistic mechanisms, as opposed to greedy winner-takes-all, is that an input vector that yields a 51% decision in one direction won't prevent other directions from being explored.  This feature is particularly important when there are multiple beneficial directions to choose from.  For instance, choice A may always give a reward of 5 while choice B gives 0 or 10, depending on the circumstances. An animat with winner-takes-all decision making could easily get stuck choosing A if it never encounters a beneficial B situation before it gets set in its ways. However, with a probabilistic architecture, even if a neural network is trained 90% towards A, B will still be tried occasionally. This functionality may not be valuable in circumstances with well-defined input sets and repeated training sessions that expose neural networks to the full range of possible input vectors, but it can be advantageous in more open-ended, constantly changing systems.

(Zhao and Schmidhuber, 1996) deal with a similar situation in which they want to use reinforcement learning to train animats without using heuristics and without repeatable trials. They use their incremental self-improvement (IS) method to train animats in an environment that is constantly changing as the animats in it learn and adapt to each other's behavior. They describe IS as "co-evolutionary" in the sense that the animats adapt to each other, but since there is no "evolution" across generations I prefer the term *co-learning*. NEC-DAC similarly consists of co-learning cooperative/competitive animats, and the history queues I use for training are similar to the history used by IS. However, IS maintains a complete history of each animat's life from creation up to the training "checkpoint" whereas NEC-DAC history queues only store the last 100 decisions and their context. IS then considers the animat's history as a set of regions between checkpoints and evaluates whether or not the modifications made at each checkpoint improved the reward rate of the animat in subsequent regions. Any modifications that are judged to be harmful are later "undone", thus guaranteeing that their "reinforcement acceleration criterion" is satisfied -- that is, that the reward rate stays the same or increases over the life of the animat. Their principle method for dealing with an environmental change that renders their current policy disadvantageous is to backtrack and undo recent modifications until the reward rate begins increasing again. In contrast, NEC-DAC's probabilistic neural network approach continually incorporates changes to its existing policy by modifying neural network weights and by refusing to commit to any specific pattern of behaviors no matter how advantageous it appears at the moment. Since the NEC-DAC's neural networks only generate probabilities rather than absolute decisions, no avenue is ever left unexplored and when the reward rate for one behavior path begins to increase the weights in the neural network will shift to favor that path. In essence, the neural weights fulfill the same purpose as the life history of the animats in IS by encoding all the decision-process history of the animat. Just as IS backtracks and undoes modifications judged to be harmful, NEC-DAC's neural weights are constantly adjusted to increase the probability of beneficial behaviors and reduce the probability of harmful behaviors.

### *5.2  Social Structures*

Imposing social structures on tribes in the form of command authority significantly affected animat performance and influenced how signals were used.

1. Tribes with no authority (NA), in which animats could not command each other, fared the worst, since no teaching took place.
2. Tribes with authority but no hierarchy (NH) performed slightly better, but learning was confounded because the commands from young, inexperienced animats were given the same weight as those from older, experienced animats.

3. Tribes with a merit-based hierarchy (MH), in which animats obey those with higher scores, surprisingly performed *worse* than NH tribes because an animat with a high score is not necessarily a better teacher than an older animat with a lower score.
4. Tribes using an age-based hierarchy (AH) or an age and merit hierarchy (AM) performed similarly, with a slight advantage going to AH because it didn't exclude commands from animats with seniority but low scores.

The difference between AH and MH shows that it can be useful to distribute decision-making across multiple neural networks. The outputs of the Action Neural Network were the primary determinants of score, and the outputs of the Signal Neural Network were used to command and train the ANNs of other animats, thereby improving the performance of the tribe as a whole across generations. The two neural networks had different purposes and worked in tandem to optimize the tribal culture. (Cangelosi and Harnad, 2002) showed similarly that what they called "symbolic theft" (learning a behavior from another, such as the learning done by the SNN) was significantly more efficient than "sensorimotor toil" (learning a behavior for yourself, such as the learning done by the ANN). Cangelosi and Harnad argue that the most fundamental advantage of language is that it enables its users to bypass the otherwise necessary trial-and-error stage of learning a behavior and jump straight to the benefits of that behavior, and the results of SOC-STRUCT bear out that result. The different social structures serve as sorting functions that ensure that younger animats learn behaviors from the "right" sources, and all except NA implement a form of symbolic theft.

Although the definition of "right" is ambiguous and necessarily depends on the circumstances in question, (Marocco, Angelo and Nolfi, 2002) demonstrated that not only are some social structures are better than others, but that some social structures can actually prevent the formation of communication. Although NEC-DAC sidesteps this potential difficulty by beginning with exogenous signals, it is nonetheless interesting to note that the successful social structure identified by Marocco et al. is somewhat similar to AH, the best social structure tested in NEC-DAC. In their experiment with communicating robots, Marocco et al. defined two social structures they called "PARENT" and "ALL". In the PARENT structure, a robot receives communication signals only from its own parent, and in the ALL structure a robot receives signals from every member of the previous generation. Although the comparison is inexact, the PARENT structure is similar to AH and the ALL structure is similar to the No Hierarchy (NH) structure in which every animat takes commands from every other. In both PARENT and AH agents only listen to their elders, whereas in ALL and NH agents listen to everyone who is speaking. Marocco et al. judged communication to have successfully evolved when at least 50% of the agents in the final generation of their simulation produced two signals that differentiated between two objects, and that only ever happened when the PARENT structure was in use.

(Buzing, Eiben and Schut, 2005) described similar, though more limited, experiments with exogenous social structures in a world called VUSCAPE that is based on SugarScape (Epstein & Axtell, 1996). Whereas most research on the emergence of language focuses on the evolution of syntax and semantics, they (as I) chose to look instead at how existing communication capabilities can be used for cooperation.

Traditional SugarScape is a exploration/scavenging model commonly-used by social scientists for simulating societies that Buzing et al. augmented by adding a simple "call for assistance" signal that an animat can utter when it encounters a pile of sugar that's too large for it to consume on its own.  The authors vary a parameter that controls the amount of sugar an animat can eat in one sitting and thereby increase or decrease the "cooperation pressure" on the animats as a whole to cooperate -- i.e., animats that can eat less in one sitting need to find a larger number of piles and suffer less from sharing the piles they find.  As with NEC-DAC their structure of cooperation is built into the system, but their VUSCAPE system is based on evolutionary development and their animats do not decide when to use their single signal.  Instead, the animats are equipped with a "talkativeness" gene that causes them to utter their signal with a random probability when they are in the vicinity of a sugar pile.  On the other side of the communication equation, listeners in NEC-DAC have considerably more flexibility than listeners in VUSCAPE.  Similar to their "talkativeness" gene, the VUSCAPE animats have a "listening" gene that determines the random probability that the animat will heed any "call for assistance" signal that it hears.  As they increase the cooperation pressure, their animats evolve greater talking and listening probabilities.  NEC-DAC extends the exploration of the use of signals for cooperation by introducing not only a substantially greater number of signals but also several different conceptions of cooperation.  A NEC-DAC animat uses its Signal Neural Network to decide when to utter which signal and its Action Neural Network to set the base probability for a given behavior, which is then modified by its obedience multiplier.  This method allows a NEC-DAC animat to dynamically influence when it will obey the commands it receives and makes for a far more complex set of interactions. By using a learning simulation such as NEC-DAC rather than an evolutionary simulation, social scientists could design more dynamic worlds that could support the more complex communication schemes required for more human-like interactions. Economists (Tesfatsion, 2002), for instance, are growing particularly interested in modeling economic systems on the micro scale, as face-to-face relationships between individual agents rather than as macrostructures of "fixed decision rules, common knowledge assumptions, representative agents, and imposed market equilibrium constraints" with micro interactions limited to tightly constrained game theory representations.

The supremacy of the Age Hierarchy (AH) and the Age and Merit Hierarchy (AM) structures over the Merit Hierarchy (MH) structure is interesting because it implies that it is beneficial for young, high-scoring animats to take orders from older, lower-scoring animats. This is odd, because it means that AH tribes are reaping a corporate benefit that is not reflected in the scores of the individuals responsible for generating that benefit. However, the result is explained by the feedback signals described in section 3.2.5. Animats' SNNs are purposefully reinforced when they utter commands that benefit their listeners. What may at first appear to be an example of altruistic communication is in fact simply the result of a second, internal scoring mechanism operating in parallel to the primary scoring mechanism. Without this internal reward, current anthropology research suggests that altruistic cooperation cannot be sustained in the long-term, even when punishment is used enforce proper altruistic behavior (Boyd, Gintis, Bowles and Richerson, 2003).

In solo trials, AH tribes reached the target score about 10% faster than the other social structures, which all performed similarly. Competition, and the attendant complexity, was a major motivation for developing culture, at least for the NEC-DAC experiments, and the various social structures evidently had little effect on performance in the absence of that additional pressure. It is not clear if the specific type of complexity introduced (competition) had more of an effect than could have other forms of complexity.

Future research should investigate dynamic social structures and would go hand-in-hand with my suggestion above to enable animats to assign obedience multipliers to individuals based on experience. It could then be studied whether or not score correlates with imputed authority in a purely voluntary social structure. Social structures based on communication paradigms other than audible sound should also be tested, such as various computer networking technologies (message passing) and bulletin board systems (shared memory). NEC-DAC animats cannot uniquely identify whom they are signaling with, and this design decision could cause scaling problems if the system were using a communication paradigm that put all animats into communication range simultaneously (e.g., over the Internet).

Investigation into non-voluntary obedience could also be interesting, if for instance animats were able to issue a command, check if it is obeyed, and then punish a disobedient listener. NEC-DAC doesn't allow combat between animats in the same tribe, but animats do theoretically have the power to determine whether or not their commands are obeyed because the two neural networks can pass information using the memory registers. Tribes themselves could also be made endogenous and opened up to voluntary membership (Odell, Parunak and Fleischer, 2003). Additionally, multi-dimensional scoring metrics might be useful for evaluating performance, particularly when multiple neural networks are involved and attempting to optimize different domains.

Incorporating additional signals and a more advanced communication paradigm would bring the simulation closer to using language rather than mere signaling. More advanced communication should include:

- grounding,
- creativity,
- arbitrariness,
- phrase structure,
- negation,
- recursive semantics,
- syntax.

Much research has gone into studying how these features of language emerge through learning or evolution, but we shouldn't shy away from investigating how these features are used for cooperation once they exist. In a sense I'm suggesting a more top-down approach to language and cooperation in which these components are engineered symbolically and then put into use by a neural network. Similar to my animats in NEC-DAC, most humans don't consider the underlying mechanism that enables them to create new sentences or to learn vocabulary; they are concerned more with how they can best use their language abilities to cooperate with others and to achieve their goals.

## *5.3 Role Differentiation*

There are three key conclusions to draw from the ROLE-DIFF experiment:

1. Roles can arise spontaneously within a population of animats with identical capabilities.

2. Differentiation can give a group a competitive advantage.

3. Spontaneous differentiation is much less pronounced in non-competitive environments.

All of the animats in all my experiments possess the same capabilities, unprejudiced towards any artificial roles, and tribes are not explicitly rewarded for promoting differentiation. Nevertheless, significant differentiation did occur across my experiments as a result of animats independently assigning themselves behavioral niches. In many cases the roles created by the differentiation could not be intuitively characterized, but the differentiation factor metric provides a mathematical way to objectively measure the differences between the behaviors of multiple animats within a tribe. Whether the roles that arise within a given tribe can be anthropomorphized or not, ROLE-DIFF demonstrates that differentiation contributes to high scoring, particularly in competitive trials.

Animats receive feedback from two sources: the effects of their own behavior and the effects of their commands on their listeners. Role differentiation results from a combination of these two factors. An animat might take on a role for itself by making choices that are profitable given pre-existing behavioral tendencies of the other animats in its tribe. In this case, the role is "discovered" without a need for signals when the animat is rewarded for falling into a niche created by a vacancy left by the behaviors of other friendly animats. For instance, if (for whatever reason) no animats in a tribe ever collect brick-5, a new animat that is initialized with an equal affinity for all bricks will quickly find brick-5 to be the most rewarding. However, ROLE-DIFF demonstrates that tribes with authority hierarchies have Action Differentiation Factors 20% higher on average than tribes without authority hierarchies, which indicates that some significant portion of differentiation is due to the use of signals. NEC-DAC is not capable of determining what percentage of differentiation is due to the use of command signals because it is not possible to run an experiment in which animats cannot "discover" niches as described above, but isolating these two factors might be a fruitful direction for future work.

NEC-DAC animats are identical when created except for their random neural network weights, and the roles that they assume over their lives arise endogenously from their experience in their environment, but there are other approaches to role assignment. For example, (Bonabeau, Dorigo and Theraulaz, 1999) provoke exogenous role assignment in their ant-like animats through the use of *response thresholds* set for each animat that determine the likelihood that a given animat will react to a task-associated stimuli. For example, in their simulation an animat with a low response threshold for hunger will go gather food for its colony before another animat with a higher hunger response threshold will. In this way, an animat is stimulated to perform a task based on how the level of perceived need compares with its response threshold for that task. These behaviors are reinforced into roles over time because when an animat works at a task its response threshold for that task is decreased -- making it more likely that the animat will take on that task in the future -- and its other response thresholds are increased, leading to specialization. Roles assignment is not rigid because if a specialized animat dies the stimulus attached to its task will increase as the task remains unsatisfied, and the stimulus will eventually cross the threshold of another animat that will then perform the task. Bonabeau et al. demonstrate strong specialization even when animat response thresholds are initialized identically, but because the roles are predefined the differentiation is still exogenous. They also show that response thresholds can lead to coordinated activities. For instance, a dirty nest may stimulate some animats to gather trash into a pile, and the pile of trash may then stimulate some animats to haul the pile of trash out of the nest. Response thresholds can give the impression that the animats are purposefully performing the tasks in sequence even though there is no intentionality. The response threshold approach is quite efficient for allocating workers to single-behavior tasks such as may be found in an artificial ant colony, but it does not provide a mechanism for creating or adapting to new tasks or for chaining behaviors together without intermediate stimuli or rewards. In contrast, each role in NEC-DAC consists of several behaviors used in sequence without intermediate rewards, and each role is identified (or not) anew in each trial. The animats decide how to accomplish their main task of stacking bricks by identifying subtasks (fetch, move, drop) and then combining behaviors into sequences. Over time preferential uses of these behavior sequences coalesce into roles. Even though NEC-DAC has a built-in reward for dropping the correct brick onto the tower, the way that intermediate behaviors were combined into roles was not defined before experimentation, which is why NEC-DACs roles are endogenous.

      As with the social structures, differentiation also appears to be less important in non-competitive environments. My experimental setup makes it difficult to isolate

complexity from competition to determine whether or not competition increases differentiation simply because of the complexity it adds or because of the competition itself. If complexity could be quantified it would be possible to construct a solo scenario with the same complexity as a competitive scenario and then compare differentiation, but that does not appear to be possible within the scope of NEC-DAC. Considering complexity and competition together, competition increases complexity in several ways:

1. Greater scarcity of resources because:
    a. there are more animats carrying bricks,
    b. there is an additional tower that can keep out of circulation up to four bricks at a time.
2. Larger behavior/signal space because attack and defense options become useful.
3. More complex input data because the presence of enemy animats and the enemy tower activate several inputs that are unused in solo trials.

Tribes in competitive scenarios had differentiation factors nearly four times higher than solo tribes, demonstrating that competition and added complexity contribute to role creation. Future research could look into more detailed metrics for measuring and characterizing behavior and signal differentiation; data representation is a significant hurdle, because role information is spread across time and distributed among all the animats in a tribe. NEC-DAC handles the complexity of the problem by averaging standard deviations within tribes over time into a single number (the differentiation factor), but more advanced techniques could, e.g., plot the behavior and signal sequence choices of individual animats against their input vectors to analyze how different animats would react to the same inputs. By controlling for inputs it might be possible to characterize behavior more thoroughly, and exhaustive boundary tests could be used to create visualizations of animat brains. A cluster analysis that categorized roles into a human-distinguishable representation would be particularly useful.

## *5.4 Architecture Comparisons*

(Maynard Smith and Harper, 1995) identify three classifications of signals that can be applied to the signals in NEC-DAC:

1. Self-reporting signals vs. other-reporting signals
    a. Self-reporting signals convey information about the signaler
    b. Other-reporting signals convey information about something other than the signaler
2. Symbols vs. icons vs. indices
    a. Symbolic signals are arbitrarily linked to objects
    b. Iconic signals resemble their objects
    c. Index signals are physically linked to their objects
3. Minimal signals vs. cost-added signals
    a. Minimal signals have costs that are no greater than is required to transmit their information
    b. Cost-added signals have costs than are directly needed

Most signals in NEC-DAC are other-reporting, except for signals that are used for giving feedback about the effectiveness of commands. Unlike the vast majority of natural signals, NEC-DAC signals are all symbolic. NEC-DAC signals are also minimal cost, with the cost being that an enemy animat that overhears a signal not intended for it may be attracted to the signaler. In the current architecture, a listening enemy animat cannot interpret or act upon a signal that it hears from another tribe (except in the Signal Interception experiment described in section 4.4.3), but it will set its Enemy Target variable to point to the signaler and can then move towards the signaler by focusing on the source of the signal rather than depending on sight, which has a shorter range. Thus, when an animat uses a signal other than Null, it may attract attention from enemies that would otherwise not have known of its presence. However, because these signals are minimal, this cost does not relate to the handicap principle proposed by (Zahavi, 1975, 1977) which hypothesizes that signals are useful precisely because they are "wasteful". Zahavi argued that expensive signals can be trusted when they impose a cost that reduces the fitness of the signaler, e.g., peacocks with elaborate tails are attractive to mates because they are able to survive despite having tails that make them easier for predators to detect. However, NEC-DAC does not concern itself with honesty and there is no sexual selection pressure (because there is no sex). In fact, the cost of signaling does not impact the ability of the signaler to win a combat if an enemy listener should decide to home in on the signal and attack, so it is not "wasteful" in Zahavi's sense. (It is possible that an animat could refrain from signaling when its Combat Power is low, but in practice this was not observed.)

NEC-DAC animats' signals are honest and altruistic because the animats are explicitly rewarded when their listeners benefit from obeying, and this is a plausible model that relates to signals in nature. For example, (Vehrencamp, 2000) examined bird songs and concluded that costly handicap signals appear to arise only when the signaler and receiver have conflicting interests, and (Lachman, Számadó and Bergstrom, 2001) argue that low-cost signaling -- and even language -- can arise when individuals have interests that are only somewhat coincident. Lachman et al. compare peacocks and sparrows and note that although both use plumage to signal fitness, the sparrows' signals are not costly. Peacocks use their tails to signal sexual fitness to potential mates, but sparrows use their chest "badges" to signal their aggressiveness. The researchers conclude that the sparrow signals don't need to be costly because it is easy for an opposing sparrow to determine the truth of the signal simply by initiating threatening behavior. A peahen, however, has no immediate way to determine before the fact whether or not the peacock she is considering will actually produce healthy offspring, and she has no way to punish a liar once the truth becomes known, so the cost of the signal must be paid up front. There is a motive to deceive when there are indivisible resources that both want to exploit or when there is sexual selection based on fitness, and in those cases the use of costly signals can discourage dishonesty. However, again, NEC-DAC animats only (purposefully) signal to others in their same tribe with whom they cooperate in stacking bricks, and there is no mating. Future work using NEC-DAC could impose a reward-related cost on signaling, but since the primary beneficiaries of the signals are the listeners such a cost would potentially eliminate useful signaling altogether.

      Other neural network models have made use of dual networks, such as such as the system described in (French, Ans & Rousset, 2001) to avoid catastrophic interference when learning sequences of patterns. Catastrophic interference occurs when an artificial neural network forgets recently learned information (last-in first-out) when it learns something new. Although a common problem with artificial neural networks, this human

brain does not exhibit this flaw but instead forgets things learned farthest in the past before forgetting more recently learned things (first-in first-out). To avoid catastrophic interference, French et al. modified the standard neural network architecture by combining two networks, one of which sees all newly-learned information first (the "hippocampal" short-term memory network) and only later transfers that information to the second network (the "neocortical" long-term memory network). The general information flow of the system during this transfer is shown in Figure 19 (initial pattern learning) and Figure 19 (knowledge transfer).

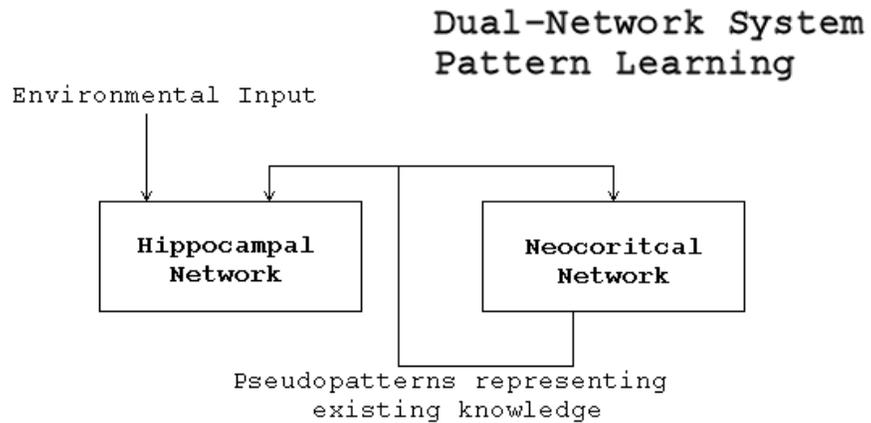

**Figure 18 -- This figure illustrates the data flow during initial pattern learning from the environment in the dual-network system described by French et al.**

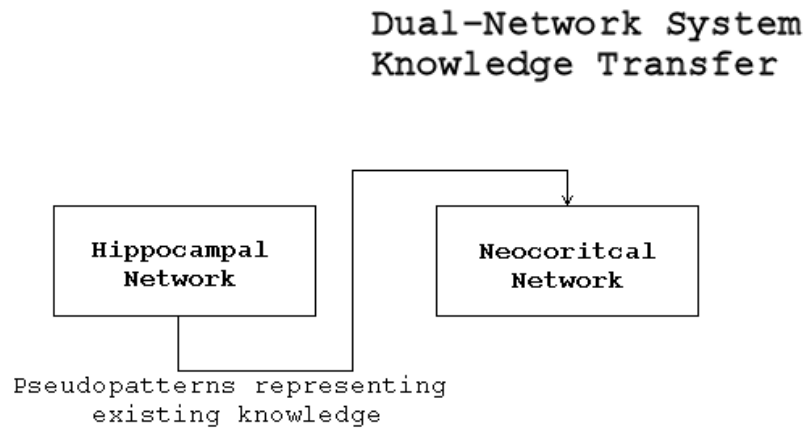

**Figure 19 -- This figure illustrates the data flow during knowledge transference between the two networks in the dual-network system described by French et al. There is no output from the Neocortical Network in this case because it is simply accepting knowledge from the Hippocampal Network.**

When new information from the environment is learned it is fed only into the hippocampal network and is interleaved with *pseudopatterns* (first proposed by (Robins, 1995)) that represent the knowledge stored in the second, neocortical network. A pseudopattern consists of a random input vector and the output vector generated as a result of feeding that input vector through a network. The new, real information that the hippocampal network is attempting to learn is interspersed with pseudopatterns generated by the neocortical network; when the hippocampal network learns from these pseudopatterns its existing knowledge is reinforced while new knowledge is introduced in between these patterns. By generating a large number of pseudopatterns it is possible to characterize the function encoded in the neocortical neural network, and by interleaving these pseudopatterns with new information during training it is possible to avoid interfering with existing knowledge in the network. Once the hippocampal network is trained to the desired criterion, its total knowledge is then transferred to the neocortical network via pseudopatterns. Thus, in a dual-network system, the neural network concerned with long-term learning only sees a fully integrated set of knowledge and will not have to assimilate new information on the fly. However, the two networks in NEC-DAC are used for an entirely different purpose: to divide the problem space into two distinct subsets. My neural networks both receive the same input data and use it to make decisions in two different domains.

## 5.5  General Suggestions for Future Improvements

There are a few general improvements that could be made that might increase the power of the animats and the usefulness of the results by increasing the complexity of the system. The added complexity might tease out new, more intelligent behaviors from the animats and could make them more human-like.

Briefly:

1. Test communication paradigms beyond that of audible sound, such as various computer networking technologies (message passing) and bulletin board systems (shared memory).

2. Modify the memory registers to be more than static storage, such as counters, random number sources, or mathematical operators.

3. Create a more powerful signaling paradigm with more characteristics of true language, such as: grounding, creativity, arbitrariness, phrase structure, negation, recursive semantics and syntax.

4. Experiment with variable levels of honesty in signaling (Noble, Di Paolo, and Bullock, 2002).

5. Replace the symbolic behaviors and signals with neural equivalents, perhaps using a version of egocentric spatial maps (Panangadan and Dyer, 2001) for tracking brick locations. Consider using data compression techniques similar to biological eyes (Meister and Berry, 1999) to reduce the number of neurons required.

6. Incorporate evolutionary behaviors and signaling capabilities to touch the third complex adaptive system (other than learning and culture) identified by (Kirby, 2002).

7. Create a neural mechanism for animats to select signal recipients, including a broadcast mode.

8. Enable enemy animats to understand and act on the contents of overheard commands in addition to overheard information signals.

9. Create a simulation with endogenous tribes so that animats can decide where to place their loyalty.
10. Create a neural mechanism for animats to select combat targets.
11. Implement a building task that requires more complexity than stacking bricks in sequential order.
12. Enable animats to assign trust/obedience values to their compatriots individually.
13. Test with more than two tribes. If animats are also enabled to understand signals from other tribes, alliances between tribes could be explored.
14. Enable combat between animats in the same tribe so that commanders can punish disobedience.
15. Increase the knowledge animats have about each other.
16. Create a neural mechanism for animats to identify each other.
17. Implement signals that influence higher-order goals rather than only single actions.
18. Develop a higher-resolution metric for analyzing role differentiation.
19. Perform boundary tests or cluster analysis to more fully characterize the neural networks.
20. Isolate motivations for differentiation, separating discovered niches from commanded roles.
21. Implement indirect, more natural rewards for combat rather than internal rewards and investigate the circumstances under which such combat is beneficial enough to spur the creation of a warrior role.

22. Replace random start phase with a form of adaptive curiosity (Schmidhuber, 1991).

23. Increase the complexity of the world by adding more features: objects obstacles or dynamic environmental conditions such as a day/night cycle or weather.

# 6 Appendix

The Appendix contains data that supplements the main body of my dissertation. Several gigabytes of raw data were generated in total so it is impractical to distribute it all, but the following samples are representative.

## *6.1 Animat Queues*

Here are three complete dumps of the input and output queues from three different animats: one top-performer (Table 30), one middling-performer (Table 31), and one poor-performer (Table 32). These queues were taken at the time of the animats' deaths, when they were the oldest animats in their tribe, which means that at the time they were ignoring any commands sent to them by their compatriots. Not every input from section 3.2.4 is shown because not every input was used in the trials that generated this data; the inputs from the memory registers were used but their values were omitted because they vary very little over the course of 100 think cycles near the end of an animat's life. Table 33 shows the queues of an animat from the middle of its lifecycle. Table 29 contains the key to the abbreviations used in the subsequent tables.

**Table 29 -- These are the abbreviations used in the subsequent animat queue tables.**

| Abbreviation | Meaning | Abbreviation | Meaning |
|---|---|---|---|
| *Column Headings* | | | |
| # | Queue index | L.Action | Last action performed (in the previous think cycle) |
| L.Signal | Last signal uttered (in the previous think cycle) | L.Seen | Observed action of target friendly animat |
| Heard | Last signal heard (only present in Table 33) | | |
| ATFT | At friendly tower | SEFT | See friendly tower |
| ATET | At enemy tower | SEET | See enemy tower |
| HT | Have enemy target | HB | Have brick |
| HBR | Have right brick | HBTL | Have brick too low |
| HBTH | Have brick too high | FTOB | Friendly tower observation time |
| FTHT | Friendly tower height | ETOB | Enemy tower observation time |
| SEFR | See friendly animat | SAL | Same friendly animat signal target as last think cycle |
| *Actions* | | | |
| NONE | Wait action | DROP | Drop brick |
| GTFT | Go to friendly tower | EXPL | Explore |
| GTET | Go to enemy tower | ATTK | Attack |
| FD<x> | Fetch brick x | DFND | Defend |
| *Signals* | | | |
| None | Null signal | C_EXPL | Command to explore |
| C_GTFT | Command to GTFT | C_ATTK | Command to attack |
| C_GTET | Command to GTET | C_DFND | Command to defend |
| C_FD<x> | Command to FD<x> | D<x>LOC | Query brick x location |
| C_DROP | Command to drop brick | THEIGHT | Query friendly tower height |

**Table 30 -- This table shows the final 100 entries in the input and output queues of animat 051116-230748-0011-2-1233 which scored a record high 20017 points. The "#" column indicates the index in the circular queue for the data shown, and the first entry is the most recent; each later row takes one step back in time, and the last entry in the table contains the earliest data. The inputs from the memory registers were omitted because they vary very little near the end of an animat's life.**

| # | L.Action | L.Signal | L.Seen | ATFT | SEFT | ATET | SEET | HT | HB | HBR | HBTL | HBTH | FTOB | FTHT | ETOB | SEFR | SAL |
|---|---|---|---|---|---|---|---|---|---|---|---|---|---|---|---|---|---|
| 88 | GTFT | C_GTFT | DROP | 1 | 1 | 0 | 0 | 1 | 1 | 1 | 0 | 0 | 0 | 0.8 | 1 | 1 | 0 |
| 87 | GTFT | C_GTFT |  | 0 | 0 | 0 | 0 | 0 | 1 | 1 | 0 | 0 | 0.64 | 0.8 | 1 | 0 | 0 |
| 86 | FD5 | C_DROP |  | 0 | 0 | 0 | 0 | 1 | 1 | 1 | 0 | 0 | 0.423 | 0.8 | 1 | 0 | 0 |
| 85 | DROP | C_DROP | DROP | 1 | 1 | 0 | 0 | 0 | 0 | 0 | 0 | 0 | 0 | 0.8 | 1 | 1 | 1 |
| 84 | GTFT | NONE | GTFT | 1 | 1 | 0 | 0 | 0 | 0 | 0 | 0 | 0 | 0 | 0.6 | 1 | 1 | 0 |

| # | L.Action | L.Signal | L.Seen | ATFT | SEFT | ATET | SEET | HT | HB | HBR | HBTL | HBTH | FTOB | FTHT | ETOB | SEFR | SAL |
|---|---|---|---|---|---|---|---|---|---|---|---|---|---|---|---|---|---|
| 83 | GTFT | C_GTFT | GTFT | 1 | 1 | 0 | 0 | 1 | 0 | 0 | 0 | 0 | 0 | 0.6 | 1 | 1 | 0 |
| 82 | GTFT | C_DROP | GTFT | 1 | 1 | 0 | 0 | 0 | 0 | 0 | 0 | 0 | 0 | 0.6 | 1 | 1 | 0 |
| 81 | GTFT | C_DROP | GTFT | 1 | 1 | 0 | 0 | 0 | 0 | 0 | 0 | 0 | 0 | 0.6 | 1 | 1 | 0 |
| 80 | DROP | C_DROP | DFND | 1 | 1 | 0 | 0 | 0 | 0 | 0 | 0 | 0 | 0 | 0.6 | 1 | 1 | 1 |
| 79 | GTFT | C_GTFT | DFND | 1 | 1 | 0 | 0 | 1 | 0 | 0 | 0 | 0 | 0 | 0.4 | 1 | 1 | 0 |
| 78 | GTFT | C_GTFT | FD4 | 1 | 1 | 0 | 0 | 1 | 0 | 0 | 0 | 0 | 0 | 0.4 | 1 | 1 | 1 |
| 77 | GTFT | C_GTFT | GTFT | 1 | 1 | 0 | 0 | 1 | 0 | 0 | 0 | 0 | 0 | 0.4 | 1 | 1 | 0 |
| 76 | GTFT | C_DROP | DROP | 1 | 1 | 0 | 0 | 1 | 0 | 0 | 0 | 0 | 0 | 0.2 | 1 | 1 | 1 |
| 75 | DROP | C_GTFT | FD2 | 1 | 1 | 0 | 0 | 1 | 0 | 0 | 0 | 0 | 0 | 0.2 | 1 | 1 | 0 |
| 74 | DROP | C_DROP | DFND | 1 | 1 | 0 | 0 | 1 | 0 | 0 | 0 | 0 | 0 | 0.2 | 1 | 1 | 1 |
| 73 | DROP | C_DROP | DFND | 1 | 1 | 0 | 0 | 1 | 0 | 0 | 0 | 0 | 0 | 0.2 | 1 | 1 | 1 |
| 72 | DROP | C_GTFT | DFND | 1 | 1 | 0 | 0 | 1 | 0 | 0 | 0 | 0 | 0 | 0.2 | 1 | 1 | 0 |
| 71 | DROP | C_GTFT | GTET | 1 | 1 | 0 | 0 | 1 | 0 | 0 | 0 | 0 | 0 | 0.2 | 1 | 1 | 0 |
| 70 | GTFT | C_DROP | GTFT | 1 | 1 | 0 | 0 | 1 | 0 | 0 | 0 | 0 | 0 | 0.2 | 1 | 1 | 0 |
| 69 | DROP | C_GTFT | FD2 | 1 | 1 | 0 | 0 | 1 | 0 | 0 | 0 | 0 | 0 | 0.2 | 1 | 1 | 0 |
| 68 | DROP | C_GTFT | GTFT | 1 | 1 | 0 | 0 | 0 | 0 | 0 | 0 | 0 | 0 | 0.2 | 1 | 1 | 0 |
| 67 | DROP | C_DROP | DROP | 1 | 1 | 0 | 0 | 1 | 0 | 0 | 0 | 0 | 0 | 0.2 | 1 | 1 | 0 |
| 66 | GTFT | C_GTFT | DROP | 1 | 1 | 0 | 0 | 1 | 0 | 0 | 0 | 0 | 0 | 0.2 | 1 | 1 | 0 |
| 65 | DROP | C_GTFT | GTFT | 0 | 1 | 0 | 0 | 1 | 0 | 0 | 0 | 0 | 0 | 0.2 | 1 | 1 | 1 |
| 64 | DROP | D4LOC | GTFT | 0 | 1 | 0 | 0 | 1 | 0 | 0 | 0 | 0 | 0 | 0.2 | 1 | 1 | 1 |
| 63 | GTFT | C_DROP | GTFT | 0 | 1 | 0 | 0 | 1 | 0 | 0 | 0 | 0 | 0 | 0.2 | 1 | 1 | 1 |
| 62 | DROP | C_GTFT | GTET | 0 | 0 | 0 | 0 | 1 | 0 | 0 | 0 | 0 | 0.113 | 0.2 | 1 | 0 | 1 |
| 61 | EXPL | C_GTFT | GTET | 0 | 0 | 0 | 0 | 1 | 1 | 0 | 1 | 0 | 0.071 | 0.2 | 1 | 0 | 0 |
| 60 | FD1 | C_GTFT | GTFT | 1 | 1 | 0 | 0 | 1 | 1 | 0 | 1 | 0 | 0 | 0.2 | 1 | 1 | 1 |
| 59 | DROP | C_FD1 | DROP | 1 | 1 | 0 | 0 | 1 | 0 | 0 | 0 | 0 | 0 | 0.2 | 1 | 1 | 1 |
| 58 | DROP | C_DROP | DROP | 1 | 1 | 0 | 0 | 0 | 0 | 0 | 0 | 0 | 0 | 0.2 | 1 | 1 | 1 |
| 57 | GTFT | C_GTFT | GTFT | 1 | 1 | 0 | 0 | 1 | 0 | 0 | 0 | 0 | 0 | 0.2 | 1 | 1 | 1 |
| 56 | DROP | C_GTFT | GTFT | 1 | 1 | 0 | 0 | 1 | 0 | 0 | 0 | 0 | 0 | 0.2 | 1 | 1 | 1 |
| 55 | DROP | C_DROP | GTFT | 1 | 1 | 0 | 0 | 1 | 0 | 0 | 0 | 0 | 0 | 0.2 | 1 | 1 | 0 |
| 54 | GTFT | C_DROP | DFND | 1 | 1 | 0 | 0 | 1 | 0 | 0 | 0 | 0 | 0 | 0 | 1 | 1 | 1 |
| 53 | GTFT | C_GTFT | GTFT | 1 | 1 | 0 | 0 | 1 | 0 | 0 | 0 | 0 | 0 | 0 | 1 | 1 | 0 |
| 52 | DROP | C_GTFT | DROP | 1 | 1 | 0 | 0 | 1 | 0 | 0 | 0 | 0 | 0 | 0 | 1 | 1 | 1 |
| 51 | GTFT | C_GTET | GTFT | 1 | 1 | 0 | 0 | 1 | 0 | 0 | 0 | 0 | 0 | 0 | 1 | 1 | 0 |
| 50 | GTFT | C_DROP | GTFT | 1 | 1 | 0 | 0 | 1 | 0 | 0 | 0 | 0 | 0 | 0 | 1 | 1 | 0 |
| 49 | DROP | C_GTFT | GTFT | 0 | 0 | 0 | 0 | 1 | 0 | 0 | 0 | 0 | 0.932 | 0 | 0.854 | 0 | 0 |
| 48 | DFND | C_GTFT | FD5 | 0 | 0 | 0 | 0 | 1 | 0 | 0 | 0 | 0 | 0.899 | 0 | 0.821 | 0 | 0 |
| 47 | EXPL | C_DROP | GTET | 0 | 0 | 0 | 0 | 1 | 0 | 0 | 0 | 0 | 0.338 | 0 | 0.26 | 1 | 0 |
| 46 | EXPL | C_GTET | FD2 | 0 | 0 | 0 | 0 | 1 | 0 | 0 | 0 | 0 | 0.248 | 0 | 0.17 | 0 | 0 |
| 45 | DROP | C_DROP | DFND | 1 | 1 | 0 | 0 | 1 | 0 | 0 | 0 | 0 | 0 | 0 | 0.833 | 1 | 1 |
| 44 | DROP | C_GTFT | DFND | 1 | 1 | 0 | 0 | 1 | 0 | 0 | 0 | 0 | 0 | 0 | 0.798 | 1 | 1 |
| 43 | GTFT | C_DROP | DFND | 1 | 1 | 0 | 0 | 1 | 0 | 0 | 0 | 0 | 0 | 0 | 0.747 | 1 | 0 |
| 42 | DROP | C_GTFT | DFND | 1 | 1 | 0 | 0 | 1 | 0 | 0 | 0 | 0 | 0 | 0 | 0.708 | 1 | 1 |
| 41 | GTFT | D2LOC | DFND | 1 | 1 | 0 | 0 | 1 | 0 | 0 | 0 | 0 | 0 | 0 | 0.674 | 1 | 0 |
| 40 | GTFT | C_GTFT | DFND | 1 | 1 | 0 | 0 | 1 | 0 | 0 | 0 | 0 | 0 | 0 | 0.636 | 1 | 0 |
| 39 | DROP | C_DROP | DFND | 1 | 1 | 0 | 0 | 1 | 0 | 0 | 0 | 0 | 0 | 0 | 0.592 | 1 | 1 |
| 38 | DROP | C_DROP | DFND | 1 | 1 | 0 | 0 | 1 | 0 | 0 | 0 | 0 | 0 | 0 | 0.549 | 1 | 1 |
| 37 | DROP | C_DROP | DFND | 1 | 1 | 0 | 0 | 1 | 0 | 0 | 0 | 0 | 0 | 0 | 0.499 | 1 | 1 |
| 36 | DROP | C_GTFT | DFND | 1 | 1 | 0 | 0 | 1 | 0 | 0 | 0 | 0 | 0 | 0 | 0.463 | 1 | 1 |
| 35 | GTFT | C_DROP | DFND | 1 | 1 | 0 | 0 | 1 | 0 | 0 | 0 | 0 | 0 | 0 | 0.431 | 1 | 0 |
| 34 | GTFT | D3LOC | DFND | 1 | 1 | 0 | 0 | 1 | 0 | 0 | 0 | 0 | 0 | 0 | 0.397 | 1 | 0 |
| 33 | GTFT | C_GTFT | GTFT | 1 | 1 | 0 | 0 | 1 | 0 | 0 | 0 | 0 | 0 | 0 | 0.355 | 1 | 1 |
| 32 | DROP | D3LOC | FD2 | 0 | 1 | 0 | 0 | 1 | 0 | 0 | 0 | 0 | 0 | 0 | 0.237 | 1 | 1 |

| # | L.Action | L.Signal | L.Seen | ATFT | SEFT | ATET | SEET | HT | HB | HBR | HBTL | HBTH | FTOB | FTHT | ETOB | SEFR | SAL |
|---|---|---|---|---|---|---|---|---|---|---|---|---|---|---|---|---|---|
| 31 | DROP | C_DROP | FD2 | 0 | 1 | 0 | 0 | 1 | 0 | 0 | 0 | 0 | 0 | 0 | 0.192 | 1 | 0 |
| 30 | GTFT | C_GTFT | FD2 | 0 | 1 | 0 | 0 | 1 | 0 | 0 | 0 | 0 | 0 | 0 | 0.141 | 1 | 0 |
| 29 | GTFT | C_FD1 | FD5 | 0 | 0 | 0 | 1 | 1 | 0 | 0 | 0 | 0 | 0.683 | 0 | 0 | 0 | 1 |
| 28 | DROP | C_GTET | FD5 | 0 | 0 | 0 | 1 | 1 | 0 | 0 | 0 | 0 | 0.621 | 0 | 0 | 1 | 1 |
| 27 | FD5 | C_DROP | FD5 | 0 | 0 | 0 | 1 | 1 | 0 | 0 | 0 | 0 | 0.587 | 0 | 0 | 1 | 0 |
| 26 | DROP | C_GTFT | GTFT | 0 | 0 | 0 | 0 | 1 | 0 | 0 | 0 | 0 | 0.409 | 0 | 0.073 | 1 | 1 |
| 25 | GTFT | C_GTFT | FD4 | 0 | 0 | 0 | 0 | 1 | 0 | 0 | 0 | 0 | 0.374 | 0 | 0.038 | 1 | 1 |
| 24 | GTET | D4LOC | DFND | 0 | 0 | 1 | 1 | 1 | 0 | 0 | 0 | 0 | 0.22 | 0 | 0 | 1 | 0 |
| 23 | GTFT | C_GTFT | GTFT | 1 | 1 | 0 | 0 | 1 | 0 | 0 | 0 | 0 | 0 | 0 | 1 | 1 | 1 |
| 22 | DROP | C_GTFT | GTFT | 1 | 1 | 0 | 0 | 1 | 0 | 0 | 0 | 0 | 0 | 0 | 1 | 1 | 1 |
| 21 | GTFT | C_DROP | GTFT | 1 | 1 | 0 | 0 | 0 | 0 | 0 | 0 | 0 | 0 | 0 | 1 | 1 | 0 |
| 20 | GTFT | C_DROP | DFND | 1 | 1 | 0 | 0 | 0 | 0 | 0 | 0 | 0 | 0 | 0 | 1 | 1 | 0 |
| 19 | DROP | C_DROP | DROP | 1 | 1 | 0 | 0 | 0 | 0 | 0 | 0 | 0 | 0 | 0.8 | 1 | 1 | 1 |
| 18 | GTFT | D4LOC | GTFT | 1 | 1 | 0 | 0 | 1 | 0 | 0 | 0 | 0 | 0 | 0.8 | 1 | 1 | 0 |
| 17 | DROP | C_GTFT | DFND | 1 | 1 | 0 | 0 | 1 | 0 | 0 | 0 | 0 | 0 | 0.8 | 1 | 1 | 1 |
| 16 | GTFT | C_GTFT | DFND | 1 | 1 | 0 | 0 | 1 | 0 | 0 | 0 | 0 | 0 | 0.8 | 1 | 1 | 0 |
| 15 | GTFT | D2LOC | GTFT | 1 | 1 | 0 | 0 | 1 | 0 | 0 | 0 | 0 | 0 | 0.8 | 1 | 1 | 0 |
| 14 | GTFT | C_GTFT | GTFT | 1 | 1 | 0 | 0 | 1 | 0 | 0 | 0 | 0 | 0 | 0.8 | 1 | 1 | 0 |
| 13 | GTFT | C_GTFT | GTET | 1 | 1 | 0 | 0 | 1 | 0 | 0 | 0 | 0 | 0 | 0.8 | 1 | 1 | 1 |
| 12 | DROP | C_DROP | FD3 | 1 | 1 | 0 | 0 | 1 | 0 | 0 | 0 | 0 | 0 | 0.8 | 1 | 1 | 1 |
| 11 | DROP | C_DROP | GTFT | 1 | 1 | 0 | 0 | 1 | 0 | 0 | 0 | 0 | 0 | 0.8 | 1 | 1 | 0 |
| 10 | GTFT | C_GTFT | DFND | 1 | 1 | 0 | 0 | 1 | 0 | 0 | 0 | 0 | 0 | 0.8 | 1 | 1 | 1 |
| 9 | DROP | C_DROP | GTFT | 1 | 1 | 0 | 0 | 1 | 0 | 0 | 0 | 0 | 0 | 0.8 | 1 | 1 | 0 |
| 8 | DROP | C_GTFT | GTFT | 1 | 1 | 0 | 0 | 1 | 0 | 0 | 0 | 0 | 0 | 0.8 | 1 | 1 | 0 |
| 7 | DROP | C_GTFT | DFND | 1 | 1 | 0 | 0 | 1 | 0 | 0 | 0 | 0 | 0 | 0.8 | 1 | 1 | 0 |
| 6 | DROP | C_GTFT | GTFT | 1 | 1 | 0 | 0 | 1 | 0 | 0 | 0 | 0 | 0 | 0.8 | 1 | 1 | 0 |
| 5 | GTFT | C_GTFT | DFND | 1 | 1 | 0 | 0 | 1 | 1 | 0 | 1 | 0 | 0 | 0.8 | 1 | 1 | 1 |
| 4 | DFND | C_GTFT | DFND | 1 | 1 | 0 | 0 | 1 | 1 | 0 | 1 | 0 | 0 | 0.8 | 1 | 1 | 0 |
| 3 | DROP | C_GTFT | WAIT | 1 | 1 | 0 | 0 | 0 | 1 | 0 | 1 | 0 | 0 | 0.8 | 0.427 | 1 | 1 |
| 2 | GTFT | C_DROP | WAIT | 1 | 1 | 0 | 0 | 0 | 1 | 0 | 1 | 0 | 0 | 0.8 | 0.406 | 1 | 0 |
| 1 | FD1 | C_DROP | GTFT | 1 | 1 | 0 | 0 | 0 | 1 | 0 | 1 | 0 | 0 | 0.8 | 0.365 | 1 | 0 |
| 0 | GTFT | C_FD2 | FD1 | 1 | 1 | 0 | 0 | 1 | 1 | 0 | 1 | 0 | 0 | 0.8 | 0.322 | 1 | 0 |
| 99 | GTFT | C_GTFT | GTFT | 1 | 1 | 0 | 0 | 1 | 1 | 0 | 1 | 0 | 0 | 0.8 | 0.287 | 1 | 1 |
| 98 | GTFT | C_DROP | DROP | 1 | 1 | 0 | 0 | 1 | 1 | 0 | 1 | 0 | 0 | 0.8 | 0.245 | 1 | 1 |
| 97 | GTFT | C_DROP | FD4 | 1 | 1 | 0 | 0 | 1 | 1 | 0 | 1 | 0 | 0 | 0.8 | 0.2 | 1 | 0 |
| 96 | FD1 | C_DROP | GTET | 0 | 0 | 0 | 1 | 1 | 1 | 0 | 1 | 0 | 1 | 0.8 | 0 | 0 | 0 |
| 95 | EXPL | C_DROP | EXPL | 0 | 0 | 0 | 0 | 1 | 0 | 0 | 0 | 0 | 1 | 0.8 | 0.044 | 0 | 0 |
| 94 | DROP | C_DROP | FD3 | 0 | 0 | 0 | 1 | 1 | 0 | 0 | 0 | 0 | 0.863 | 0.8 | 0 | 1 | 0 |
| 93 | GTFT | C_GTET | FD5 | 0 | 0 | 0 | 1 | 1 | 0 | 0 | 0 | 0 | 0.823 | 0.8 | 0 | 1 | 0 |
| 92 | DROP | C_GTFT | GTFT | 0 | 0 | 0 | 0 | 1 | 0 | 0 | 0 | 0 | 0.558 | 0.8 | 0.195 | 1 | 1 |
| 91 | EXPL | C_DROP | FD3 | 0 | 0 | 0 | 0 | 1 | 0 | 0 | 0 | 0 | 0.507 | 0.8 | 0.144 | 1 | 0 |
| 90 | DROP | C_GTFT | GTFT | 0 | 0 | 0 | 1 | 1 | 0 | 0 | 0 | 0 | 0.329 | 0.8 | 0 | 1 | 1 |
| 89 | DROP | C_DROP | GTFT | 0 | 0 | 0 | 1 | 1 | 0 | 0 | 0 | 0 | 0.284 | 0.8 | 0 | 1 | 1 |

In Table 30 we see that even one of the best performing animats, Animat 1233, doesn't appear to behave in an "optimal" manner, which might be considered to be fetching, returning and dropping bricks with the sizes required to stack on top of its friendly tower. In fact, the signals uttered by Animat 1233 appear to command behavior that is more optimal than its own actions. For instance, in rows 29, 30 and 31 we can see it issue commands to fetch brick-1, go to friendly tower, and then drop the brick (even though the SAL column tells us that these commands were issued to at least two different animats). However, we do see in rows 86, 87 and 88 that the animat fetches the right brick and then returns to its tower (only to be eliminated before it score a final time). Similar behavior can also be seen in rows 96 through 3, though the brick fetched (brick-1) is too low to fit onto the tower.

Table 31 -- This table shows the final 100 entries in the input and output queues of animat 051116-230748-0008-1-888 which scored the median score of 5705 points. The "#" column indicates the index in the circular queue for the data shown, and the first entry is the most recent; each later row takes one step back in time, and the last entry in the table contains the earliest data. The inputs from the memory registers were omitted because they vary very little near the end of an animat's life.

| # | L.Action | L.Signal | L.Seen | ATFT | SEFT | ATET | SEET | HT | HB | HBR | HBTL | HBTH | FTOB | FTHT | ETOB | SEFR | SAL |
|---|---|---|---|---|---|---|---|---|---|---|---|---|---|---|---|---|---|
| 67 | FD3 | C_DROP | FD1 | 0 | 1 | 0 | 0 | 1 | 1 | 0 | 1 | 0 | 0 | 0.6 | 1 | 1 | 0 |
| 66 | FD1 | C_DROP | DROP | 0 | 1 | 0 | 0 | 1 | 1 | 0 | 1 | 0 | 0 | 0.6 | 1 | 1 | 0 |
| 65 | DROP | C_DROP | DROP | 0 | 1 | 0 | 0 | 1 | 0 | 0 | 0 | 0 | 0 | 0.6 | 1 | 0 | 1 |
| 64 | DROP | C_DROP | GTFT | 0 | 1 | 0 | 0 | 0 | 0 | 0 | 0 | 0 | 0 | 0.6 | 1 | 0 | 0 |
| 63 | FD3 | C_DROP | FD3 | 0 | 1 | 0 | 0 | 1 | 1 | 0 | 1 | 0 | 0 | 0.6 | 1 | 0 | 0 |
| 62 | DROP | C_DROP | FD4 | 0 | 1 | 0 | 0 | 1 | 0 | 0 | 0 | 0 | 0 | 0.6 | 1 | 1 | 1 |
| 61 | DROP | C_DROP | FD4 | 0 | 1 | 0 | 0 | 1 | 0 | 0 | 0 | 0 | 0 | 0.6 | 1 | 1 | 1 |
| 60 | DROP | C_DROP | FD4 | 0 | 1 | 0 | 0 | 1 | 0 | 0 | 0 | 0 | 0 | 0.6 | 1 | 1 | 1 |
| 59 | FD4 | C_DROP | FD4 | 0 | 1 | 0 | 0 | 1 | 1 | 1 | 0 | 0 | 0 | 0.6 | 1 | 1 | 0 |
| 58 | FD3 | C_DROP | DROP | 0 | 1 | 0 | 0 | 1 | 1 | 0 | 1 | 0 | 0 | 0.6 | 1 | 1 | 1 |
| 57 | DROP | C_DROP | FD3 | 0 | 1 | 0 | 0 | 1 | 0 | 0 | 0 | 0 | 0 | 0.6 | 1 | 1 | 1 |
| 56 | FD3 | C_DROP | FD3 | 0 | 1 | 0 | 0 | 1 | 1 | 0 | 1 | 0 | 0 | 0.6 | 1 | 1 | 0 |
| 55 | FD3 | C_DROP | FD2 | 0 | 1 | 0 | 0 | 1 | 1 | 0 | 1 | 0 | 0 | 0.6 | 1 | 1 | 0 |
| 54 | WAIT | C_DROP |  | 0 | 0 | 0 | 0 | 0 | 0 | 0 | 0 | 0 | 1 | 0.6 | 0.738 | 0 | 0 |
| 53 | FD3 | C_DROP |  | 0 | 0 | 0 | 0 | 1 | 0 | 0 | 0 | 0 | 1 | 0.6 | 0.709 | 0 | 0 |
| 52 | DROP | C_FD3 | FD1 | 0 | 0 | 0 | 0 | 1 | 0 | 0 | 0 | 0 | 0.827 | 0.6 | 0.312 | 0 | 1 |
| 51 | FD3 | C_FD3 | FD5 | 0 | 0 | 0 | 0 | 1 | 0 | 0 | 0 | 0 | 0.789 | 0.6 | 0.274 | 0 | 0 |
| 50 | FD2 | C_DROP | GTFT | 0 | 0 | 0 | 0 | 1 | 0 | 0 | 0 | 0 | 0.624 | 0.6 | 0.109 | 1 | 0 |
| 49 | DROP | C_DROP | FD3 | 0 | 0 | 1 | 1 | 1 | 0 | 0 | 0 | 0 | 0.386 | 0.6 | 0 | 1 | 1 |
| 48 | GTET | C_DROP | FD1 | 0 | 0 | 1 | 1 | 1 | 0 | 0 | 0 | 0 | 0.347 | 0.6 | 0 | 1 | 0 |
| 47 | FD3 | C_DROP | FD2 | 0 | 0 | 0 | 0 | 1 | 0 | 0 | 0 | 0 | 0.13 | 0.6 | 1 | 1 | 1 |
| 46 | FD2 | C_DROP | FD2 | 0 | 0 | 0 | 0 | 1 | 1 | 0 | 1 | 0 | 0.057 | 0.6 | 1 | 1 | 0 |
| 45 | FD3 | C_DROP | GTFT | 1 | 1 | 0 | 0 | 1 | 1 | 0 | 1 | 0 | 0 | 0.6 | 1 | 1 | 1 |
| 44 | DROP | C_DROP | DROP | 1 | 1 | 0 | 0 | 1 | 0 | 0 | 0 | 0 | 0 | 0.6 | 1 | 1 | 1 |
| 43 | FD3 | C_DROP | DROP | 1 | 1 | 0 | 0 | 1 | 1 | 0 | 1 | 0 | 0 | 0.6 | 1 | 1 | 1 |

| #  | L.Action | L.Signal | L.Seen | ATFT | SEFT | ATET | SEET | HT | HB | HBR | HBTL | HBTH | FTOB  | FTHT | ETOB  | SEFR | SAL |
|----|----------|----------|--------|------|------|------|------|----|----|-----|------|------|-------|------|-------|------|-----|
| 42 | DROP     | C_DROP   | DROP   | 0    | 1    | 0    | 0    | 1  | 0  | 0   | 0    | 0    | 0     | 0.6  | 1     | 1    | 1   |
| 41 | FD3      | C_DROP   | GTFT   | 0    | 1    | 0    | 0    | 1  | 0  | 0   | 0    | 0    | 0     | 0.6  | 1     | 1    | 0   |
| 40 | FD3      | C_DROP   | DROP   | 0    | 1    | 0    | 0    | 1  | 0  | 0   | 0    | 0    | 0     | 0.6  | 1     | 1    | 1   |
| 39 | DROP     | C_DROP   | DFND   | 0    | 1    | 0    | 0    | 1  | 0  | 0   | 0    | 0    | 0     | 0.6  | 1     | 1    | 0   |
| 38 | FD1      | C_DROP   | FD3    | 0    | 1    | 0    | 0    | 1  | 1  | 0   | 1    | 0    | 0     | 0.6  | 1     | 1    | 0   |
| 37 | DROP     | C_DROP   | DROP   | 0    | 1    | 0    | 0    | 1  | 0  | 0   | 0    | 0    | 0     | 0.6  | 1     | 1    | 1   |
| 36 | DROP     | C_DROP   | DROP   | 0    | 1    | 0    | 0    | 1  | 0  | 0   | 0    | 0    | 0     | 0.6  | 1     | 1    | 1   |
| 35 | DROP     | NONE     | FD3    | 0    | 1    | 0    | 0    | 1  | 0  | 0   | 0    | 0    | 0     | 0.6  | 1     | 1    | 0   |
| 34 | FD3      | C_DROP   | DROP   | 0    | 1    | 0    | 0    | 1  | 1  | 0   | 1    | 0    | 0     | 0.6  | 1     | 1    | 1   |
| 33 | FD3      | C_DROP   | DROP   | 0    | 1    | 0    | 0    | 1  | 1  | 0   | 1    | 0    | 0     | 0.6  | 1     | 1    | 1   |
| 32 | DROP     | C_DROP   | DROP   | 0    | 1    | 0    | 0    | 1  | 0  | 0   | 0    | 0    | 0     | 0.6  | 1     | 1    | 1   |
| 31 | FD3      | C_DROP   | DROP   | 0    | 1    | 0    | 0    | 1  | 1  | 0   | 1    | 0    | 0     | 0.6  | 1     | 1    | 1   |
| 30 | DROP     | C_DROP   | DROP   | 0    | 1    | 0    | 0    | 1  | 0  | 0   | 0    | 0    | 0     | 0.6  | 1     | 1    | 1   |
| 29 | DROP     | C_DROP   | FD1    | 0    | 1    | 0    | 0    | 1  | 0  | 0   | 0    | 0    | 0     | 0.6  | 1     | 1    | 1   |
| 28 | FD3      | C_DROP   | FD1    | 0    | 1    | 0    | 0    | 1  | 1  | 0   | 1    | 0    | 0     | 0.6  | 1     | 1    | 0   |
| 27 | DROP     | C_DROP   | GTFT   | 0    | 1    | 0    | 0    | 1  | 0  | 0   | 0    | 0    | 0     | 0.6  | 1     | 1    | 0   |
| 26 | DROP     | C_DROP   | FD4    | 0    | 1    | 0    | 0    | 1  | 0  | 0   | 0    | 0    | 0     | 0.6  | 1     | 1    | 1   |
| 25 | FD3      | C_DROP   | GTFT   | 1    | 1    | 0    | 0    | 1  | 1  | 0   | 1    | 0    | 0     | 0.6  | 1     | 1    | 0   |
| 24 | DROP     | C_DROP   | FD5    | 0    | 1    | 0    | 0    | 1  | 0  | 0   | 0    | 0    | 0     | 0.6  | 1     | 1    | 1   |
| 23 | FD1      | C_DROP   | GTFT   | 0    | 1    | 0    | 0    | 1  | 1  | 0   | 1    | 0    | 0     | 0.6  | 1     | 1    | 0   |
| 22 | FD3      | C_DROP   |        | 0    | 0    | 0    | 0    | 1  | 1  | 0   | 1    | 0    | 0.758 | 0.6  | 1     | 0    | 0   |
| 21 | DROP     | C_DROP   | DFND   | 0    | 0    | 0    | 0    | 0  | 0  | 0   | 0    | 0    | 0.453 | 0.6  | 1     | 0    | 1   |
| 20 | DROP     | C_DROP   | DFND   | 0    | 0    | 0    | 0    | 0  | 0  | 0   | 0    | 0    | 0.412 | 0.6  | 1     | 0    | 1   |
| 19 | FD3      | C_DROP   | DFND   | 0    | 0    | 0    | 0    | 1  | 0  | 0   | 0    | 0    | 0.37  | 0.6  | 1     | 0    | 0   |
| 18 | FD3      | C_DROP   | DROP   | 0    | 0    | 0    | 0    | 1  | 0  | 0   | 0    | 0    | 0.198 | 0.6  | 0.973 | 1    | 1   |
| 17 | FD2      | C_DROP   | FD2    | 0    | 0    | 0    | 0    | 1  | 1  | 0   | 1    | 0    | 0.073 | 0.6  | 0.848 | 1    | 0   |
| 16 | FD3      | C_DROP   | FD2    | 0    | 1    | 0    | 0    | 1  | 1  | 0   | 1    | 0    | 0     | 0.6  | 0.58  | 1    | 0   |
| 15 | DROP     | C_DROP   | FD3    | 0    | 1    | 0    | 0    | 0  | 0  | 0   | 0    | 0    | 0     | 0.6  | 0.496 | 1    | 0   |
| 14 | FD3      | C_DROP   | DFND   | 1    | 1    | 0    | 0    | 0  | 1  | 0   | 1    | 0    | 0     | 0.6  | 0.424 | 1    | 0   |
| 13 | FD3      | C_DROP   | FD3    | 1    | 1    | 0    | 0    | 0  | 1  | 0   | 1    | 0    | 0     | 0.6  | 0.39  | 1    | 0   |
| 12 | DROP     | C_DROP   | GTFT   | 1    | 1    | 0    | 0    | 1  | 0  | 0   | 0    | 0    | 0     | 0.6  | 0.285 | 1    | 0   |
| 11 | GTFT     | C_DROP   | FD5    | 1    | 1    | 0    | 0    | 1  | 0  | 0   | 0    | 0    | 0     | 0.6  | 0.231 | 1    | 0   |
| 10 | DROP     | C_DROP   | FD5    | 0    | 0    | 0    | 1    | 1  | 0  | 0   | 0    | 0    | 1     | 0.6  | 0     | 1    | 1   |
| 9  | FD1      | D1LOC    | FD5    | 0    | 0    | 0    | 1    | 1  | 1  | 0   | 1    | 0    | 1     | 0.6  | 0     | 1    | 0   |
| 8  | DROP     | C_DROP   | DFND   | 0    | 0    | 1    | 1    | 1  | 0  | 0   | 0    | 0    | 1     | 0.6  | 0     | 1    | 1   |
| 7  | DROP     | C_DROP   | DFND   | 0    | 0    | 1    | 1    | 1  | 0  | 0   | 0    | 0    | 1     | 0.6  | 0     | 1    | 0   |
| 6  | WAIT     | C_DROP   | GTET   | 0    | 0    | 1    | 1    | 1  | 1  | 0   | 1    | 0    | 1     | 0.6  | 0     | 1    | 1   |
| 5  | FD3      | C_DROP   | GTET   | 0    | 0    | 1    | 1    | 1  | 1  | 0   | 1    | 0    | 1     | 0.6  | 0     | 1    | 1   |
| 4  | FD3      | C_DROP   | GTET   | 0    | 0    | 1    | 1    | 1  | 1  | 0   | 1    | 0    | 1     | 0.6  | 0     | 1    | 1   |
| 3  | GTET     | C_DROP   | FD5    | 0    | 0    | 1    | 1    | 1  | 1  | 0   | 1    | 0    | 1     | 0.6  | 0     | 1    | 0   |
| 2  | FD3      | C_DROP   | DFND   | 0    | 0    | 0    | 1    | 1  | 1  | 0   | 1    | 0    | 1     | 0.6  | 0     | 1    | 1   |
| 1  | DROP     | C_DROP   | DFND   | 0    | 0    | 0    | 1    | 1  | 0  | 0   | 0    | 0    | 1     | 0.6  | 0     | 1    | 1   |
| 0  | DROP     | C_DROP   | DFND   | 0    | 0    | 0    | 1    | 1  | 0  | 0   | 0    | 0    | 1     | 0.6  | 0     | 1    | 1   |
| 99 | DFND     | C_DROP   | WAIT   | 0    | 0    | 1    | 1    | 1  | 1  | 0   | 1    | 0    | 1     | 0.6  | 0     | 1    | 0   |
| 98 | GTET     | C_DROP   | EXPL   | 0    | 0    | 1    | 1    | 1  | 1  | 0   | 1    | 0    | 0.83  | 0.6  | 0     | 1    | 0   |
| 97 | FD3      | C_DROP   | DROP   | 0    | 0    | 0    | 0    | 1  | 1  | 0   | 1    | 0    | 0.507 | 0.6  | 0.922 | 0    | 1   |
| 96 | FD3      | C_DROP   | DROP   | 0    | 0    | 0    | 0    | 1  | 1  | 0   | 1    | 0    | 0.455 | 0.6  | 0.87  | 0    | 1   |
| 95 | FD3      | C_DROP   | FD1    | 0    | 0    | 0    | 0    | 0  | 1  | 0   | 1    | 0    | 0.417 | 0.6  | 0.832 | 0    | 1   |
| 94 | DROP     | C_DROP   | FD1    | 0    | 0    | 0    | 0    | 1  | 0  | 0   | 0    | 0    | 0.382 | 0.6  | 0.797 | 0    | 0   |
| 93 | FD3      | C_DROP   | FD1    | 0    | 0    | 0    | 0    | 0  | 1  | 0   | 1    | 0    | 0.349 | 0.6  | 0.764 | 0    | 0   |
| 92 | ATTK     | C_DROP   | FD5    | 0    | 0    | 0    | 0    | 1  | 0  | 0   | 0    | 0    | 0.307 | 0.6  | 0.722 | 0    | 0   |
| 91 | FD4      | C_DROP   | DROP   | 0    | 0    | 0    | 0    | 1  | 0  | 0   | 0    | 0    | 0.224 | 0.6  | 0.639 | 0    | 0   |

| #  | L.Action | L.Signal | L.Seen | ATFT | SEFT | ATET | SEET | HT | HB | HBR | HBTL | HBTH | FTOB  | FTHT | ETOB  | SEFR | SAL |
|----|----------|----------|--------|------|------|------|------|----|----|-----|------|------|-------|------|-------|------|-----|
| 90 | FD3      | C_DROP   | DROP   | 0    | 0    | 0    | 0    | 1  | 1  | 0   | 1    | 0    | 0.12  | 0.6  | 0.535 | 0    | 0   |
| 89 | DROP     | D1LOC    | GTFT   | 1    | 1    | 0    | 0    | 1  | 0  | 0   | 0    | 0    | 0     | 0.6  | 0.289 | 1    | 0   |
| 88 | FD1      | C_DROP   | GTFT   | 1    | 1    | 0    | 0    | 1  | 1  | 0   | 1    | 0    | 0     | 0.6  | 0.252 | 1    | 0   |
| 87 | WAIT     | C_DROP   | GTFT   | 1    | 1    | 0    | 0    | 1  | 1  | 0   | 1    | 0    | 0     | 0.6  | 0.212 | 1    | 1   |
| 86 | FD1      | C_DROP   | GTFT   | 1    | 1    | 0    | 0    | 1  | 1  | 0   | 1    | 0    | 0     | 0.6  | 0.186 | 1    | 0   |
| 85 | FD3      | C_DROP   | FD3    | 0    | 0    | 0    | 1    | 1  | 1  | 1   | 0    | 0    | 0.817 | 0.4  | 0     | 1    | 0   |
| 84 | EXPL     | C_DROP   | EXPL   | 0    | 0    | 0    | 0    | 1  | 0  | 0   | 0    | 0    | 0.617 | 0.4  | 1     | 1    | 0   |
| 83 | DROP     | C_DROP   | DROP   | 0    | 0    | 0    | 0    | 1  | 0  | 0   | 0    | 0    | 0.313 | 0.4  | 0.625 | 1    | 1   |
| 82 | DROP     | C_DROP   | DROP   | 0    | 0    | 0    | 0    | 0  | 0  | 0   | 0    | 0    | 0.268 | 0.4  | 0.58  | 1    | 1   |
| 81 | DROP     | C_DROP   | DROP   | 0    | 0    | 0    | 0    | 1  | 0  | 0   | 0    | 0    | 0.235 | 0.4  | 0.547 | 1    | 1   |
| 80 | DFND     | C_DROP   | DROP   | 0    | 0    | 0    | 0    | 1  | 1  | 1   | 0    | 0    | 0.2   | 0.4  | 0.512 | 1    | 1   |
| 79 | WAIT     | C_DROP   | DROP   | 0    | 0    | 0    | 0    | 1  | 1  | 1   | 0    | 0    | 0.13  | 0.4  | 0.442 | 1    | 1   |
| 78 | FD3      | C_DROP   | DROP   | 0    | 0    | 0    | 0    | 1  | 1  | 1   | 0    | 0    | 0.111 | 0.4  | 0.423 | 1    | 0   |
| 77 | FD2      | C_DROP   | GTFT   | 1    | 1    | 0    | 0    | 1  | 0  | 0   | 0    | 0    | 0     | 0.4  | 0.185 | 1    | 0   |
| 76 | DROP     | C_DROP   | GTFT   | 0    | 0    | 0    | 1    | 1  | 0  | 0   | 0    | 0    | 0.396 | 0.4  | 0     | 1    | 1   |
| 75 | GTET     | C_DROP   | GTFT   | 0    | 0    | 1    | 1    | 1  | 1  | 1   | 0    | 0    | 0.331 | 0.4  | 0     | 1    | 1   |
| 74 | FD3      | C_DROP   | DROP   | 0    | 0    | 0    | 1    | 1  | 1  | 1   | 0    | 0    | 0.285 | 0.4  | 0     | 1    | 1   |
| 73 | GTET     | C_DROP   | FD1    | 0    | 0    | 1    | 1    | 1  | 0  | 0   | 0    | 0    | 0.217 | 0.4  | 0     | 1    | 0   |
| 72 | FD2      | NONE     | FD2    | 0    | 1    | 0    | 0    | 1  | 0  | 0   | 0    | 0    | 0     | 0.4  | 0.243 | 1    | 0   |
| 71 | GTET     | C_DROP   | GTET   | 0    | 0    | 1    | 1    | 1  | 0  | 0   | 0    | 0    | 0.19  | 0.4  | 0     | 0    | 0   |
| 70 | DROP     | C_DROP   | DROP   | 0    | 1    | 0    | 0    | 0  | 0  | 0   | 0    | 0    | 0     | 0.4  | 1     | 1    | 1   |
| 69 | DROP     | C_DROP   | DROP   | 0    | 1    | 0    | 0    | 0  | 0  | 0   | 0    | 0    | 0     | 0.4  | 1     | 1    | 1   |
| 68 | DROP     | C_DROP   | FD1    | 0    | 1    | 0    | 0    | 0  | 0  | 0   | 0    | 0    | 0     | 0.4  | 1     | 1    | 0   |

Table 31, above, shows an excerpt from the queues of Animat 888, a middling performer. In rows 74 through 76 we can see the animat fetch the right brick but then take it to the enemy tower rather than the friendly tower before dropping it. Also notice that compared to Animat 1233 in Table 30, Animat 888 does a lot more fetching and a lot less returning to its friendly tower and dropping, the behaviors that are closest in time to the scoring event. Animat 888 also uses the drop command signal almost exclusively.

Table 32 -- This table shows the final 100 entries in the input and output queues of animat 051116-230748-0005-2-518 which scored a record low 89 points. The "#" column indicates the index in the circular queue for the data shown, and the first entry is the most recent; each later row takes one step back in time, and the last entry in the table contains the earliest data. The inputs from the memory registers were omitted because they vary very little near the end of an animat's life.

| #  | L.Action | L.Signal | L.Seen | ATFT | SEFT | ATET | SEET | HT | HB | HBR | HBTL | HBTH | FTOB  | FTHT | ETOB | SEFR | SAL |
|----|----------|----------|--------|------|------|------|------|----|----|-----|------|------|-------|------|------|------|-----|
| 14 | GTFT     | D5LOC    | GTFT   | 1    | 1    | 0    | 0    | 1  | 1  | 0   | 1    | 0    | 0     | 0.8  | 1    | 1    | 0   |
| 13 | FD1      | D5LOC    | EXPL   | 0    | 0    | 0    | 0    | 1  | 1  | 0   | 1    | 0    | 0.043 | 0.8  | 1    | 1    | 0   |
| 12 | DFND     | NONE     | FD3    | 0    | 1    | 0    | 0    | 1  | 0  | 0   | 0    | 0    | 0     | 0.8  | 1    | 1    | 0   |
| 11 | GTFT     | D1LOC    | WAIT   | 0    | 1    | 0    | 0    | 1  | 0  | 0   | 0    | 0    | 0     | 0.8  | 1    | 1    | 0   |

| # | L.Action | L.Signal | L.Seen | ATFT | SEFT | ATET | SEET | HT | HB | HBR | HBTL | HBTH | FTOB | FTHT | ETOB | SEFR | SAL |
|---|---|---|---|---|---|---|---|---|---|---|---|---|---|---|---|---|---|
| 10 | EXPL | NONE | DROP | 0 | 1 | 0 | 0 | 1 | 0 | 0 | 0 | 0 | 0 | 0.8 | 1 | 1 | 0 |
| 9 | FD5 | D4LOC | FD2 | 0 | 0 | 0 | 0 | 1 | 0 | 0 | 0 | 0 | 0.187 | 0.8 | 1 | 0 | 0 |
| 8 | WAIT | D4LOC | FD4 | 1 | 1 | 0 | 0 | 1 | 1 | 0 | 1 | 0 | 0 | 0.8 | 1 | 1 | 1 |
| 7 | FD1 | NONE | FD4 | 1 | 1 | 0 | 0 | 1 | 1 | 0 | 1 | 0 | 0 | 0.8 | 1 | 1 | 1 |
| 6 | GTFT | C_ATTK | FD4 | 1 | 1 | 0 | 0 | 1 | 1 | 0 | 1 | 0 | 0 | 0.8 | 1 | 1 | 0 |
| 5 | GTET | D1LOC |  | 0 | 0 | 0 | 0 | 0 | 1 | 0 | 1 | 0 | 1 | 0.8 | 1 | 0 | 0 |
| 4 | FD1 | NONE | EXPL | 0 | 0 | 0 | 0 | 0 | 1 | 0 | 1 | 0 | 1 | 0.8 | 1 | 1 | 1 |
| 3 | FD1 | NONE | EXPL | 0 | 0 | 0 | 0 | 0 | 1 | 0 | 1 | 0 | 1 | 0.8 | 1 | 0 | 0 |
| 2 | FD2 | D4LOC |  | 0 | 0 | 0 | 0 | 0 | 1 | 0 | 1 | 0 | 1 | 0.8 | 1 | 0 | 0 |
| 1 | FD4 | C_ATTK |  | 0 | 0 | 0 | 0 | 1 | 1 | 0 | 1 | 0 | 1 | 0.8 | 1 | 0 | 0 |
| 0 | EXPL | D1LOC |  | 0 | 0 | 0 | 0 | 1 | 1 | 1 | 0 | 0 | 1 | 0.8 | 0.939 | 0 | 0 |
| 99 | FD5 | D1LOC |  | 0 | 0 | 0 | 0 | 0 | 1 | 1 | 0 | 0 | 1 | 0.8 | 0.918 | 0 | 0 |
| 98 | FD4 | D5LOC |  | 0 | 0 | 0 | 0 | 1 | 1 | 0 | 1 | 0 | 1 | 0.8 | 0.847 | 0 | 0 |
| 97 | FD5 | D5LOC | EXPL | 0 | 0 | 0 | 0 | 0 | 1 | 1 | 0 | 0 | 1 | 0.8 | 0.75 | 0 | 1 |
| 96 | FD4 | D5LOC | EXPL | 0 | 0 | 0 | 0 | 0 | 1 | 0 | 1 | 0 | 1 | 0.8 | 0.587 | 0 | 0 |
| 95 | DROP | C_ATTK |  | 0 | 0 | 0 | 0 | 0 | 0 | 0 | 0 | 0 | 1 | 0.8 | 0.544 | 0 | 0 |
| 94 | GTET | D5LOC |  | 0 | 0 | 0 | 0 | 0 | 0 | 0 | 0 | 0 | 1 | 0.8 | 0.492 | 0 | 0 |
| 93 | ATTK | D4LOC |  | 0 | 0 | 0 | 0 | 0 | 0 | 0 | 0 | 0 | 1 | 0.8 | 0.443 | 0 | 0 |
| 92 | DROP | D1LOC | EXPL | 0 | 0 | 0 | 0 | 0 | 0 | 0 | 0 | 0 | 1 | 0.8 | 0.422 | 0 | 1 |
| 91 | EXPL | D4LOC | EXPL | 0 | 0 | 0 | 0 | 0 | 1 | 0 | 1 | 0 | 1 | 0.8 | 0.38 | 1 | 1 |
| 90 | FD4 | C_ATTK | EXPL | 0 | 0 | 0 | 0 | 1 | 1 | 0 | 1 | 0 | 1 | 0.8 | 0.345 | 1 | 0 |
| 89 | EXPL | D5LOC |  | 0 | 0 | 0 | 0 | 1 | 1 | 0 | 1 | 0 | 1 | 0.8 | 0.244 | 0 | 0 |
| 88 | FD2 | D4LOC | FD1 | 0 | 0 | 0 | 1 | 1 | 1 | 0 | 1 | 0 | 1 | 0.8 | 0 | 1 | 0 |
| 87 | DFND | D5LOC | DFND | 0 | 0 | 1 | 1 | 1 | 0 | 0 | 0 | 0 | 1 | 0.8 | 0 | 1 | 0 |
| 86 | GTET | D4LOC | FD5 | 0 | 0 | 1 | 1 | 1 | 0 | 0 | 0 | 0 | 1 | 0.8 | 0 | 1 | 0 |
| 85 | EXPL | D5LOC | FD1 | 0 | 0 | 0 | 0 | 1 | 0 | 0 | 0 | 0 | 1 | 0.8 | 0.383 | 1 | 0 |
| 84 | GTET | D1LOC | FD3 | 0 | 0 | 1 | 1 | 1 | 0 | 0 | 0 | 0 | 1 | 0.8 | 0 | 1 | 0 |
| 83 | ATTK | D5LOC | DFND | 0 | 0 | 0 | 0 | 1 | 0 | 0 | 0 | 0 | 1 | 0.8 | 1 | 1 | 1 |
| 82 | FD3 | NONE | DFND | 0 | 0 | 0 | 0 | 1 | 0 | 0 | 0 | 0 | 1 | 0.8 | 1 | 1 | 1 |
| 81 | FD1 | C_ATTK | DFND | 0 | 0 | 0 | 0 | 0 | 0 | 0 | 0 | 0 | 1 | 0.8 | 1 | 1 | 0 |
| 80 | FD3 | D5LOC |  | 0 | 0 | 0 | 0 | 0 | 0 | 0 | 0 | 0 | 1 | 0.8 | 1 | 0 | 0 |
| 79 | FD1 | NONE |  | 0 | 0 | 0 | 0 | 0 | 0 | 0 | 0 | 0 | 1 | 0.8 | 1 | 0 | 0 |
| 78 | ATTK | D4LOC |  | 0 | 0 | 0 | 0 | 0 | 1 | 0 | 1 | 0 | 1 | 0.8 | 1 | 0 | 0 |
| 77 | DFND | D1LOC |  | 0 | 0 | 0 | 0 | 0 | 1 | 0 | 1 | 0 | 1 | 0.8 | 1 | 0 | 0 |
| 76 | ATTK | D1LOC |  | 0 | 0 | 0 | 0 | 0 | 1 | 0 | 1 | 0 | 1 | 0.8 | 1 | 0 | 0 |
| 75 | WAIT | D5LOC |  | 0 | 0 | 0 | 0 | 0 | 1 | 0 | 1 | 0 | 1 | 0.8 | 0.991 | 0 | 0 |
| 74 | FD4 | D5LOC |  | 0 | 0 | 0 | 0 | 0 | 1 | 0 | 1 | 0 | 1 | 0.8 | 0.964 | 0 | 0 |
| 73 | GTET | NONE |  | 0 | 0 | 0 | 0 | 0 | 0 | 0 | 0 | 0 | 0.825 | 0.8 | 0.533 | 0 | 0 |
| 72 | GTET | C_ATTK |  | 0 | 0 | 0 | 0 | 0 | 0 | 0 | 0 | 0 | 0.788 | 0.8 | 0.496 | 0 | 0 |
| 71 | FD5 | D4LOC |  | 0 | 0 | 0 | 0 | 0 | 0 | 0 | 0 | 0 | 0.753 | 0.8 | 0.461 | 0 | 0 |
| 70 | FD2 | D1LOC |  | 0 | 0 | 0 | 0 | 1 | 1 | 0 | 1 | 0 | 0.715 | 0.8 | 0.423 | 0 | 0 |
| 69 | EXPL | C_ATTK | FD5 | 0 | 0 | 0 | 1 | 1 | 0 | 0 | 0 | 0 | 0.159 | 0.8 | 0 | 1 | 0 |
| 68 | WAIT | C_ATTK | GTFT | 0 | 1 | 0 | 0 | 1 | 0 | 0 | 0 | 0 | 0 | 0.8 | 0.382 | 1 | 1 |
| 67 | DROP | NONE | GTFT | 0 | 1 | 0 | 0 | 1 | 0 | 0 | 0 | 0 | 0 | 0.8 | 0.36 | 1 | 0 |
| 66 | DROP | D1LOC | GTFT | 0 | 1 | 0 | 0 | 0 | 0 | 0 | 0 | 0 | 0 | 0.8 | 0.317 | 0 | 0 |
| 65 | DROP | D5LOC | FD4 | 0 | 1 | 0 | 0 | 0 | 0 | 0 | 0 | 0 | 0 | 0.8 | 0.274 | 0 | 0 |
| 64 | GTFT | D1LOC | EXPL | 1 | 1 | 0 | 0 | 1 | 1 | 0 | 1 | 0 | 0 | 0.8 | 0.207 | 1 | 1 |
| 63 | ATTK | NONE | EXPL | 0 | 0 | 0 | 0 | 1 | 1 | 0 | 1 | 0 | 1 | 0.8 | 0.076 | 1 | 0 |
| 62 | ATTK | NONE | DFND | 0 | 0 | 0 | 0 | 1 | 1 | 0 | 1 | 0 | 0.928 | 0.8 | 0.07 | 1 | 0 |
| 61 | FD4 | D4LOC | FD3 | 0 | 0 | 0 | 0 | 1 | 1 | 0 | 1 | 0 | 0.655 | 0.8 | 0.278 | 0 | 1 |
| 60 | GTET | D1LOC | GTFT | 0 | 0 | 1 | 1 | 1 | 0 | 0 | 0 | 0 | 0.277 | 0.8 | 0 | 1 | 1 |
| 59 | WAIT | C_ATTK | GTFT | 0 | 0 | 1 | 1 | 1 | 0 | 0 | 0 | 0 | 0.242 | 0.8 | 0 | 1 | 1 |

| # | L.Action | L.Signal | L.Seen | ATFT | SEFT | ATET | SEET | HT | HB | HBR | HBTL | HBTH | FTOB | FTHT | ETOB | SEFR | SAL |
|---|---|---|---|---|---|---|---|---|---|---|---|---|---|---|---|---|---|
| 58 | GTET | D4LOC | FD1 | 0 | 0 | 1 | 1 | 1 | 0 | 0 | 0 | 0 | 0.214 | 0.8 | 0 | 1 | 0 |
| 57 | FD1 | D5LOC | GTFT | 0 | 1 | 0 | 0 | 1 | 0 | 0 | 0 | 0 | 0 | 0.8 | 1 | 1 | 0 |
| 56 | FD1 | NONE | DFND | 0 | 1 | 0 | 0 | 1 | 0 | 0 | 0 | 0 | 0 | 0.8 | 1 | 1 | 0 |
| 55 | DFND | D5LOC | FD5 | 0 | 0 | 0 | 0 | 1 | 1 | 0 | 1 | 0 | 0.47 | 0.6 | 0.878 | 1 | 0 |
| 54 | FD2 | D5LOC | GTFT | 0 | 0 | 0 | 0 | 1 | 1 | 0 | 1 | 0 | 0.134 | 0.6 | 0.542 | 1 | 1 |
| 53 | ATTK | C_ATTK | GTFT | 0 | 0 | 0 | 0 | 1 | 1 | 0 | 1 | 0 | 0.087 | 0.6 | 0.495 | 1 | 0 |
| 52 | FD2 | D1LOC | FD3 | 0 | 1 | 0 | 0 | 1 | 1 | 0 | 1 | 0 | 0 | 0.6 | 0.292 | 1 | 0 |
| 51 | FD2 | D5LOC | FD5 | 0 | 1 | 0 | 0 | 1 | 0 | 0 | 0 | 0 | 0 | 0.6 | 0.192 | 1 | 0 |
| 50 | GTFT | D4LOC | DFND | 0 | 0 | 0 | 0 | 1 | 0 | 0 | 0 | 0 | 0.69 | 0.6 | 0.057 | 1 | 1 |
| 49 | DROP | C_ATTK | DFND | 0 | 0 | 0 | 1 | 1 | 0 | 0 | 0 | 0 | 0.633 | 0.6 | 0 | 1 | 0 |
| 48 | GTET | D1LOC | ATTK | 0 | 0 | 0 | 1 | 1 | 0 | 0 | 0 | 0 | 0.584 | 0.6 | 0 | 1 | 0 |
| 47 | DFND | NONE | DFND | 0 | 0 | 0 | 0 | 1 | 0 | 0 | 0 | 0 | 0.525 | 0.6 | 0.342 | 1 | 1 |
| 46 | FD2 | C_ATTK | ATTK | 0 | 0 | 0 | 0 | 1 | 0 | 0 | 0 | 0 | 0.256 | 0.6 | 0.073 | 1 | 0 |
| 45 | FD5 | D4LOC | FD3 | 0 | 0 | 0 | 1 | 1 | 1 | 0 | 0 | 1 | 0.115 | 0.6 | 0 | 1 | 0 |
| 44 | GTET | C_ATTK | DFND | 0 | 1 | 0 | 0 | 1 | 1 | 1 | 0 | 0 | 0 | 0.6 | 0.906 | 1 | 1 |
| 43 | DFND | D5LOC | DFND | 1 | 1 | 0 | 0 | 1 | 1 | 1 | 0 | 0 | 0 | 0.6 | 0.861 | 1 | 1 |
| 42 | FD4 | NONE | DFND | 1 | 1 | 0 | 0 | 1 | 1 | 1 | 0 | 0 | 0 | 0.6 | 0.84 | 1 | 1 |
| 41 | FD5 | D4LOC | GTFT | 1 | 1 | 0 | 0 | 1 | 1 | 0 | 1 | 0 | 0 | 0.6 | 0.69 | 1 | 0 |
| 40 | FD2 | C_ATTK | FD1 | 1 | 1 | 0 | 0 | 1 | 1 | 0 | 1 | 0 | 0 | 0.6 | 0.628 | 1 | 1 |
| 39 | WAIT | D4LOC | FD1 | 1 | 1 | 0 | 0 | 1 | 1 | 0 | 1 | 0 | 0 | 0.6 | 0.594 | 1 | 1 |
| 38 | GTFT | C_ATTK | FD1 | 1 | 1 | 0 | 0 | 1 | 1 | 0 | 1 | 0 | 0 | 0.6 | 0.569 | 1 | 1 |
| 37 | FD1 | C_ATTK | DROP | 0 | 1 | 0 | 0 | 1 | 1 | 0 | 1 | 0 | 0 | 0.6 | 0.52 | 1 | 1 |
| 36 | FD1 | D1LOC | DROP | 0 | 1 | 0 | 0 | 1 | 0 | 0 | 0 | 0 | 0 | 0.6 | 0.376 | 1 | 0 |
| 35 | FD3 | D1LOC | FD5 | 0 | 1 | 0 | 0 | 1 | 1 | 0 | 1 | 0 | 0 | 0.6 | 0.224 | 1 | 0 |
| 34 | FD3 | D4LOC | FD3 | 0 | 1 | 0 | 0 | 1 | 1 | 0 | 1 | 0 | 0 | 0.6 | 0.174 | 1 | 0 |
| 33 | DROP | C_ATTK | FD1 | 0 | 0 | 1 | 1 | 1 | 0 | 0 | 0 | 0 | 1 | 0.6 | 0 | 1 | 1 |
| 32 | FD4 | D5LOC | EXPL | 0 | 0 | 0 | 1 | 1 | 1 | 0 | 0 | 1 | 1 | 0.6 | 0 | 1 | 0 |
| 31 | GTET | D4LOC | DFND | 0 | 0 | 1 | 1 | 1 | 1 | 0 | 0 | 1 | 1 | 0.6 | 0 | 0 | 0 |
| 30 | FD5 | C_ATTK | ATTK | 0 | 0 | 0 | 1 | 1 | 1 | 0 | 0 | 1 | 1 | 0.6 | 0 | 1 | 0 |
| 29 | FD4 | D4LOC | ATTK | 0 | 0 | 0 | 1 | 1 | 1 | 1 | 0 | 0 | 1 | 0.6 | 0 | 1 | 0 |
| 28 | WAIT | C_ATTK | FD3 | 0 | 0 | 1 | 1 | 1 | 1 | 0 | 1 | 0 | 1 | 0.6 | 0 | 1 | 1 |
| 27 | FD1 | D5LOC | FD3 | 0 | 0 | 1 | 1 | 1 | 1 | 0 | 1 | 0 | 1 | 0.6 | 0 | 1 | 1 |
| 26 | DROP | D1LOC | FD3 | 0 | 0 | 0 | 1 | 1 | 0 | 0 | 0 | 0 | 1 | 0.6 | 0 | 1 | 1 |
| 25 | FD2 | C_ATTK | GTET | 0 | 0 | 0 | 1 | 1 | 1 | 0 | 1 | 0 | 1 | 0.6 | 0 | 1 | 1 |
| 24 | FD3 | C_ATTK | GTET | 0 | 0 | 0 | 1 | 1 | 0 | 0 | 0 | 0 | 1 | 0.6 | 0 | 1 | 1 |
| 23 | WAIT | D5LOC | FD1 | 0 | 0 | 1 | 1 | 1 | 0 | 0 | 0 | 0 | 1 | 0.6 | 0 | 1 | 1 |
| 22 | GTET | D4LOC | FD1 | 0 | 0 | 1 | 1 | 1 | 0 | 0 | 0 | 0 | 1 | 0.6 | 0 | 1 | 1 |
| 21 | GTET | D5LOC | FD1 | 0 | 0 | 1 | 1 | 1 | 0 | 0 | 0 | 0 | 1 | 0.6 | 0 | 1 | 1 |
| 20 | ATTK | D5LOC | GTET | 0 | 0 | 0 | 0 | 1 | 0 | 0 | 0 | 0 | 1 | 0.6 | 0.077 | 1 | 0 |
| 19 | GTET | C_ATTK | GTET | 0 | 0 | 0 | 0 | 1 | 0 | 0 | 0 | 0 | 0.909 | 0.6 | 0.099 | 1 | 1 |
| 18 | FD4 | NONE | GTET | 0 | 0 | 0 | 0 | 1 | 0 | 0 | 0 | 0 | 0.876 | 0.6 | 0.066 | 1 | 1 |
| 17 | DROP | D5LOC | GTET | 0 | 0 | 0 | 1 | 1 | 0 | 0 | 0 | 0 | 0.789 | 0.6 | 0 | 0 | 0 |
| 16 | EXPL | NONE | FD2 | 0 | 0 | 0 | 1 | 1 | 1 | 0 | 1 | 0 | 0.752 | 0.6 | 0 | 0 | 0 |
| 15 | ATTK | D4LOC | GTET | 0 | 1 | 0 | 0 | 1 | 1 | 0 | 1 | 0 | 0 | 0.6 | 0.072 | 1 | 0 |

Animat 518, whose history queues are shown in Table 32, behaves in an essentially random manner, which probably explains why it scored so incredibly poorly.

**Table 33** -- This table shows the final 100 entries in the input and output queues of animat 051116-230748-0011-2-1287 which scored 7679 by the midpoint of its lifespan when this snapshot was taken. The "#" column indicates the index in the circular queue for the data shown, and the first entry is the most recent; each later row takes one step back in time, and the last entry in the table contains the earliest data.

| # | L.Act | L.Signal | L.Seen | Heard | ATFT | SEFT | ATET | SEET | HT | HB | HBR | HBTL | HBTH | FTOB | FTHT | ETOB | SEFR | SAL |
|---|---|---|---|---|---|---|---|---|---|---|---|---|---|---|---|---|---|---|
| 40 | GTFT | D2LOC | DFND | | 1 | 1 | 0 | 0 | 1 | 0 | 0 | 0 | 0 | 0 | 0.4 | 1 | 1 | 0 |
| 39 | DROP | C_DROP | DFND | | 0 | 1 | 0 | 0 | 1 | 0 | 0 | 0 | 0 | 0 | 0.4 | 1 | 1 | 1 |
| 38 | FD2 | C_FD5 | DFND | | 0 | 1 | 0 | 0 | 1 | 0 | 0 | 0 | 0 | 0 | 0.4 | 1 | 1 | 0 |
| 37 | DROP | C_DROP | GTFT | | 0 | 1 | 0 | 0 | 1 | 0 | 0 | 0 | 0 | 0 | 0.4 | 0.996 | 1 | 1 |
| 36 | DROP | D5LOC | FD1 | DROP | 0 | 1 | 0 | 0 | 1 | 0 | 0 | 0 | 0 | 0 | 0.4 | 0.951 | 1 | 1 |
| 35 | FD4 | C_DROP | DROP | | 0 | 1 | 0 | 0 | 1 | 1 | 0 | 0 | 1 | 0 | 0.4 | 0.908 | 1 | 1 |
| 34 | GTFT | C_DROP | GTFT | | 1 | 1 | 0 | 0 | 0 | 0 | 0 | 0 | 0 | 0 | 0.4 | 0.837 | 1 | 1 |
| 33 | GTFT | C_GTFT | FD1 | GTFT | 0 | 1 | 0 | 0 | 0 | 0 | 0 | 0 | 0 | 0 | 0.4 | 0.786 | 1 | 1 |
| 32 | DROP | C_DROP | FD1 | GTFT | 0 | 1 | 0 | 0 | 0 | 0 | 0 | 0 | 0 | 0 | 0.4 | 0.765 | 1 | 0 |
| 31 | GTFT | C_GTFT | WAIT | DROP | 1 | 1 | 0 | 0 | 1 | 1 | 0 | 1 | 0 | 0 | 0.4 | 0.648 | 1 | 0 |
| 30 | DROP | C_DROP | EXPL | | 0 | 0 | 0 | 0 | 1 | 1 | 0 | 1 | 0 | 0.852 | 0.4 | 0.35 | 0 | 1 |
| 29 | FD1 | C_GTFT | EXPL | | 0 | 0 | 0 | 0 | 1 | 1 | 0 | 1 | 0 | 0.831 | 0.4 | 0.329 | 0 | 0 |
| 28 | EXPL | C_DROP | | | 0 | 0 | 0 | 0 | 1 | 0 | 0 | 0 | 0 | 0.617 | 0.4 | 0.115 | 0 | 0 |
| 27 | DROP | C_DROP | DROP | | 0 | 1 | 0 | 0 | 0 | 0 | 0 | 0 | 0 | 0 | 0.4 | 0.796 | 1 | 1 |
| 26 | FD2 | C_DROP | DROP | | 0 | 1 | 0 | 0 | 1 | 1 | 0 | 1 | 0 | 0 | 0.4 | 0.763 | 1 | 0 |
| 25 | DROP | C_DROP | GTFT | GTFT | 1 | 1 | 0 | 0 | 1 | 0 | 0 | 0 | 0 | 0 | 0.4 | 0.687 | 1 | 0 |
| 24 | GTFT | C_GTFT | GTFT | DROP | 1 | 1 | 0 | 0 | 0 | 0 | 0 | 0 | 0 | 0 | 0.4 | 0.643 | 1 | 1 |
| 23 | ATTK | C_DROP | DROP | | 1 | 1 | 0 | 0 | 0 | 0 | 0 | 0 | 0 | 0 | 0.4 | 0.595 | 1 | 1 |
| 22 | DROP | C_GTFT | DROP | GTFT | 1 | 1 | 0 | 0 | 0 | 0 | 0 | 0 | 0 | 0 | 0.4 | 0.573 | 1 | 0 |
| 21 | GTFT | C_DROP | GTFT | | 1 | 1 | 0 | 0 | 0 | 0 | 0 | 0 | 0 | 0 | 0.4 | 0.525 | 1 | 0 |
| 20 | GTFT | C_DROP | EXPL | | 1 | 1 | 0 | 0 | 1 | 0 | 0 | 0 | 0 | 0 | 0.4 | 0.466 | 1 | 1 |
| 19 | FD4 | C_GTFT | GTFT | | 0 | 1 | 0 | 0 | 1 | 0 | 0 | 0 | 0 | 0 | 0.4 | 0.418 | 1 | 1 |
| 18 | DROP | C_FD5 | DROP | | 0 | 1 | 0 | 0 | 1 | 0 | 0 | 0 | 0 | 0 | 0.4 | 0.397 | 1 | 0 |
| 17 | DROP | C_DROP | GTET | DROP | 0 | 1 | 0 | 0 | 1 | 0 | 0 | 0 | 0 | 0 | 0.4 | 0.358 | 1 | 1 |
| 16 | GTFT | C_GTFT | DROP | DROP | 1 | 1 | 0 | 0 | 1 | 1 | 0 | 1 | 0 | 0 | 0.4 | 0.281 | 1 | 0 |
| 15 | GTFT | D4LOC | GTFT | DROP | 1 | 1 | 0 | 0 | 1 | 1 | 0 | 1 | 0 | 0 | 0.4 | 0.233 | 1 | 0 |
| 14 | GTFT | C_GTFT | GTET | | 0 | 0 | 0 | 0 | 1 | 1 | 0 | 1 | 0 | 0.245 | 0.4 | 0.089 | 1 | 0 |
| 13 | FD2 | C_GTFT | FD5 | | 0 | 0 | 0 | 1 | 1 | 1 | 0 | 1 | 0 | 0.134 | 0.4 | 0 | 0 | 0 |
| 12 | GTFT | C_GTFT | DFND | | 1 | 1 | 0 | 0 | 1 | 0 | 0 | 0 | 0 | 0 | 0.4 | 1 | 1 | 0 |
| 11 | FD3 | C_GTET | FD5 | | 0 | 1 | 0 | 0 | 0 | 0 | 0 | 0 | 0 | 0 | 0.4 | 1 | 1 | 0 |
| 10 | DROP | D5LOC | DFND | | 0 | 1 | 0 | 0 | 0 | 0 | 0 | 0 | 0 | 0 | 0.4 | 1 | 1 | 1 |
| 9 | WAIT | C_DROP | DFND | | 0 | 1 | 0 | 0 | 0 | 1 | 0 | 0 | 1 | 0 | 0.4 | 1 | 1 | 0 |
| 8 | FD5 | C_GTFT | FD5 | | 0 | 1 | 0 | 0 | 1 | 1 | 0 | 0 | 1 | 0 | 0.4 | 1 | 1 | 0 |
| 7 | GTFT | C_GTFT | GTFT | | 1 | 1 | 0 | 0 | 1 | 0 | 0 | 0 | 0 | 0 | 0.4 | 1 | 1 | 0 |
| 6 | WAIT | C_GTFT | FD4 | GTFT | 1 | 1 | 0 | 0 | 1 | 0 | 0 | 0 | 0 | 0 | 0.4 | 1 | 1 | 1 |
| 5 | DROP | C_GTFT | GTFT | | 1 | 1 | 0 | 0 | 1 | 0 | 0 | 0 | 0 | 0 | 0.4 | 1 | 1 | 0 |
| 4 | DROP | C_GTFT | DROP | DROP | 1 | 1 | 0 | 0 | 1 | 0 | 0 | 0 | 0 | 0 | 0.4 | 1 | 1 | 0 |
| 3 | GTFT | C_DROP | DROP | DROP | 1 | 1 | 0 | 0 | 1 | 0 | 0 | 0 | 0 | 0 | 0.4 | 1 | 1 | 0 |
| 2 | GTFT | C_GTFT | DFND | GTFT | 1 | 1 | 0 | 0 | 1 | 0 | 0 | 0 | 0 | 0 | 0.4 | 1 | 1 | 0 |
| 1 | DROP | C_GTFT | FD1 | | 0 | 0 | 0 | 0 | 1 | 0 | 0 | 0 | 0 | 0.414 | 0.4 | 1 | 1 | 1 |
| 0 | DROP | C_DROP | FD1 | | 0 | 0 | 0 | 0 | 1 | 0 | 0 | 0 | 0 | 0.368 | 0.4 | 1 | 1 | 1 |
| 99 | DFND | C_DROP | FD1 | DROP | 0 | 0 | 0 | 0 | 1 | 1 | 0 | 1 | 0 | 0.317 | 0.4 | 1 | 1 | 1 |
| 98 | DROP | C_FD2 | DFND | | 0 | 0 | 0 | 0 | 1 | 1 | 0 | 1 | 0 | 0.127 | 0.4 | 1 | 0 | 0 |
| 97 | FD1 | D4LOC | GTFT | | 0 | 0 | 0 | 0 | 1 | 1 | 0 | 1 | 0 | 0.105 | 0.4 | 1 | 0 | 0 |
| 96 | DROP | C_GTFT | GTFT | GTFT | 0 | 1 | 0 | 0 | 1 | 0 | 0 | 0 | 0 | 0 | 0.4 | 1 | 1 | 1 |
| 95 | GTFT | C_GTFT | GTFT | | 1 | 1 | 0 | 0 | 1 | 1 | 0 | 0 | 1 | 0 | 0.4 | 1 | 1 | 1 |
| 94 | GTFT | C_GTFT | GTFT | | 1 | 1 | 0 | 0 | 0 | 1 | 0 | 0 | 1 | 0 | 0.4 | 1 | 1 | 1 |

| # | L.Act | L.Signal | L.Seen | Heard | ATFT | SEFT | ATET | SEET | HT | HB | HBR | HBTL | HBTH | FTOB | FTHT | ETOB | SEFR | SAL |
|---|---|---|---|---|---|---|---|---|---|---|---|---|---|---|---|---|---|---|
| 93 | FD5 | C_DROP | WAIT | | 0 | 1 | 0 | 0 | 0 | 1 | 0 | 0 | 1 | 0 | 0.4 | 1 | 1 | 0 |
| 92 | DROP | C_DROP | GTFT | | 1 | 1 | 0 | 0 | 1 | 0 | 0 | 0 | 0 | 0 | 0.4 | 1 | 1 | 0 |
| 91 | GTFT | C_DROP | GTFT | | 1 | 1 | 0 | 0 | 1 | 0 | 0 | 0 | 0 | 0 | 0.4 | 1 | 1 | 0 |
| 90 | DROP | C_DROP | DROP | GTFT | 1 | 1 | 0 | 0 | 1 | 0 | 0 | 0 | 0 | 0 | 0.4 | 1 | 1 | 0 |
| 89 | GTFT | C_GTFT | DROP | FD1 | 1 | 1 | 0 | 0 | 1 | 1 | 1 | 0 | 0 | 0 | 0.2 | 1 | 1 | 0 |
| 88 | GTFT | C_DROP | DROP | GTFT | 1 | 1 | 0 | 0 | 1 | 1 | 1 | 0 | 0 | 0 | 0.2 | 1 | 1 | 0 |
| 87 | FD2 | C_DROP | DROP | GTFT | 0 | 1 | 0 | 0 | 1 | 1 | 1 | 0 | 0 | 0 | 0.2 | 1 | 1 | 0 |
| 86 | GTFT | C_DROP | DROP | DROP | 1 | 1 | 0 | 0 | 1 | 0 | 0 | 0 | 0 | 0 | 0.2 | 1 | 1 | 0 |
| 85 | GTFT | C_DROP | DROP | GTFT | 0 | 1 | 0 | 0 | 1 | 0 | 0 | 0 | 0 | 0 | 0 | 1 | 1 | 0 |
| 84 | DROP | C_GTFT | FD3 | GTFT | 0 | 1 | 0 | 0 | 1 | 0 | 0 | 0 | 0 | 0 | 0 | 1 | 1 | 0 |
| 83 | FD2 | NONE | DROP | DROP | 0 | 1 | 0 | 0 | 1 | 1 | 0 | 0 | 1 | 0 | 0 | 1 | 1 | 0 |
| 82 | DROP | C_GTFT | DFND | | 0 | 1 | 0 | 0 | 1 | 0 | 0 | 0 | 0 | 0 | 0 | 1 | 1 | 0 |
| 81 | DROP | C_DROP | FD5 | DROP | 0 | 1 | 0 | 0 | 1 | 0 | 0 | 0 | 0 | 0 | 0 | 1 | 1 | 0 |
| 80 | DROP | C_DROP | DROP | | 0 | 1 | 0 | 0 | 1 | 0 | 0 | 0 | 0 | 0 | 0 | 1 | 1 | 1 |
| 79 | DROP | C_GTFT | DROP | | 0 | 1 | 0 | 0 | 1 | 0 | 0 | 0 | 0 | 0 | 0 | 1 | 1 | 0 |
| 78 | DROP | C_DROP | ATTK | | 0 | 1 | 0 | 0 | 1 | 0 | 0 | 0 | 0 | 0 | 0 | 1 | 1 | 0 |
| 77 | FD2 | C_DROP | DFND | DROP | 0 | 1 | 0 | 0 | 1 | 1 | 0 | 0 | 1 | 0 | 0 | 1 | 1 | 0 |
| 76 | DROP | C_DROP | GTFT | | 1 | 1 | 0 | 0 | 0 | 0 | 0 | 0 | 0 | 0 | 0 | 1 | 1 | 1 |
| 75 | GTFT | D5LOC | GTFT | | 1 | 1 | 0 | 0 | 0 | 0 | 0 | 0 | 0 | 0 | 0 | 1 | 1 | 0 |
| 74 | DFND | C_GTFT | GTFT | GTFT | 1 | 1 | 0 | 0 | 1 | 0 | 0 | 0 | 0 | 0 | 0 | 1 | 1 | 0 |
| 73 | DROP | C_DROP | WAIT | | 1 | 1 | 0 | 0 | 1 | 0 | 0 | 0 | 0 | 0 | 0 | 1 | 1 | 0 |
| 72 | FD4 | C_GTFT | FD1 | | 1 | 1 | 0 | 0 | 1 | 0 | 0 | 0 | 0 | 0 | 0 | 1 | 1 | 0 |
| 71 | GTET | C_GTFT | GTFT | | 1 | 1 | 0 | 0 | 1 | 0 | 0 | 0 | 0 | 0 | 0 | 1 | 1 | 0 |
| 70 | DFND | C_DROP | FD2 | DROP | 1 | 1 | 0 | 0 | 0 | 0 | 0 | 0 | 0 | 0 | 0 | 1 | 1 | 0 |
| 69 | GTET | C_GTFT | ATTK | | 1 | 1 | 0 | 0 | 0 | 0 | 0 | 0 | 0 | 0 | 0 | 1 | 1 | 0 |
| 68 | DFND | C_DROP | GTFT | | 1 | 1 | 0 | 0 | 0 | 0 | 0 | 0 | 0 | 0 | 0 | 1 | 1 | 1 |
| 67 | GTFT | C_DROP | GTFT | | 1 | 1 | 0 | 0 | 0 | 0 | 0 | 0 | 0 | 0 | 0 | 1 | 1 | 1 |
| 66 | DROP | C_GTFT | GTFT | | 0 | 1 | 0 | 0 | 1 | 0 | 0 | 0 | 0 | 0 | 0 | 1 | 1 | 0 |
| 65 | FD2 | C_DROP | DROP | | 0 | 1 | 0 | 0 | 1 | 1 | 0 | 0 | 1 | 0 | 0 | 1 | 1 | 1 |
| 64 | FD5 | C_GTFT | GTFT | | 0 | 1 | 0 | 0 | 1 | 0 | 0 | 0 | 0 | 0 | 0 | 0.99 | 1 | 1 |
| 63 | DROP | C_DROP | GTFT | | 0 | 1 | 0 | 0 | 1 | 1 | 0 | 0 | 1 | 0 | 0 | 0.949 | 1 | 1 |
| 62 | FD2 | C_FD5 | GTFT | GTET | 0 | 1 | 0 | 0 | 1 | 1 | 0 | 0 | 1 | 0 | 0 | 0.927 | 1 | 1 |
| 61 | GTFT | D5LOC | GTFT | | 1 | 1 | 0 | 0 | 1 | 0 | 0 | 0 | 0 | 0 | 0 | 0.865 | 1 | 1 |
| 60 | GTFT | C_DROP | DROP | | 1 | 1 | 0 | 0 | 1 | 0 | 0 | 0 | 0 | 0 | 0 | 0.827 | 1 | 0 |
| 59 | DROP | C_GTFT | FD4 | GTFT | 1 | 1 | 0 | 0 | 1 | 0 | 0 | 0 | 0 | 0 | 0 | 0.777 | 1 | 0 |
| 58 | GTFT | C_GTFT | ATTK | DROP | 1 | 1 | 0 | 0 | 1 | 0 | 0 | 0 | 0 | 0 | 0 | 0.742 | 1 | 0 |
| 57 | GTFT | C_DROP | GTFT | GTFT | 1 | 1 | 0 | 0 | 1 | 0 | 0 | 0 | 0 | 0 | 0 | 0.69 | 1 | 0 |
| 56 | GTFT | C_DROP | DROP | | 1 | 1 | 0 | 0 | 1 | 0 | 0 | 0 | 0 | 0 | 0 | 0.639 | 1 | 1 |
| 55 | DROP | C_GTET | DROP | | 0 | 1 | 0 | 0 | 1 | 0 | 0 | 0 | 0 | 0 | 0 | 0.593 | 1 | 1 |
| 54 | DROP | C_DROP | DROP | | 0 | 1 | 0 | 0 | 1 | 0 | 0 | 0 | 0 | 0 | 0 | 0.544 | 1 | 0 |
| 53 | DFND | C_DROP | DROP | GTET | 1 | 1 | 0 | 0 | 1 | 1 | 0 | 0 | 1 | 0 | 0 | 0.475 | 1 | 0 |
| 52 | GTFT | C_GTFT | GTFT | | 1 | 1 | 0 | 0 | 1 | 1 | 0 | 0 | 1 | 0 | 0 | 0.355 | 1 | 0 |
| 51 | GTFT | C_DROP | GTFT | | 1 | 1 | 0 | 0 | 1 | 1 | 0 | 0 | 1 | 0 | 0 | 0.314 | 1 | 0 |
| 50 | FD2 | C_GTFT | GTFT | | 1 | 1 | 0 | 0 | 1 | 1 | 0 | 0 | 1 | 0 | 0 | 0.292 | 1 | 0 |
| 49 | DROP | C_GTFT | EXPL | | 0 | 0 | 0 | 0 | 1 | 0 | 0 | 0 | 0 | 0.33 | 0 | 0.129 | 1 | 1 |
| 48 | GTFT | C_GTFT | EXPL | | 0 | 0 | 0 | 0 | 1 | 1 | 0 | 0 | 1 | 0.281 | 0 | 0.08 | 1 | 0 |
| 47 | FD3 | C_DROP | DFND | DROP | 0 | 0 | 0 | 1 | 1 | 1 | 0 | 0 | 1 | 0.16 | 0 | 0 | 1 | 0 |
| 46 | DROP | C_DROP | DROP | GTFT | 1 | 1 | 0 | 0 | 1 | 0 | 0 | 0 | 0 | 0 | 0 | 1 | 1 | 0 |
| 45 | DROP | C_DROP | GTFT | GTFT | 1 | 1 | 0 | 0 | 1 | 0 | 0 | 0 | 0 | 0 | 0 | 1 | 1 | 1 |
| 44 | DROP | D3LOC | GTFT | DROP | 1 | 1 | 0 | 0 | 1 | 0 | 0 | 0 | 0 | 0 | 0 | 1 | 1 | 0 |
| 43 | FD2 | C_GTFT | FD4 | DROP | 1 | 1 | 0 | 0 | 1 | 1 | 0 | 0 | 1 | 0 | 0 | 1 | 1 | 0 |
| 42 | DROP | C_DROP | DROP | GTFT | 1 | 1 | 0 | 0 | 1 | 0 | 0 | 0 | 0 | 0 | 0 | 1 | 1 | 1 |

| # | L.Act | L.Signal | L.Seen | Heard | ATFT | SEFT | ATET | SEET | HT | HB | HBR | HBTL | HBTH | FTOB | FTHT | ETOB | SEFR | SAL |
|---|---|---|---|---|---|---|---|---|---|---|---|---|---|---|---|---|---|---|
| 41 | ATTK | C_DROP | GTFT | | 1 | 1 | 0 | 0 | 1 | 0 | 0 | 0 | 0 | 0 | 0 | 1 | 1 | 1 |

In Table 33 we can see a snapshot of the behavior of an above-average animat, Animat 1287, taken near the midpoint in its life. This table also includes a column indicating the signals that Animat 1287 heard, so we can connect its selection actions with the commands it hears. For example, in rows 32 through 34 we see that the animat obeys the three commands it heard in rows 31 through 33; however, in rows 42 through 48 the animat doesn't obey the commands it hears quite as reliably, perhaps because its own inputs strongly push it in a different direction.

## *6.2 Code Samples*

This section contains code samples from selected parts of the NEC-DAC system. All code is in C and is commented throughout.

### 6.2.1 Animat Think Loop

Here is the main function of the animats' think loop. This function is called every time an animat thinks and is responsible to handling all its behavior. In this sample, *ent* refers to the animat being processed at the time.

```c
void Animat_Think (edict_t *ent)
{
    // Declare and initialize local variables
    int cur_goal=0, i=0, j=0, cur_cmd=0;
    edict_t *target=NULL;
    edict_t *best = NULL;
    edict_t *enemy = NULL;
    edict_t *fr = NULL;
    edict_t *temp = NULL;
    double near_enemies = 0, near_friends = 0;
    int qi=0;
    double nt = 0;
    int count_score=0;
```

```c
    double t_i_state[I_NUM*2]={0};
    char out[200]={0};

    // Verify animat integrity if desired
    if (GS_VERIFYANIMATS)
    {
        VerifyAnimats();
    }

    // Verify that animats are named and anchored properly in the world.
    if (ent->classname == NULL)
    {
        gi.dprintf ("*** Animat_Think(): animat without name!\n");
    }

    if (ent->cache == NULL)
    {
        gi.dprintf ("*** Animat_Think(): animat without cache!\n");
    }

    // Check if animat has surpassed lifespan and if so, kill it.  If it is
    // carrying a brick then wait to kill it until it drops (to prevent
    // bricks from being lost.
    if (ent->birth_frame + C_LIFESPAN < level.framenum && ent->disc_carried == 0)
    {
        // Write input data to file for records
        Write_Inputs(-ent->id, out);
        gi.dprintf("Output to file: %s\n", out);

        // Write animat obituary to file for records
        Write_Obit(ent, out);
        gi.dprintf("Output to file: %s\n", out);

        VectorSet(ent->s.origin, -1000, -1000, C_MAPPLANE);
        VectorClear(ent->velocity);
        ent->nextthink = level.time + 9999999;

        // Tell the anchor to create a new animat to replace this one
        if (ent->health > 0)
        {
            ent->cache->res[1] += 1;
            ent->health = 0;
        }

        // Shuffle off this mortal coil
        G_FreeEdict(ent);
        return;
    }

    // Keep track of animat's location from when it last thought
    VectorCopy(ent->s.origin, ent->last_origin);

    // Find a target enemy if we don't have one
    if (ent->target_enemy == NULL || ent->target_enemy->inuse == 0)
    {
        ent->target_enemy = Nearby_Enemy(ent, &near_enemies);
    }

    // Sense any bricks within sight range
    Sense_Discs(ent);

    // Determine the queue index for this think cycle
    qi = ent->think_count%TQ_LENGTH;

    // If we're not busy, then let's think!
    if (ent->a_state >= 0)
    {
```

```c
		// Increment the think counter
		ent->think_count++;

		// Calculate the new queue index
		qi = ent->think_count%TQ_LENGTH;

		// Populate the animat's ANN inputs from sensory information
		Fill_ActionInputs(ent);

		// Copy the input vector to the input queue
		memcpy(ent->IQ[qi], ent->a_inputs, NN_INPUTS*sizeof(double));

		// If there's a command in the command queue, pop it
		A_QueuePop(ent, &ent->CQ[qi]);

		// Evaluate the ANN to choose a behavior
		Eval_State(ent, 1);

		// Set the behavioral state to the one selected by the ANN
		Set_State(ent);

		// Populate the animat's SNN inputs from sensory information
		Fill_SignalInputs(ent);

		// Copy the input vector to the input queue
		memcpy(ent->HQ[qi], ent->s_inputs, NN_INPUTS*sizeof(double));

		// Evaluate the SNN to choose a signal
		Eval_Signal(ent, 1);

		// Emit the signal chosen by the SNN
		Set_Signal(ent);

		// Copy the ANN and SNN outputs to the output history queues
		memcpy(ent->LQ[qi], ent->s_outputs, NN_OUTPUTS*sizeof(double));
		memcpy(ent->AQ[qi], ent->a_outputs, NN_OUTPUTS*sizeof(double));

		// Check if we should do a learn (C_LEARNCHANCE == 0.01)
		if (random()<C_LEARNCHANCE)
		{
			// Save the contents of the memory registers
			memcpy(t_i_state, ent->i_state, I_NUM*2*sizeof(double));

			// Train the networks.
			Train_State(ent);
			Train_Signal(ent);

			// Restore the contents of the memory registers
			memcpy(ent->i_state, t_i_state, I_NUM*2*sizeof(double));
		}

		// Check if we're past the random start phase
		if (level.framenum > C_RANDFRAMES)
		{
			// Update the behavior count for the records
			TotalCount_Update(ent);
		}

	}
	else // if we're busy, keep doing what we're doing
	{
	}

	// Increase our nextthink time
	if (ent->a_state > 0)
	{
		ent->nextthink = level.time + 2.1;
```

```c
	}
	else
	{
		ent->nextthink = level.time + (crandom() + 2.1);
	}

	// Execute our selected action
	switch (ent->a_state)
	{
	// If we're waiting, increase our combat power and do nothing
	case S_WAIT:
		VectorClear(ent->velocity);
		VectorCopy(ent->s.origin, ent->dest);
		ent->combat_power = MIN(ent->combat_power + 0.1, 1);
		ent->target_enemy = NULL; // 050203
		if (random()>C_GIVEUP)
		{
		}
		break;
	// If we're going to friendly tower...
	case S_GTFT:
		VectorClear(ent->velocity);
		VectorCopy(ent->s.origin, ent->dest);
		ent->target_enemy = NULL;
		// Move towards our tower
		if (ent->cache != NULL)
		{
			A_MoveTower(ent, 1);
		}
		else
		{
			gi.dprintf ("*** Animat_Think(): animat without cache 2!\n");
		}
		// Set to GTFT busy state
		ent->a_state = S_GTFTd;
		break;
	// GTFT busy state
	case S_GTFTd:
		// If we're at the tower, stop
		if (GRIDX(ent) == GRIDX(ent->cache) && GRIDY(ent) == GRIDY(ent->cache))
		{
			VectorClear(ent->velocity);
			VectorCopy(ent->s.origin, ent->dest);
			ent->a_state = S_GTFT;
		}
		// Otherwise, keep moving towards the tower
		else
		{
			A_MoveTower(ent, 1);
		}
		break;
	// Go to enemy tower
	case S_GTET:
		// Move towards the enemy tower
		VectorClear(ent->velocity);
		VectorCopy(ent->s.origin, ent->dest);
		A_MoveTower(ent, 0);
		// Set to GTET busy state
		ent->a_state = S_GTETd;
		break;
	// GTET busy state
	case S_GTETd:
		// If we no longer have a target enemy quit
		if (ent->target_enemy == NULL)
		{
			ent->a_state = S_GTET;
```

```
                }
                else
                {
                        // If we're at the enemy tower, stop
                        if (GRIDX(ent) == GRIDX(ent->target_enemy->cache) &&
                        GRIDY(ent) == GRIDY(ent->target_enemy->cache))
                        {
                                VectorClear(ent->velocity);
                                VectorCopy(ent->s.origin, ent->dest);
                                ent->a_state = S_GTET;
                        }
                        // Otherwise, move towards the enemy tower
                        else
                        {
                                A_MoveTower(ent, 0);
                        }
                }
                break;
// Fetch brick-1
case S_FD1:
        // Set to fetch brick-1 busy state
        ent->a_state = S_FD1d;
        VectorClear(ent->velocity);
        VectorCopy(ent->s.origin, ent->dest);
        ent->target_enemy = NULL;
        break;
// Fetch brick-1 busy state
case S_FD1d:
        // Fetch the brick, and if we get it then stop
        if (A_DiscFetch(ent, 1) == 1)
        {
                VectorClear(ent->velocity);
                VectorCopy(ent->s.origin, ent->dest);
                ent->a_state = S_FD1;
        }
        // Otherwise just keep on going
        else
        {
        }
        break;
// Fetch brick-2
case S_FD2:
        ent->a_state = S_FD2d;
        VectorClear(ent->velocity);
        VectorCopy(ent->s.origin, ent->dest);
        ent->target_enemy = NULL;
        break;
// Fetch brick-2 busy state
case S_FD2d:
        if (A_DiscFetch(ent, 2) == 1)
        {
                VectorClear(ent->velocity);
                VectorCopy(ent->s.origin, ent->dest);
                ent->a_state = S_FD2;
        }
        else
        {
        }
        break;
// Fetch brick-3
case S_FD3:
        ent->a_state = S_FD3d;
        VectorClear(ent->velocity);
        VectorCopy(ent->s.origin, ent->dest);
        ent->target_enemy = NULL;
        break;
// Fetch brick-3 busy state
```

```
                case S_FD3d:
                        if (A_DiscFetch(ent, 3) == 1)
                        {
                                VectorClear(ent->velocity);
                                VectorCopy(ent->s.origin, ent->dest);
                                ent->a_state = S_FD3;
                        }
                        else
                        {
                        }
                        break;
// Fetch brick-4
                case S_FD4:
                        ent->a_state = S_FD4d;
                        VectorClear(ent->velocity);
                        VectorCopy(ent->s.origin, ent->dest);
                        ent->target_enemy = NULL;
                        break;
// Fetch brick-4 busy state
                case S_FD4d:
                        if (A_DiscFetch(ent, 4) == 1)
                        {
                                VectorClear(ent->velocity);
                                VectorCopy(ent->s.origin, ent->dest);
                                ent->a_state = S_FD4;
                        }
                        else
                        {
                        }
                        break;
// Fetch brick-5
                case S_FD5:
                        ent->a_state = S_FD5d;
                        VectorClear(ent->velocity);
                        VectorCopy(ent->s.origin, ent->dest);
                        ent->target_enemy = NULL;
                        break;
// Fetch brick-5 busy state
                case S_FD5d:
                        if (A_DiscFetch(ent, 5) == 1)
                        {
                                VectorClear(ent->velocity);
                                VectorCopy(ent->s.origin, ent->dest);
                                ent->a_state = S_FD5;
                        }
                        else
                        {
                        }
                        break;
// Drop brick
                case S_DROP:
                        // Set to drop brick busy state
                        ent->a_state = S_DROPd;
                        VectorClear(ent->velocity);
                        VectorCopy(ent->s.origin, ent->dest);
                        ent->target_enemy = NULL;
                        break;
// Drop brick busy state
                case S_DROPd:
                        // Try to drop it, and if we succeed then we're done
                        if (A_DiscDrop(ent) == 1)
                        {
                                VectorClear(ent->velocity);
                                VectorCopy(ent->s.origin, ent->dest);
                                ent->a_state = S_DROP;
                        }
                        else
```

```
                {
                }
                break;
        // Explore
        case S_EXPLORE:
                // Move towards a random location
                A_MoveRandom(ent);
                // Set explore busy state
                ent->a_state = S_EXPLOREd;
                ent->target_enemy = NULL;
                break;
        // Explore busy state
        case S_EXPLOREd:
                // Possibly change our random movement
                if (random()<0.1)
                {
                        A_MoveRandom(ent);
                }
                else
                {
                        // Continue moving in current direction
                }
                break;
        // Attack
        case S_ATTACK:
                // If we have an enemy targeted...
                if (ent->target_enemy != NULL)
                {
                        // Move towards the enemy and set to the attack busy state
                        VectorCopy(ent->target_enemy->s.origin, ent->dest);
                        ent->a_state = S_ATTACKd;
                }
                else
                {
                        // Otherwise stay where we are
                        VectorClear(ent->velocity);
                        VectorCopy(ent->s.origin, ent->dest);
                }
                break;
        // Attack busy state
        case S_ATTACKd:
                if (ent->target_enemy != NULL)
                {
                        // If the enemy is within sight range...
                        if (Dist(ent->s.origin, ent->target_enemy->s.origin) <= C_SIGHT)
                        {
                                // Attack!
                                A_Attack(ent, ent->target_enemy);
                                ent->a_state = S_ATTACK;
                        }
                        else
                        {
                                // Otherwise keep moving towards the enemy
                                VectorCopy(ent->target_enemy->s.origin, ent->dest);
                        }
                }
                // If we don't have an enemy, then quit
                else
                {
                        ent->a_state = S_ATTACK;
                }
                break;
        // Defend
        case S_DEFEND:
                // Stay where we are end enter the defend busy state
                VectorClear(ent->velocity);
                VectorCopy(ent->s.origin, ent->dest);
```

```c
            ent->a_state = S_DEFENDd;
            ent->target_enemy = NULL;
            break;
// Defend busy state
case S_DEFENDd:
            // When we get attacked, then quit defending
            if (ent->just_attacked == 1)
            {
                    ent->a_state = S_DEFEND;
            }
            // Otherwise just keep going
            else
            {
            }

            break;
// Uh oh, something's broken!
default:
            gi.dprintf ("*** Animat_Think(): bad behavior, doesn't exist!\n");
            break;
}

// If we're in a busy state then check to see if we should give up.
// There's nothing shameful about quitting!
if (ent->a_state < 0 && random()<C_GIVEUP)
{
        VectorClear(ent->velocity);
        VectorCopy(ent->s.origin, ent->dest);

        // Come out of the busy state
        ent->a_state = -ent->a_state;
}

// If our action resulted in a place we should move to...
if (ent->dest[0] != -1)
{
        if (ent->dest[2] == 0)
        {
                gi.dprintf ("*** Animat_Think(): dest cleared by behavior, \
    should never get here!\n");
        }

        // If we're close enough to our destination, stop
        if (Dist(ent->dest, ent->s.origin) > 5)
        {
                VectorSubtract(ent->dest, ent->s.origin, ent->velocity);
                VectorNormalize(ent->velocity);
                VectorScale(ent->velocity,
                (ent->a_state==S_EXPLOREd?
                3*(C_VELOCITY):(C_VELOCITY)), ent->velocity);
        }
        // Otherwise, move towards it
        else
        {
                VectorClear(ent->velocity);
                VectorCopy(ent->s.origin, ent->dest);
        }
}
else
{
        gi.dprintf ("*** Animat_Think(): dest not set by behavior, \
    should never get here!\n");
}

// Make sure the animat stays on the plane of the 2D map
ent->s.origin[2] = C_MAPPLANE;
```

```
    // Check if the animat escapes the world somehow and put it back
    if (ent->s.origin[0] > C_MAPXMAX+C_GRIDSIZE ||
ent->s.origin[0] < C_MAPXMIN-C_GRIDSIZE ||
ent->s.origin[1] > C_MAPYMAX+C_GRIDSIZE ||
ent->s.origin[1] < C_MAPYMIN-C_GRIDSIZE)
    {
        gi.dprintf ("*** Animat_Think(): animat outside of world!\n");

        VectorSet(ent->s.origin,
      (((int)(random()*2))-1)*25 + ent->cache->s.origin[0],
      (((int)(random()*2))-1)*25 + ent->cache->s.origin[1],
     C_MAPPLANE);

        if (ent->s.origin[0] > C_MAPXMAX+C_GRIDSIZE ||
     ent->s.origin[0] < C_MAPXMIN-C_GRIDSIZE ||
     ent->s.origin[1] > C_MAPYMAX+C_GRIDSIZE ||
     ent->s.origin[1] < C_MAPYMIN-C_GRIDSIZE)
        {
            gi.dprintf ("*** Animat_Think(): cache outside of world! \
              putting animat in center of map\n");

            VectorSet(ent->cache->s.origin, (C_MAPXMAX-C_MAPXMIN)/2,
         (C_MAPYMAX-C_MAPYMIN)/2, C_MAPPLANE);
            gi.linkentity(ent->cache);

            VectorSet(ent->s.origin,
         (((int)(random()*2))-1)*25 + ent->cache->s.origin[0],
         (((int)(random()*2))-1)*25 + ent->cache->s.origin[1],
        C_MAPPLANE);
        }
        VectorClear(ent->velocity);
    }

    // Animats can't have less than 1 health
    ent->health = MAX(ent->health, 1);

    // Only update score if we just finished an action, not if we're busy
    if (ent->a_state >= 0)
    {
        // Update the score queue with our current health level
        ent->SQ[qi] = ent->health;

        // Increase combat power and decrease our just-attacked counter
        ent->combat_power = MIN(ent->combat_power + 0.1, 1);
        ent->just_attacked = MAX(ent->just_attacked - 0.2, 0);
    }

    // Link the animat with the world, and we'll be on our way!
    gi.linkentity(ent);
}
```

### 6.2.2 Calculating the Differentiation Factor

This code sample shows how the action differentiation factor is calculated for tribe *t*. The signal differentiation factor is calculated similarly.

```c
double GetTribeSeqDeviation(int t)
{
        // Declare and initialize the local variables
        edict_t *from=NULL;
        int i, j;
        double  src=0;
        double  srd=0;
        int             sr[C_ANIMATSPERTRIBE][S_NUM*S_NUM]={0};
        double  srs=0;

        // Initialize the entity pointer to the head of the array
        from = g_edicts;
        i = 0;

        // Loop through all the entities
        for ( ; from < &g_edicts[globals.num_edicts]; from++)
        {
                // If from isn't active, continue
                if (!from->inuse)
                        continue;

                // If from isn't an animat, continue
                if (Q_stricmp(from->classname, "~animat") != 0)
                        continue;

                // If from isn't in tribe t, continue
                if (from->tribe != t)
                        continue;

                // Copy the array that stores the count of the number of times each
                // behavior sequence was chosen to the local array
                memcpy(sr[i], from->a_state_count, S_NUM*S_NUM*sizeof(int));

                // Normalize the counts to get the frequency for each sequence
                RankSequences(sr[i], (S_NUM*S_NUM));

                // Increment the count of animats in tribe t that we've found so far
                i++;
        }

        // For each behavior sequence we're going to calculate the
        // standard deviations and add them up
        for (i=0;i<S_NUM*S_NUM;i++)
        {
                src = 0;
                srd = 0;

                // Calculate the mean sequence frequency for the tribe
                for (j=0;j<C_ANIMATSPERTRIBE;j++)
                {
                        src += (double)sr[j][i]/(double)C_ANIMATSPERTRIBE;
                }

                // Calculate the sum of the squares of the difference
                for (j=0;j<C_ANIMATSPERTRIBE;j++)
                {
                        srd += pow((sr[j][i] - src),2);
                }

                // Add up the standard deviations
                srs += sqrt(srd/(double)C_ANIMATSPERTRIBE);
        }

        // Find the mean standard deviation by dividing by the number of
        // sequences.  That's our Differentiation Factor!
        return srs/(S_NUM*S_NUM);
}
```

### 6.2.3 Evaluating the Neural Networks

This code sample shows how the Action Neural Network is evaluated. It selects a behavior and fills in the memory register with the appropriate values. The animat being trained is *ent*, and the *allow_random* parameter indicates whether or not the neural network should generate random behavior. The Signal Neural Network is evaluated similarly.

```
void Eval_State(edict_t *ent, int allow_random)
{
        // Declare and initialize the local variables
        int i, j;
        double ot=0;
        int cs=0, cm=0, hidden_to_use=NN_HIDDEN;
        static int r=0;

        // Clear the outputs and the intermediate values
        memset(ent->a_outputs, 0, NN_OUTPUTS*sizeof(double));
        memset(ent->a_vals, 0, NN_THRESH*sizeof(double));

        // If we're in the random start phase and or we're otherwise supposed
        // to act randomly, generate some random behavior
        if (allow_random == 1 && (ent->tribe > C_LEARNTRIBES ||
                (level.framenum < C_RANDFRAMES)))
        {
                // Set a random output high
                cs = random()*S_NUM;
                ent->a_outputs[cs] = 1;

                // Set the memory register outputs to random values
                for (i=0;i<I_NUM;i++)
                {
                        ent->i_state[i] = ent->a_outputs[OUT_I_START+i] = random();
                }
                return;
        }
        // Otherwise...
        else
        {
                // normal operation
        }

        // Use the default number of hidden units
        hidden_to_use = NN_HIDDEN;

        // Scale (0,1) inputs to (-1,1) range and put them into the input neurons
        for (i=0;i<NN_INPUTS;i++)
        {
                ent->a_vals[i] = (ent->a_inputs[i] * 2) - 1;
        }
```

```c
        // For every hidden unit...
        for (i=0;i<hidden_to_use;i++)
        {
                // For every input...
                for (j=0;j<NN_INPUTS;j++)
                {
                        // Multiply the inputs by the connection weights and populate the
                        // hidden neurons with their activation value
                        ent->a_vals[NN_HIDDEN_START+i] += ent->a_weights[AW_IH(j, i)] *
                                NN_Activation(ent->a_vals[NN_INPUTS_START+j]);
                }
        }

        // For every output neuron...
        for (i=0;i<NN_OUTPUTS;i++)
        {
                // For every hidden neuron...
                for (j=0;j<hidden_to_use;j++)
                {
                        // Multiply the hidden unit activation by the connection weights
                        // and populate the output neurons with their activation value
                        ent->a_vals[NN_OUTPUTS_START+i] += ent->a_weights[AW_HO(j, i)] *
                                NN_Activation(ent->a_vals[NN_HIDDEN_START+j]);
                }

                // Set the neural network outputs
                ent->a_outputs[i] = NN_Activation(ent->a_vals[NN_OUTPUTS_START+i]);
        }

        // Scale behavior outputs from (-1,1) range to (0,2) range
        for (i=OUT_STATES_START;i<=OUT_STATES_END;i++)
        {
                ent->a_outputs[i] += 1;
        }

        // For all the behavior outputs add the output bonus
        for (i=OUT_STATES_START;i<=OUT_STATES_END;i++)
        {
                ent->a_outputs[i] += C_OUTPUTBONUS;
        }

        // If we have a command in the command queue, multiple the appropriate
        // output by the obedience multiplier
        if (allow_random == 1 &&
                ent->CQ[ent->think_count%TQ_LENGTH].command != 0)
        {
                ent->a_outputs[ent->CQ[ent->think_count%TQ_LENGTH].command] *=
                        g_commult[ent->tribe];

        }
        else
        {}

        // Normalize the array of behavior outputs
        Array_Normalize(ent->a_outputs, OUT_STATES_START, OUT_STATES_END);

        // Select a random behavior based on the probabilities generated by
        // the neural network
        ot = random();
        for (i=OUT_STATES_START;i<=OUT_STATES_END;i++)
        {
                if (ot < ent->a_outputs[i])
                {
                        cs = i;
```

```
                        break;
                }
                else
                {
                        ot -= ent->a_outputs[i];
                }
        }

        // For all the memory register outputs...
        for (i=0;i<I_NUM;i++)
        {
                // Scale them from (-1,1) range to (0,1) range
                ent->a_outputs[OUT_I_START+i] = (ent->a_outputs[OUT_I_START+i]+1)/2;

                // Store the memory register outputs into the memory registers
                if (allow_random == 1)
                {
                        ent->i_state[i] = ent->a_outputs[OUT_I_START+i];
                }
        }

        // Clear the outputs of the neural network
        memset(ent->a_outputs, 0, OUT_I_START*sizeof(double));

        // Set the output of the selected behavior high
        ent->a_outputs[cs] = 1;
}
```

## 6.2.4 Training the Neural Networks

This section of code is the function used to train the Action Neural Network of animat *ent*. The Signal Neural Network is trained similarly.

```
void Train_State(edict_t *ent)
{
        // Declare and initialize the local variables
        int ts[TQ_LENGTH]={0};
        double o_error[NN_OUTPUTS]={0};
        double h_error[NN_HIDDEN]={0};
        double o_expected[NN_OUTPUTS]={0};
        double lr[TQ_LENGTH]={0}, lrp[TQ_LENGTH]={0};
        int i=0, j=0, k=0, top=0;
        int t1=0, t2=0, t3=0;
        double d1=0, delta=0;
        int sl=0, fg=0;
        double cur_improvement;
        int d_count=0,d_max=0;
        double no_error=0;

        // Temporary inputs and outputs used for saving the initial state
        double           t_inputs[NN_INPUTS];
        double           t_outputs[NN_OUTPUTS];

        // Use the default number of hidden units
        int hidden_to_use=NN_HIDDEN;
```

```c
        // Quit if tribe is not supposed to learn
        if (ent->tribe > C_LEARNTRIBES)
        {
                return;
        }

        // Quit if the animat hasn't filled up its queues yet
        if (ent->think_count < TQ_LENGTH)
        {
                return;
        }

        // Set ts[] to hold the index values in time order
        for (i=0;i<TQ_LENGTH;i++)
        {
                ts[i] = ((ent->think_count+i)%TQ_LENGTH);
        }
        //      ts[0] == current think, do not use
        //      ts[1] == farthest back data, most in the past
        //      ts[TQ_LENGTH - 1] == most recent good data
        //      in general, ts[i] is more recent than ts[i-1]

        // Save the current inputs and outputs for later
        memcpy(t_inputs, ent->a_inputs, NN_INPUTS*sizeof(double));
        memcpy(t_outputs, ent->a_outputs, NN_OUTPUTS*sizeof(double));

        // Clear the inputs and outputs
        memset(ent->a_inputs, 0, NN_INPUTS*sizeof(double));
        memset(ent->a_outputs, 0, NN_OUTPUTS*sizeof(double));

        // Loop through the queues
        for (top=TQ_LENGTH-1;top>1;top--)
// walk though backwards in time, increasing lr[] as we go
        {
                // Calculate the improvement between now and the previous think cycle
                cur_improvement = ent->SQ[ts[top]] - ent->SQ[ts[top-1]];

                // If there's no change, continue
                if (cur_improvement == 0)
                {
                        continue;
                }

                // Check the improvement level against the max known improvement
                // in order to scale the reward appropriately.  This allows us to
                // not know in advance what range of rewards will be given and makes
                // the learning routine more generic.
                (void) CheckMaxImprovement(cur_improvement);

                // Set the learning rate equal to the coefficient of learning times
                // the ratio of the current improvement to the max improvement.
                lr[ts[top]] = C_LCOEFFICIENT * (cur_improvement/g_max_improvement);

                // Set the current number of forgivenesses at 0.
                fg = 0;

                // Loop while the learning rate isn't zero, we haven't forgiven the
                // max number of times, and we haven't crossed the zero boundary.
                while (lr[ts[top]] != 0 && fg <= C_FORGIVESTATE && top - fg > 0)
                {
                        // Consider the time fg before the think cycle being trained

                        i = top - fg;

                        // Discount the learning rate based on how far we are back in
                        // time (as evidenced by how many states we've stepped back)
                        lr[ts[i]] = lr[ts[top]] * pow(C_LDISCOUNT, fg);
```

```
            // Increment the number of forgivenesses
            fg += 1;

            // If this behavior was the result of a command, utter a feedback
            // signal to everyone in range.
            if (ent->CQ[ts[i]].commander != NULL)
            {
                    if (ent->AQ[ts[i]][ent->CQ[ts[i]].command] == 1)
                    {
                            Train_SignalFeedback(ent, ts[i], lr[ts[i]]);
                    }
            }

            // Copy the input queue onto the neural network inputs
            memcpy(ent->a_inputs, ent->IQ[ts[i]], NN_INPUTS*sizeof(double));

            // Clear the error arrays
            memset(o_expected, 0, NN_OUTPUTS*sizeof(double));
            memset(o_error, 0, NN_OUTPUTS*sizeof(double));

            // Set the expected values to be the outputs from the actual
            // history.
            memcpy(o_expected, ent->AQ[ts[i]], NN_OUTPUTS*sizeof(double));

            // Evaluate the neural network and set allow_random to 0
            Eval_State(ent, 0);

            // For every output...
            for (j=0;j<NN_OUTPUTS;j++)
            {
                    // Calculate the difference between the actual and expected
                    o_error[j] = (o_expected[j] - ent->a_outputs[j]);

                    // Save up the total error for later
                    no_error += fabs((o_expected[j] - ent->a_outputs[j]));
            }

            // If there's no difference, move along
            if (no_error == 0)
            {
                    continue;
            }

            // For every hidden unit...
            for (j=0;j<hidden_to_use;j++)
            {
                    // Zero the hidden unit error
                    h_error[j] = 0;

                    // For every output...
                    for (k=0;k<NN_OUTPUTS;k++)
                    {
                            // Get the weight index for the connection between
                            // j and k
                            t1 = AW_HO(j, k);

                            // Get the neuron indexes
                            t2 = NN_HIDDEN_START+j;
                            t3 = NN_OUTPUTS_START+k;

                            // Calculate the hidden unit error
                            h_error[j] += (o_error[k]*ent->a_weights[t1]);

                            // If our error and learning rate are non-zero...
                            if (lr[ts[i]] != 0 && o_error[k] != 0)
                            {
```

```c
                                        // Calculate the weight delta
                                        delta = lr[ts[i]]*o_error[k]*
                                                (NN_Activation_d(ent->a_vals[t3]))*
                                        (NN_Activation(ent->a_vals[t2]));

                                        // Add the delta and bound the weight between (-1,1)
                                        ent->a_weights[t1] = BOUND(-1,
                                        ent->a_weights[t1] + delta, 1);
                                }
                        }
                }

                // For every input...
                for (j=0;j<NN_INPUTS;j++)
                {
                        // For every hidden unit...
                        for (k=0;k<hidden_to_use;k++)
                        {
                                // Get the index for the connection between j and k

                                t1 = AW_IH(j, k);

                                // Get the neuron indexes
                                t2 = NN_INPUTS_START+j;
                                t3 = NN_HIDDEN_START+k;

                                // If the learning rate is non-zero...
                                if (lr[ts[i]] != 0)
                                {
                                        // Calculate the delta
                                        delta = lr[ts[i]]*h_error[k]*
                                        ent->a_vals[t2]*
                                                (NN_Activation_d(ent->a_vals[t3]));
                                        // Add the delta and bound within (-1,1)
                                        ent->a_weights[t1] = BOUND(-1,
                                        ent->a_weights[t1] + delta, 1);
                                }
                        }
                }
                d_count = 0;
        }
        // Restore the inputs and outputs to what they were before
        memcpy(ent->a_inputs, t_inputs, NN_INPUTS*sizeof(double));
        memcpy(ent->a_outputs, t_outputs, NN_OUTPUTS*sizeof(double));
}
```

- Jim, K., & Giles, C. Talking Helps: Evolving Communicating Agents for the Predator-Prey Pursuit Problem. Artificial Life 6, 237-254 (2000).

- Kirby, S. Natural Language from Artificial Life. Artificial Life 8, 185-215 (2002).

- Kirkpatrick, S., Gelatt, C. D. and Vecchi, M. P. Optimization by Simulated Annealing, Science, Vol 220, Number 4598, pages 671-680 (1983).

- Klapper-Rybicka, M., Schraudolph, N., Schmidhuber, J. Unsupervised Learning in Recurrent Neural Networks. Proc. International Conference on Artificial Neural Networks, Vienna. Springer Verlag, Berlin (2001).

- Lachman, M., Számadó, S., and Bergstrom, C. Cost and conflict in animal signals and human language. Proceedings of the National Academy of Sciences, Volume 98, Number 23, pages 13189-13194 (2001).

- Marocco, D., Angelo, C., Nolfi, S. The Role of Social and Cognitive Abilities in the Emergence of Communication: Experiments in Evolutionary Robotics. EPSRC/BBSRC International Workshop Biologically-Inspired Robotics Bristol. pp. 174-181 (2002).

- Matsumoto, Makoto. Mersenne Twister random number generator (2004). (http://www.math.sci.hiroshima-u.ac.jp/~m-mat/MT/emt.html)

- Maynard Smith, J. & Harper, D. G. C. Animal Signals: Models and Terminology. *Journal of Theoretical Biology*, Volume 177, Issue 3, pages 305-311 (1995).

- Meister, M., Berry, M. The Neural Code of the Retina. Neuron, Vol. 22, 435–450, March, 1999.

- Noble, J., Di Paolo, E., Bullock, S. Adaptive factors in the evolution of signalling [sic] systems. Cangelosi, A & Parisi, D (editors) Simulating the Evolution of Language, pp. 53-77 Springer-Verlag (2002).

- Odell, J., Parunak, H. V., Fleischer, M. "The Role of Roles in Designing Effective Agent Organizations," in *Software Engineering for Large-Scale Multi-Agent Systems*, Garcia, A.; Lucena, C.; Zambonelli, F.; Omicini, A.; Castro, J. (eds.), Lecture Notes on Computer Science volume 2603, Springer, Berlin, 2003, pp 27-38.

- Oliphant, M. Formal Approaches to Innate and Learned Communication: Laying the Foundation for Language. Ph.D. Thesis, University of California San Diego, Cognitive Science Department (1997).

- Omlin, C. W., and Giles, C. L. "Extraction and Insertion of Symbolic Information in Recurrent Neural Networks", in *Artificial Intelligence and Neural Networks: Steps*
165